\documentclass[pdflatex,sn-mathphys-num]{sn-jnl}%

\usepackage{graphicx}%
\usepackage{multirow}%
\usepackage{amsmath,amssymb,amsfonts}%
\usepackage{amsthm}%
\usepackage{mathrsfs}%
\usepackage[title]{appendix}%
\usepackage{xcolor}%
\usepackage{textcomp}%
\usepackage{manyfoot}%
\usepackage{booktabs}%
\usepackage{algorithm}%
\usepackage{algorithmicx}%
\usepackage{algpseudocode}%
\usepackage{listings}%
\usepackage{xspace}%
\ifdefined\pdfsuppressptexinfo\pdfsuppressptexinfo=-1\fi%
\makeatletter\@twosidefalse\@mparswitchfalse\makeatother
\makeatletter%
\renewcommand\paragraph{\@startsection{paragraph}{4}{\z@}{1.2ex \@plus .4ex \@minus .2ex}{-0.6em}{\normalfont\normalsize\bfseries}}
\makeatother

\theoremstyle{thmstyleone}%

\theoremstyle{thmstyletwo}%

\theoremstyle{thmstylethree}%

\newcommand{\relay}{RELAY\xspace}
\newcommand{\uno}[1]{\textbf{#1}}%
\newcommand{\due}[1]{\underline{#1}}%
\newcommand{\tre}[1]{\textit{#1}}%
\newsavebox{\tabbox}%
\newcommand{\figpending}[2]{\fbox{\parbox[c][#1][c]{\dimexpr\linewidth-2\fboxsep-2\fboxrule\relax}{\centering\textcolor{red}{#2}}}}

\begin{document}

\newcommand{\papertitle}{An ECG foundation model for zero-shot transfer from hospital to point-of-care devices}
\title[Article Title]{\papertitle}

\author*[1]{\fnm{Edoardo} \sur{Coppola}}\email{ec2013@cam.ac.uk}
\equalcont{These authors contributed equally to this work.}
\author[2]{\fnm{Stefano} \sur{Fiorini}}\email{s.fiorini1994@gmail.com}
\equalcont{These authors contributed equally to this work.}
\author[1]{\fnm{Pietro} \sur{Liò}}\email{pl219@cam.ac.uk}
\author[3]{\fnm{Mattia} \sur{Savardi}}\email{mattia.savardi@unibs.it}
\author[3]{\fnm{Alberto} \sur{Signoroni}}\email{alberto.signoroni@unibs.it}
\affil*[1]{\orgdiv{Department of Computer Science and Technology}, \orgname{University of Cambridge}, \orgaddress{\street{William Gates Building, 15 JJ Thomson Avenue}, \city{Cambridge}, \postcode{CB3 0FD}, \country{United Kingdom}}}
\affil[2]{\orgname{Independent Researcher}, \orgaddress{\street{Pepparedsleden 1}, \city{Mölndal}, \postcode{431 83}, \country{Sweden}}}
\affil[3]{\orgdiv{Department of Medical and Surgical Specialties, Radiological Sciences and Public Health}, \orgname{University of Brescia}, \orgaddress{\street{Viale Europa, 11}, \city{Brescia}, \postcode{25123}, \country{Italy}}}

\abstract{
Electrocardiogram (ECG) recording is extending beyond hospitals to point-of-care devices that record one or two leads, where public labelled data is too scarce to build foundation models (FMs) as for 12-lead ECGs.
Whether existing 12-lead FMs transfer reliably to point-of-care ECGs without retraining remains unclear.
We show that seven published FMs reach a zero-shot AUROC of 41--81\% on two real point-of-care datasets and, under simulated reduction to one or two leads, lose 12--35 AUROC points on average and gain 0.15--0.23 in mean absolute error across 29 benchmarks covering 1,673 clinical targets.
We present RELAY, a 7-million-parameter FM that represents ECGs as graphs of segments from whichever leads are recorded, without filling missing ones or knowing their identities.
Pretrained on 9.2 million 12-lead ECGs, RELAY is competitive with 12 leads, ranks first on both point-of-care datasets and on 26 and 22 of 29 benchmarks with one and two leads sampled randomly, and keeps the most informative operating point when 12-lead thresholds are applied to device recordings.
Matched ablations show that omitting absent leads, rather than filling them, reduces what RELAY loses in transfer.
ECG FMs should therefore be judged across all acquisition settings, not on 12-lead benchmarks alone.
}


\maketitle

\section{Introduction}
\label{sec:introduction}

The electrocardiogram (ECG) is a cheap, non-invasive tool for assessing cardiac health.
In hospitals it is routinely recorded from 12 leads, each capturing a different projection of the heart's electrical activity~\cite{friedman2024electrocardiogram}. This abundance of recordings and annotations has driven the development of 12-lead ECG foundation models (FMs), pretrained on hundreds of thousands to millions of recordings and finetuned for tens to hundreds of diagnostic and prognostic targets~\cite{Li2025, Tian2024, Coppola2024, Na2024, liu2024zero, mckeen2025ecg, kim2024learning, almasud2026benchmarkingecgfmsreality}.
Meanwhile, ECG recording is increasingly extending beyond hospitals to point-of-care devices, such as handheld recorders, smartwatches and adhesive patches, which usually capture only one or two leads~\cite{el2026wrist}.
However, large collections of point-of-care ECG recordings are held privately by device manufacturers, while the publicly available ones are labelled almost exclusively for rhythm-related conditions~\cite{bouzid2022remote, Clifford2017, icentia11k}.
This data scarcity has constrained the development of point-of-care FMs, with the few existing models trained on proprietary datasets and not publicly released~\cite{abbaspourazad2023large}. Consequently, the pretraining and finetuning paradigm established for hospital ECGs cannot be readily replicated in point-of-care settings.

This leaves a practical question: can 12-lead models be used on point-of-care ECGs?
The closest evidence comes from task-specific networks trained on lead I extracted from 12-lead ECGs, which lost accuracy on smartwatch and handheld recordings~\cite{lin2026bridging}, and from the PhysioNet/Computing in Cardiology Challenge 2021, ``\texttt{Will two do?}'', in which scores changed by less than 2\% between the 12-lead and 2-lead test  sets~\cite{PerezAlday2020,Reyna2021,pollard_physionet_2026,Reyna2022}. Both, however, addressed %
a development question: the target lead set was known in advance and models were trained for it~\cite{Reyna2021,monachino2025self}. The Challenge also scored only 30 conditions defined by ECG criteria, for which the ECG is itself the diagnostic reference~\cite{Coppola2024}, whereas targets with non-ECG ground truth (e.g. imaging) remain less explored~\cite{dhingra2025artificial,alrumayh2025artificial,musa2026artificial}, and its reduced-lead inputs were 12-lead clinical recordings with leads removed, not point-of-care recordings. In contrast, we argue that a general-purpose ECG FM
should be pretrained once, before the leads of a future device are known, and be expected to serve many tasks and devices.

Therefore, we investigate the readiness of existing 12-lead FMs for the shift from hospital to point-of-care ECGs without retraining or adaptation, assessing how much of their 12-lead performance is retained and which design choices determine it. A key challenge lies in the input interface.
Most FMs expect a fixed 12-channel input, requiring missing leads to be filled, typically with zeros, to avoid computational failures. Besides introducing unnecessary computation, zero-filling has also been employed as a training augmentation to improve robustness to missing leads~\cite{oh2022, mckeen2025ecg}. Even approaches that accommodate reduced-lead inputs either bind each lead to a predefined input slot~\cite{Na2024} or average the feature maps of the available leads into a single representation, which is then attended to only along the temporal axis~\cite{monachino2025self}, discarding the cross-lead relationships.

Our findings highlight the limitations of existing 12-lead ECG FMs under reduced-lead inference. In particular, of seven published 12-lead ECG FMs finetuned on 12-lead recordings and transferred without retraining, six need the missing leads to be filled in order to run at all, and the seventh is tied to lead identity and fragile when a lead matches none of the standard twelve. On two real point-of-care datasets, from a handheld recorder and an adhesive patch, they reach a macro AUROC of only 41.3--80.8\% and 48.1--70.4\%, respectively. Extending previous simulation frameworks~\cite{Reyna2021,lin2026bridging}, in which leads are removed from 12-lead recordings, to 29 benchmarks spanning 13 datasets and 1,673 diagnostic, structural, functional and prognostic targets, we find that their mean AUROC drops by 11.9--34.9 points with one or two leads, along with an increase in mean absolute error (MAE) of 0.149--0.230 across five regression benchmarks.

Rather than adapting the input to the architecture, we make the architecture independent of lead count and lead identity. We introduce \relay, a 7.1-million-parameter FM that represents any ECG as a variable-size spatiotemporal graph, whose nodes correspond to short segments of the recorded leads and whose edges follow electrophysiologically motivated intra- and inter-lead relationships. A missing lead simply yields a smaller graph: no zero-filling or explicit lead-identity information is required. At its core, \relay employs a Graph Attention Network (GAT)~\cite{velickovic2017graph}, whose computation scales naturally with the number of available leads. \relay is pretrained on 9.2 million 12-lead ECGs from five countries through masked node modelling with random lead sampling, leveraging inter-lead redundancy~\cite{kors1990reconstruction, nelwan2000minimal} to learn representations robust to lead availability.
Transferred in the same way as the other FMs, \relay ranks first on 26 and 22 of 29 benchmarks with one and two randomly sampled leads, on 15 of 29 benchmarks for each of lead I and lead II, and on both real point-of-care datasets. At 12 leads, where no FM consistently dominates, \relay achieves competitive performance despite being more than 12 times smaller than the largest models. With thresholds chosen on 12-lead data and applied unchanged to device recordings, it also keeps the most informative operating point. Across the four reduced-lead configurations, its mean AUROC drops by 7.7--13.0 points relative to 12-lead inference, compared with 11.9--19.9 points for the best-performing baseline in each configuration.
Matched ablations further reveal the importance of explicitly modelling lead absence. Within \relay, filling absent leads with zeros instead of omitting them reduces zero-shot transfer performance on a held-out development set by 5.6--6.4 AUROC points, with most of the degradation arising at inference time rather than during pretraining. Separately, removing random lead sampling during pretraining reduces performance by 1.8--3.9 points.

These results suggest that current 12-lead ECG FMs are not robust to the realities of point-of-care acquisition, where lead availability, one of multiple factors of variation, is often incomplete. Within \relay, how absent leads are represented is a key determinant of this robustness. By treating lead absence as an explicit architectural primitive, and reinforcing this representation through random lead sampling, \relay narrows the gap between the 12 leads on which FMs are trained and the few that reach them at the point of care.

\begin{figure}[htb!]
\centering
\IfFileExists{figures/out/Fig1-Relay.drawio.pdf}{\includegraphics[width=\linewidth]{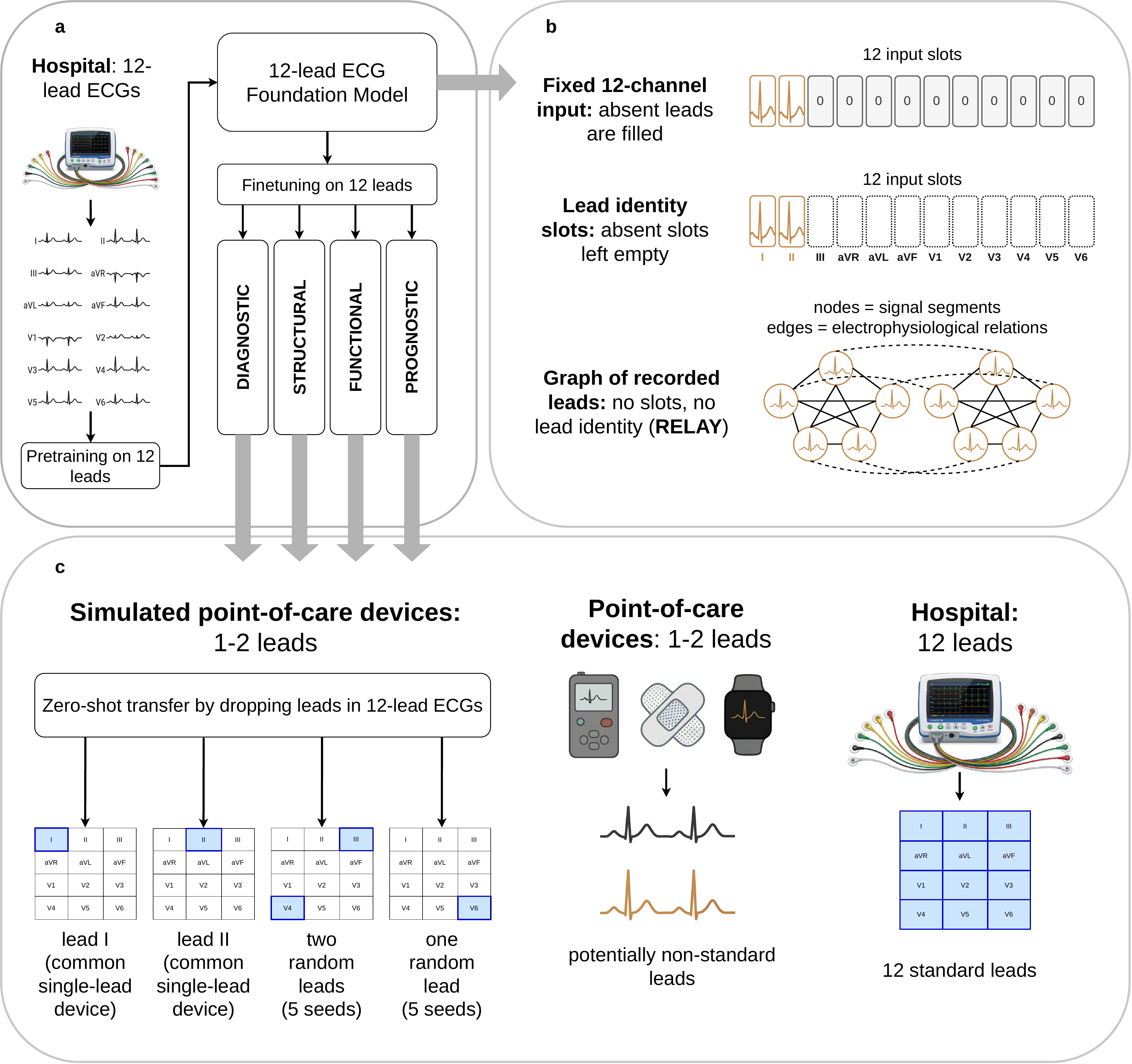}}{\figpending{70mm}{Fig.~1 pending}}
\caption{
\textbf{Study design.}
\textbf{a,} ECGs recorded in hospital have 12 leads and are used for pretraining and finetuning 12-lead ECG FMs.
\textbf{b,} When current ECG FMs receive a recording with absent leads,
six of them have a fixed 12-channel input in which absent traces are filled (e.g. with zeros),
one has a template of lead-identity slots in which absent slots are left empty (ST-MEM).
\relay represents ECGs as a graph that contains only the recorded leads.
\textbf{c,} After pretraining and finetuning on 12 leads, all ECG FMs (including \relay) are evaluated on 12 leads and,
without retraining (zero-shot transfer), on four simulated reduced-lead configurations
(lead I, lead II, and one and two leads sampled randomly, over five seeds) and on two real point-of-care datasets.}
\label{fig:overview}
\end{figure}

\section{Results}

\subsection{Transferring 12-lead FMs to fewer leads without retraining}
\label{sec:overview}

We ask how 12-lead ECG FMs, finetuned on 12-lead recordings, perform when applied without retraining to the fewer leads recorded by point-of-care devices.
We evaluate seven published FMs that differ in architecture, pretraining objective, and size (Table~\ref{tab:models}):
ECGFounder~\cite{Li2025}, ECG-JEPA~\cite{kim2024learning}, ST-MEM~\cite{Na2024}, MERL~\cite{liu2024zero}, ECGFM-KED~\cite{Tian2024},
HuBERT-ECG~\cite{Coppola2024} and ECG-CPC~\cite{almasud2026benchmarkingecgfmsreality}. We compare them with our \relay, introduced below.
We design \relay both to be competitive with 12 leads and to address the shift from hospital to point-of-care devices through zero-shot transfer. To ensure a fair comparison, it is finetuned and evaluated under the same protocol as the baselines and, like them, sees no point-of-care data at any stage.

Each model, including \relay, is finetuned once per dataset on 12-lead recordings, with no data augmentation and no removal of leads, following a unified and established benchmarking protocol~\cite{almasud2026benchmarkingecgfmsreality},
and is then evaluated without further training (zero-shot transfer) under five input configurations (Fig.~\ref{fig:overview}c):
all 12 leads; lead I alone and lead II alone, the leads most often recorded by single-lead devices; and one or two leads drawn at random for each record from a uniform distribution.
To assess overall lead coverage, random draws are generated for five seeds and shared by all models,
so that every model receives the same leads and the comparisons are paired and fair (Section~\ref{sec:methods-evaluation-protocol}).
Following established practice~\cite{Li2025, Tian2024, Coppola2024, Na2024, liu2024zero, mckeen2025ecg, kim2024learning, almasud2026benchmarkingecgfmsreality},
we report macro AUROC for classification targets and mean absolute error (MAE) for continuous ones, with standard deviations over the five seeds and bootstrap 95\% CIs ($n=1000$);
a paired bootstrap of the difference ($n=1000$) marks models not significantly worse than the best on each benchmark
(asterisks in tables, dots in Figs.~\ref{fig:twelve} and~\ref{fig:sim}).
Results for every model, benchmark and input configuration are in Supplementary Tables~\ref{tab:supp_all12}--\ref{tab:supp_random2}.
Threshold-based metrics are deliberately read from the ROC curve of each test set, and on the real point-of-care recordings, we also apply thresholds chosen on 12-lead data,
as in deployment without labelled device recordings (Section~\ref{sec:methods-evaluation-protocol}).

The way in which a model receives fewer leads depends on its interface, which we audit model by model.
Six of the seven baselines receive zeros in place of the missing leads, and ST-MEM, which requires lead identities,
is run through its native reduced-lead path (Table~\ref{tab:models}; Fig.~\ref{fig:overview}b).

The evaluation comprises 29 benchmarks derived from 13 datasets, which cover diagnostic, structural, functional and prognostic tasks, and two real point-of-care datasets.
The real point-of-care datasets are CinC 2017~\cite{Clifford2017}, recorded with a handheld device,
and Icentia11k~\cite{icentia11k}, recorded with an adhesive patch, both with single-lead recordings.
Further information on the datasets, their roles, patient-level splitting and preprocessing is given in Section~\ref{sec:method-data-and-preprocessing}.

\begin{tableorg}%
\centering
\caption{
\textbf{FMs evaluated.}
Architecture, pretraining objective, size, and the way in which each model receives a recording with fewer than 12 leads.
Six baselines have a fixed 12-channel input and receive zeros in place of absent leads.
ST-MEM accepts an arbitrary number of leads but requires their identities.
\relay processes only the recorded leads, uses no lead identity and is pretrained with random lead sampling.
CNN, convolutional neural network; GNN, graph neural network; SSM, state-space model.}
\label{tab:models}
\scriptsize
\setlength{\tabcolsep}{2.5pt}
\renewcommand{\arraystretch}{0.9}
\begin{lrbox}{\tabbox}%
\begin{tabular}{@{}llllr@{}}
\toprule
Model & Architecture & Pretraining objective & Reduced-lead input & Params (M) \\
\midrule
ECGFounder~\cite{Li2025}          & CNN         & Supervised                     & Zero-filled           & 33.8 \\
ECG-JEPA~\cite{kim2024learning}   & Transformer & JEPA     & Zero-filled           & 87.2 \\
ST-MEM~\cite{Na2024}              & Transformer & Masked autoencoding            & Native, identity-bound & 90.3 \\
MERL~\cite{liu2024zero}           & CNN         & Weak superv., contrastive  & Zero-filled           & 4.6 \\
ECGFM-KED~\cite{Tian2024}         & CNN         & Bimodal contrastive            & Zero-filled           & 9.7 \\
HuBERT-ECG~\cite{Coppola2024}     & Transformer & Iter. masked modelling     & Zero-filled           & 97.2 \\
ECG-CPC~\cite{almasud2026benchmarkingecgfmsreality} & SSM         & CPC  & Zero-filled           & 3.8 \\
\midrule
\relay (ours)                            & GNN         & Iter. masked modelling     & Native, identity-free & 7.1 \\
\bottomrule
\end{tabular}%
\end{lrbox}%
\ifdim\wd\tabbox>\linewidth \resizebox{\linewidth}{!}{\usebox{\tabbox}}\else \usebox{\tabbox}\fi
\end{tableorg}

\subsection{Twelve-lead performance of the baselines}
\label{sec:twelve}

\begin{figure*}[htb!]
\centering
\IfFileExists{figures/out/fig2_twelve_lead.pdf}{\includegraphics[width=\linewidth]{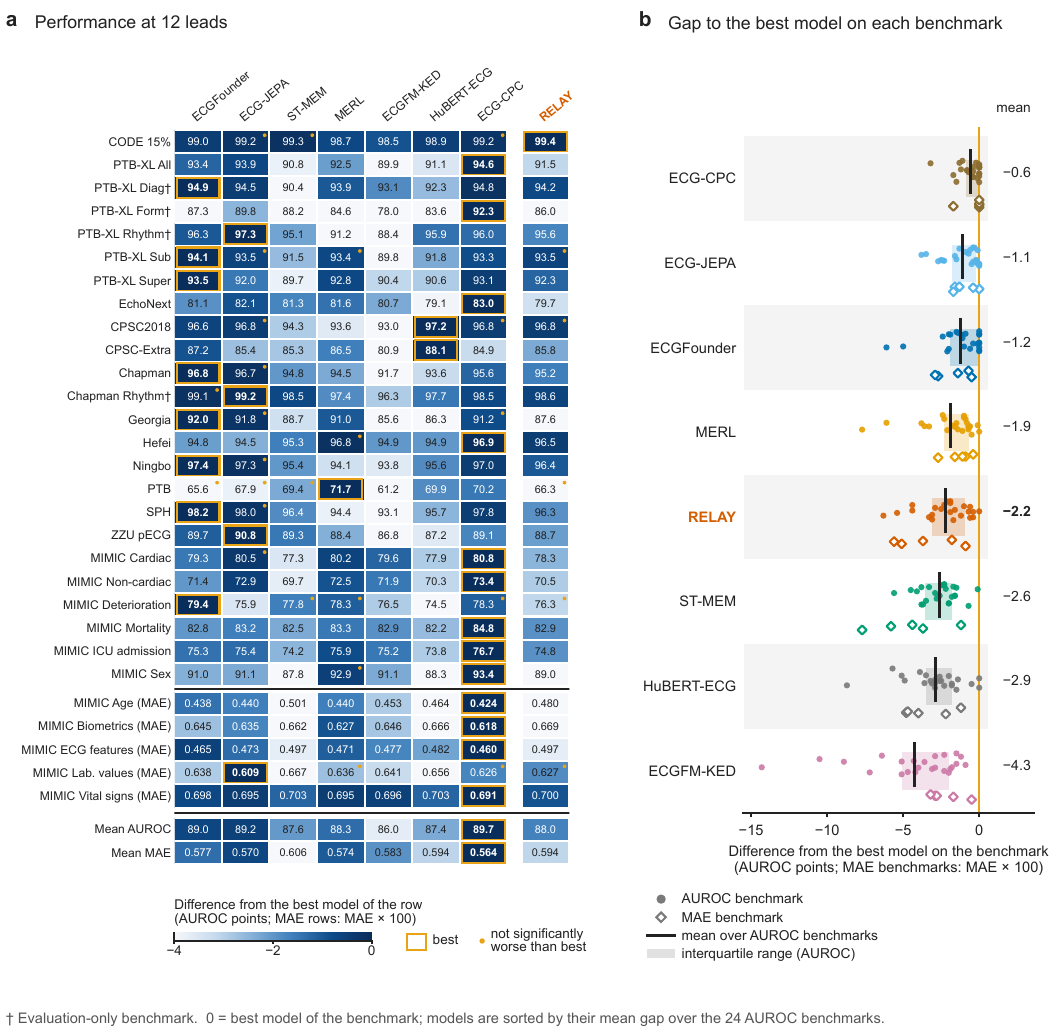}}{\figpending{85mm}{Fig.~2 pending}}
\caption{\textbf{Performance of every model at 12 leads.} \textbf{a,} Macro AUROC (\%) of the seven baselines and \relay on the 24 classification benchmarks, and mean absolute error (MAE, lower is better) on the five MIMIC-IV-ECG regression benchmarks, when finetuned and tested on 12 leads. Colour shows the difference from the best model of each row (AUROC points, or MAE $\times100$); outlines mark the best model, and dots mark models not significantly worse than it (paired bootstrap). The bottom rows give the mean AUROC over the 24 classification benchmarks and the mean MAE over the five regression benchmarks. \textbf{b,} Gap of each model to the best model on each benchmark (AUROC points; for the regression benchmarks, MAE $\times100$): points are benchmarks (circles, AUROC; diamonds, MAE), shaded boxes the interquartile range over the 24 AUROC benchmarks, and black bars and the numbers on the right their mean; models are sorted by mean gap. Values for every benchmark, with 95\% CIs, are in Supplementary Table~\ref{tab:supp_all12}. $^\dagger$Evaluation-only benchmark.}
\label{fig:twelve}
\end{figure*}

With all 12 leads, the mean AUROC of the baselines ranges from 86.0\% (ECGFM-KED) to 89.7\% (ECG-CPC), with ECG-JEPA and ECGFounder close behind (89.2\% and 89.0\%), and their mean MAE from 0.564 (ECG-CPC) to 0.606 (ST-MEM) (Fig.~\ref{fig:twelve}a; Supplementary Table~\ref{tab:supp_all12}).
Among the baselines, ECG-CPC ranks first on 13 of 29 benchmarks, ECGFounder on 8, ECG-JEPA on 4, HuBERT-ECG on 2, and ST-MEM and MERL on 1 each;
ECGFM-KED ranks first on none (Fig.~\ref{fig:twelve}a).
On 13 of the 24 benchmarks evaluated by AUROC, two or more baselines are not statistically worse than the best model (dots in Fig.~\ref{fig:twelve}a), including three of the six MIMIC-IV-ECG benchmarks (cardiac outcomes, clinical deterioration and sex).
On half of the benchmarks, the AUROC of the best and the third-best baseline differs by 1.1 AUROC points or less (maximum 4.1 AUROC points, on PTB-XL Form).
These 12-lead AUROC values, which are the reference for the performance drops reported below,
indicate that \emph{no FM achieves performance consistently better than that of competitors across benchmarks}.

\subsection{Zero-shot transfer of the baselines to real point-of-care recordings}
\label{sec:poc}

We first evaluate the models on recordings from real devices.
These contain every difference between hospital and device at once:
fewer leads, different electrode placement, consumer-grade hardware, and noise from unsupervised acquisition.
Both are labelled almost exclusively for rhythm and ectopy~\cite{bouzid2022remote}: CinC 2017 for sinus rhythm (SR) and atrial fibrillation (AFIB), and Icentia11k for SR, AFIB, atrial flutter (AFLT), premature ventricular contractions (PVC) and premature atrial contractions (PAC).
We therefore transfer models finetuned on 12-lead PTB-XL All, whose 71 labels include these statements, and map their outputs to the device labels without retraining, reporting macro AUROC on the device labels.

Models with a fixed 12-channel input receive the single lead in the lead I slot for CinC 2017,
since the handheld device (AliveCor) approximates lead I. For Icentia11k, whose patch lead matches no standard lead,
the single lead is placed in each of the 12 slots in turn, covering every possible configuration (Table~\ref{tab:poc}).
ECG-JEPA, whose input interface includes only the eight independent leads, receives the patch lead in each of the eight slots in turn.

On CinC 2017, the macro AUROC of the baselines ranges from 41.3\% (ECGFM-KED) to 80.8\% (ECG-JEPA),
and on Icentia11k, the mean macro AUROC over slots ranges from 48.1\% (MERL) to 70.4\% (ECG-JEPA) (Table~\ref{tab:poc}).

The AUROC of the baselines varies with the slot: it ranges from 51.6\% (aVR) to 73.9\% (V1) for ECGFounder, from 67.1\% (V6) to 74.0\% (V2) for ECG-JEPA, and from 50.0\% (III) to 82.0\% (V4) for ST-MEM.
Per label (Table~\ref{tab:poc}, bottom), the baseline with the highest AUROC is ECG-JEPA for AFIB on CinC 2017 (87.6\%) and for AFIB, AFLT, PVC and PAC on Icentia11k, ECGFounder for SR on CinC 2017 (83.4\%), and ST-MEM for SR on Icentia11k (72.5\%); for PAC, the AUROC of all baselines is between 50.5\% and 57.9\%.

\begin{table*}[htb!]
\centering
\setlength{\tabcolsep}{3pt}
\renewcommand{\arraystretch}{0.9}
\resizebox{\textwidth}{!}{
\begin{tabular}{ll cccccccc}
\toprule
\multirow{2}{*}{Dataset} & Lead slot & ECGFounder & ECG-JEPA & ST-MEM & MERL & ECGFM-KED & HuBERT-ECG & ECG-CPC & \textbf{RELAY} \\
        & or label & (33.8M) & (87.2M) & (90.3M) & (4.6M) & (9.7M) & (97.2M) & (3.8M) & \textbf{(7.1M)} \\
\midrule
\multicolumn{10}{c}{\textit{Macro AUROC}} \\
\midrule
CinC2017 & lead I & \tre{80.6} {\scriptsize (79.3--81.9)} & \due{80.8} {\scriptsize (79.8--81.7)} & 49.9 {\scriptsize (48.3--51.6)} & 63.5 {\scriptsize (61.8--65.3)} & 41.3 {\scriptsize (39.2--43.5)} & 61.2 {\scriptsize (60.0--62.4)} & 59.4 {\scriptsize (57.8--60.9)} & \uno{87.6} {\scriptsize (86.6--88.6)} \\
\midrule
\multirow{13}{*}{Icentia11k}
 & I & \tre{61.4} {\scriptsize (60.7--62.1)} & \due{67.5} {\scriptsize (66.8--68.1)} & 52.5 {\scriptsize (51.9--53.1)} & 51.3 {\scriptsize (50.6--52.0)} & 49.2 {\scriptsize (48.5--49.9)} & 56.9 {\scriptsize (56.2--57.6)} & 56.4 {\scriptsize (55.6--57.1)} & \uno{81.5} {\scriptsize (81.2--81.9)} \\
 & II & 59.4 {\scriptsize (58.6--60.2)} & \tre{71.5} {\scriptsize (71.0--72.0)} & \due{79.9} {\scriptsize (79.5--80.4)} & 45.8 {\scriptsize (45.0--46.6)} & 46.6 {\scriptsize (45.9--47.5)} & 57.7 {\scriptsize (57.0--58.5)} & 55.9 {\scriptsize (55.2--56.7)} & \uno{81.5} {\scriptsize (81.2--81.9)} \\
 & III & \due{57.9} {\scriptsize (57.1--58.7)} & N/A & 50.0 {\scriptsize (49.3--50.6)} & 52.4 {\scriptsize (51.6--53.1)} & 49.6 {\scriptsize (48.8--50.3)} & 55.7 {\scriptsize (55.0--56.4)} & \tre{56.8} {\scriptsize (56.1--57.5)} & \uno{81.5} {\scriptsize (81.2--81.9)} \\
 & aVR & 51.6 {\scriptsize (50.8--52.5)} & N/A & 52.3 {\scriptsize (51.6--52.9)} & 52.0 {\scriptsize (51.3--52.8)} & 48.0 {\scriptsize (47.3--48.8)} & \due{58.5} {\scriptsize (57.9--59.3)} & \tre{57.1} {\scriptsize (56.4--57.8)} & \uno{81.5} {\scriptsize (81.2--81.9)} \\
 & aVL & \tre{61.5} {\scriptsize (60.8--62.3)} & N/A & \uno{81.8} {\scriptsize (81.5--82.3)} & 45.1 {\scriptsize (44.2--45.9)} & 50.9 {\scriptsize (50.2--51.5)} & 55.6 {\scriptsize (54.8--56.4)} & 52.6 {\scriptsize (52.0--53.2)} & \due{81.5} {\scriptsize (81.2--81.9)}* \\
 & aVF & \due{60.7} {\scriptsize (59.9--61.4)} & N/A & 50.9 {\scriptsize (50.3--51.6)} & 51.1 {\scriptsize (50.3--51.8)} & 50.1 {\scriptsize (49.4--50.8)} & 55.6 {\scriptsize (54.9--56.4)} & \tre{56.1} {\scriptsize (55.3--56.9)} & \uno{81.5} {\scriptsize (81.2--81.9)} \\
 & V1 & \tre{73.9} {\scriptsize (73.3--74.6)} & 72.3 {\scriptsize (71.7--72.9)} & \due{75.4} {\scriptsize (74.9--75.8)} & 50.4 {\scriptsize (49.7--51.0)} & 49.0 {\scriptsize (48.3--49.8)} & 56.2 {\scriptsize (55.4--57.1)} & 59.2 {\scriptsize (58.4--60.0)} & \uno{81.5} {\scriptsize (81.2--81.9)} \\
 & V2 & 69.0 {\scriptsize (68.4--69.8)} & \due{74.0} {\scriptsize (73.4--74.5)} & \tre{72.2} {\scriptsize (71.7--72.7)} & 50.9 {\scriptsize (50.1--51.6)} & 49.9 {\scriptsize (49.1--50.7)} & 58.0 {\scriptsize (57.2--58.8)} & 58.9 {\scriptsize (58.2--59.5)} & \uno{81.5} {\scriptsize (81.2--81.9)} \\
 & V3 & 70.6 {\scriptsize (69.9--71.3)} & \tre{72.4} {\scriptsize (71.7--72.9)} & \due{77.5} {\scriptsize (77.0--78.1)} & 44.6 {\scriptsize (43.8--45.3)} & 48.7 {\scriptsize (47.9--49.4)} & 58.9 {\scriptsize (58.2--59.7)} & 56.8 {\scriptsize (56.3--57.4)} & \uno{81.5} {\scriptsize (81.2--81.9)} \\
 & V4 & 60.0 {\scriptsize (59.3--60.8)} & \tre{70.1} {\scriptsize (69.4--70.7)} & \uno{82.0} {\scriptsize (81.6--82.5)} & 43.6 {\scriptsize (42.9--44.3)} & 49.9 {\scriptsize (49.3--50.6)} & 58.0 {\scriptsize (57.3--58.8)} & 55.2 {\scriptsize (54.6--55.8)} & \due{81.5} {\scriptsize (81.2--81.9)} \\
 & V5 & \due{69.9} {\scriptsize (69.2--70.6)} & \tre{68.0} {\scriptsize (67.4--68.7)} & 53.6 {\scriptsize (53.0--54.2)} & 45.2 {\scriptsize (44.4--45.8)} & 50.4 {\scriptsize (49.7--51.2)} & 57.8 {\scriptsize (57.1--58.6)} & 56.1 {\scriptsize (55.3--56.9)} & \uno{81.5} {\scriptsize (81.2--81.9)} \\
 & V6 & 62.7 {\scriptsize (62.0--63.5)} & \tre{67.1} {\scriptsize (66.6--67.7)} & \due{76.8} {\scriptsize (76.4--77.3)} & 44.5 {\scriptsize (43.8--45.2)} & 49.6 {\scriptsize (48.9--50.4)} & 57.2 {\scriptsize (56.5--58.0)} & 57.9 {\scriptsize (57.2--58.6)} & \uno{81.5} {\scriptsize (81.2--81.9)} \\
\cmidrule{2-10}
 & Mean & 63.2 {\scriptsize (62.7--63.8)} & \due{70.4} {\scriptsize (69.9--70.8)} & \tre{67.1} {\scriptsize (66.8--67.4)} & 48.1 {\scriptsize (47.5--48.6)} & 49.3 {\scriptsize (48.7--50.0)} & 57.2 {\scriptsize (56.5--57.9)} & 56.6 {\scriptsize (56.1--57.1)} & \uno{81.5} {\scriptsize (81.2--81.9)} \\
\midrule
\multicolumn{10}{c}{\textit{Per-label AUROC}} \\
\midrule
\multirow{2}{*}{CinC2017}
 & AFIB & \tre{77.9} {\scriptsize (76.5--79.3)} & \uno{87.6} {\scriptsize (86.2--88.9)} & 48.8 {\scriptsize (46.7--51.1)} & 62.0 {\scriptsize (59.7--64.2)} & 41.0 {\scriptsize (38.6--43.4)} & 52.7 {\scriptsize (50.7--54.6)} & 61.8 {\scriptsize (59.5--64.0)} & \due{86.1} {\scriptsize (84.9--87.2)} \\
 & SR & \due{83.4} {\scriptsize (82.1--84.8)} & \tre{74.0} {\scriptsize (72.2--75.7)} & 51.0 {\scriptsize (48.7--53.4)} & 65.1 {\scriptsize (62.7--67.3)} & 41.7 {\scriptsize (39.4--44.0)} & 69.8 {\scriptsize (67.8--71.9)} & 57.0 {\scriptsize (55.1--59.1)} & \uno{89.0} {\scriptsize (87.9--90.2)} \\
\midrule
\multirow{5}{*}{Icentia11k}
 & SR & \tre{71.7} {\scriptsize (70.8--72.7)} & 63.1 {\scriptsize (62.2--64.0)} & \due{72.5} {\scriptsize (71.9--73.0)} & 47.4 {\scriptsize (46.2--48.6)} & 42.2 {\scriptsize (40.6--43.7)} & 65.2 {\scriptsize (63.9--66.5)} & 57.1 {\scriptsize (56.3--57.9)} & \uno{87.8} {\scriptsize (87.2--88.5)} \\
 & AFIB & 69.4 {\scriptsize (68.1--70.6)} & \due{79.7} {\scriptsize (78.3--81.1)} & \tre{75.1} {\scriptsize (74.5--75.7)} & 57.5 {\scriptsize (56.2--58.8)} & 48.5 {\scriptsize (46.7--50.4)} & 54.5 {\scriptsize (52.8--56.3)} & 65.3 {\scriptsize (64.2--66.4)} & \uno{91.5} {\scriptsize (90.8--92.1)} \\
 & AFLT & 61.0 {\scriptsize (59.1--63.0)} & \due{74.0} {\scriptsize (72.2--75.9)} & \tre{68.8} {\scriptsize (67.9--69.7)} & 36.2 {\scriptsize (34.2--38.1)} & 60.2 {\scriptsize (58.4--62.1)} & 56.8 {\scriptsize (54.7--59.1)} & 63.2 {\scriptsize (61.2--65.4)} & \uno{88.0} {\scriptsize (87.0--89.0)} \\
 & PVC & 58.3 {\scriptsize (57.9--58.7)} & \due{77.1} {\scriptsize (76.7--77.6)} & \tre{62.6} {\scriptsize (62.4--62.8)} & 46.9 {\scriptsize (46.5--47.3)} & 44.3 {\scriptsize (43.9--44.7)} & 58.8 {\scriptsize (58.5--59.2)} & 46.4 {\scriptsize (46.0--46.7)} & \uno{84.0} {\scriptsize (83.6--84.4)} \\
 & PAC & 55.8 {\scriptsize (55.5--56.1)} & \uno{57.9} {\scriptsize (57.6--58.3)} & \due{56.5} {\scriptsize (56.3--56.7)} & 52.4 {\scriptsize (52.0--52.7)} & 51.5 {\scriptsize (51.2--51.7)} & 50.5 {\scriptsize (50.2--50.8)} & 50.7 {\scriptsize (50.5--51.0)} & \tre{56.3} {\scriptsize (55.9--56.7)} \\
\bottomrule
\end{tabular}
}
\caption{\textbf{Zero-shot transfer to real point-of-care recordings.} Macro AUROC (top) and per-label AUROC (bottom), in \%, of the models finetuned on 12-lead PTB-XL All (heads mapped to the dataset labels: CinC 2017 AFIB, SR; Icentia11k SR, AFIB, AFLT, PVC, PAC), with 95\% CIs (bootstrap of the test set, 1000 resamples). The single recorded lead is placed in one lead slot of a zero-filled 12-lead input: lead I for CinC 2017 (AliveCor); every slot for Icentia11k, whose patch lead has no standard identity. \relay uses no lead identity, so its output does not depend on the slot. N/A (not available): ECG-JEPA uses only the 8 independent leads (I, II, V1--V6), so a lead placed in III, aVR, aVL or aVF never reaches it. Mean, and every per-label value for Icentia11k: mean over the lead slots (ECG-JEPA: its 8 applicable slots), with the CI of the slot mean. Parameter counts are in parentheses. *: not significantly worse than the best model of the row (paired bootstrap of the difference, 1000 resamples, 95\% CI includes 0). Best, second-best and third-best values are in bold, underlined and italic type.}
\label{tab:poc}
\end{table*}

\subsection{Performance of the baselines under simulated lead reduction}
\label{sec:sim}

Because the labels of CinC2017 and Icentia11k are limited to rhythm and ectopy, we also simulate one axis of the
hospital-to-device shift, the number of leads. We do so by evaluating the 12-lead finetuned models on reduced-lead
inputs from the clinical test sets of 29 benchmarks from 13 datasets covering 1,673 diagnostic, structural,
functional and prognostic targets. This mimics prior works~\cite{Reyna2021,lin2026bridging} but does not involve
training on the reduced-lead configurations.
With standard electrode placement, the simulation changes the number of leads and no other property of the signal.

\begin{figure*}[htb!]
\centering
\IfFileExists{figures/out/fig3_reduced.pdf}{\includegraphics[width=\linewidth]{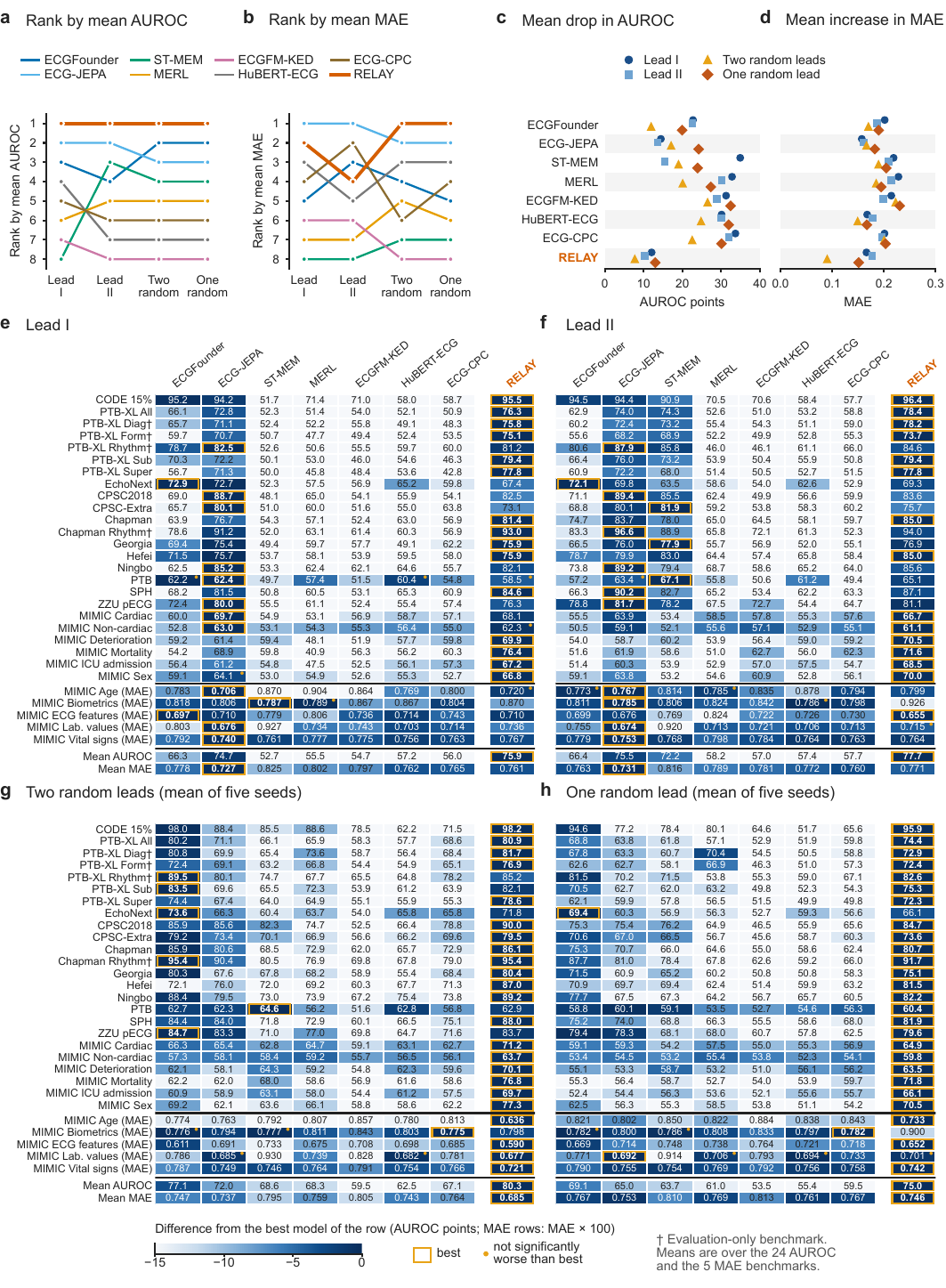}}{\figpending{160mm}{Fig.~3 pending}}
\caption{\textbf{Performance of every model under simulated lead reduction.} Models finetuned on 12-lead ECGs are evaluated without retraining with lead I, lead II, two leads sampled randomly and one lead sampled randomly; random-lead values are means over five seeds. \textbf{a,b,} Rank of each model by mean AUROC over the 24 classification benchmarks (\textbf{a}) and by mean MAE over the five MIMIC-IV-ECG regression benchmarks (\textbf{b}) in each reduced-lead configuration (the 12-lead ranking is in Fig.~\ref{fig:twelve}). \textbf{c,} Mean drop in AUROC from 12 leads (12-lead AUROC minus reduced-input AUROC of the same model on the same benchmark, averaged over the 24 classification benchmarks). \textbf{d,} Mean increase in MAE from 12 leads, averaged over the five regression benchmarks. \textbf{e--h,} Macro AUROC (\%) on the 24 classification benchmarks and MAE (lower is better) on the five regression benchmarks with lead I (\textbf{e}), lead II (\textbf{f}), two leads sampled randomly (\textbf{g}) and one lead sampled randomly (\textbf{h}); the bottom rows give the mean AUROC and the mean MAE. Colour shows the difference from the best model of each row (AUROC points, or MAE $\times100$; differences of 15 or more share the lightest colour); outlines mark the best model, and dots mark models not significantly worse than it (paired bootstrap). Values with 95\% CIs are in Supplementary Tables~\ref{tab:supp_leadI}--\ref{tab:supp_random2}. $^\dagger$Evaluation-only benchmark.}
\label{fig:sim}
\end{figure*}

With lead I, the mean AUROC of the baselines ranges from 52.7\% (ST-MEM) to 74.7\% (ECG-JEPA), with lead II from 57.0\% (ECGFM-KED) to 75.5\% (ECG-JEPA),
with two leads sampled randomly from 59.5\% (ECGFM-KED) to 77.1\% (ECGFounder), and with one lead sampled randomly from 53.5\% (ECGFM-KED) to 69.1\% (ECGFounder) (Fig.~\ref{fig:sim}e--h).
The mean drop from 12 leads ranges from 14.5 (ECG-JEPA) to 34.9 (ST-MEM) AUROC points with lead I, from 13.6 (ECG-JEPA) to 32.0 (ECG-CPC) AUROC points with lead II,
from 11.9 (ECGFounder) to 26.5 (ECGFM-KED) AUROC points with two leads sampled randomly, and from 19.9 (ECGFounder) to 32.4 (ECGFM-KED) AUROC points with one lead sampled randomly (Fig.~\ref{fig:sim}c).
With one lead sampled randomly, the mean drop of the baselines is largest on PTB-XL Superclass (36.4 AUROC points), MIMIC-IV-ECG Sex classification (34.8), PTB-XL Subclass (33.2) and PTB-XL All (32.9), and smallest on PTB (11.5).
It is 27.6 AUROC points on the rhythm benchmarks (PTB-XL Rhythm and Chapman Rhythm) and 27.1 AUROC points on the other benchmarks.

On the five MIMIC-IV-ECG regression benchmarks, the mean MAE of the baselines ranges from 0.564 to 0.606 with 12 leads, from 0.727 (ECG-JEPA) to 0.825 (ST-MEM) with lead I,
from 0.731 (ECG-JEPA) to 0.816 (ST-MEM) with lead II, from 0.737 (ECG-JEPA) to 0.805 (ECGFM-KED) with two leads sampled randomly,
and from 0.753 (ECG-JEPA) to 0.813 (ECGFM-KED) with one lead sampled randomly (Fig.~\ref{fig:sim}e--h).
Its increase from 12 leads ranges from 0.149 (HuBERT-ECG, two leads sampled randomly) to 0.230 (ECGFM-KED, one lead sampled randomly) (Fig.~\ref{fig:sim}d).

The ranking of the baselines differs between fixed and random leads (Fig.~\ref{fig:sim}a,b).
ECG-JEPA has the smallest drop on lead I and lead II (14.5 and 13.6 AUROC points) and a drop of 24.2 AUROC points with one lead sampled randomly.
ECGFounder has a drop of 22.7 and 22.6 AUROC points on lead I and lead II and 19.9 AUROC points with one lead sampled randomly.

ST-MEM, run through its native reduced-lead path, depends on which lead it receives: its mean drop is 34.9 AUROC points with lead I but 15.5 with lead II, 19.0 with two leads sampled randomly and 23.9 with one.

\subsection{\relay}
\label{sec:relay}

\begin{figure*}[htb!]
\centering
\IfFileExists{figures/out/ECG_lettered.pdf}{\includegraphics[width=\linewidth]{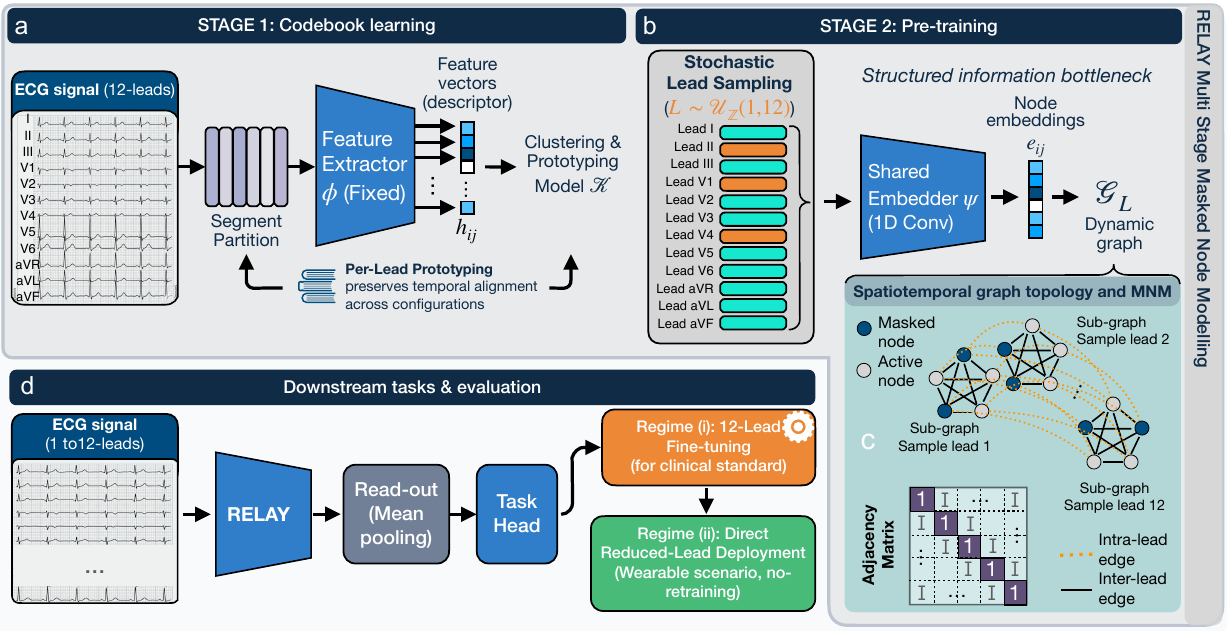}}{\figpending{100mm}{Fig.~6 pending}}

\caption{
\textbf{\relay architecture and pretraining.}
\textbf{a,} Feature extraction and clustering.
The 12 leads of every pretraining ECG are split into segments, each segment of each lead is described individually (stage 1: mel-frequency cepstral coefficients; stage 2: latent node representations from the stage-1 network), and a clustering model fitted on all descriptors assigns every segment to a class.
\textbf{b,} Masked node modelling with random lead sampling.
For each ECG, a random subset of leads is sampled, with the number of leads drawn uniformly from 1 to 12; a segment embedder (1D convolution) maps each segment of the sampled leads to a node embedding, part of the nodes of each lead is masked, and a graph attention network predicts the classes of the masked nodes.
Feature extraction, clustering and masked node modelling are repeated in stage 2 with the refined classes.
\textbf{c,} Electrophysiologically motivated spatiotemporal graph topology.
Nodes are segments; edges connect all segments within a lead and time-aligned segments across leads.
The adjacency matrix has all-ones blocks on the diagonal (within a lead) and identity blocks elsewhere (across leads).
\textbf{d,} After pretraining, \relay is finetuned on 12-lead ECGs, with a read-out by mean pooling and a task head, and, like the baselines, is evaluated without retraining with 12 leads, lead I, lead II, and one or two leads sampled randomly.}
\label{fig:arch}
\end{figure*}

To test whether absent leads, and lead identity, need to be represented at all, we develop \relay (REduced-Lead trAnsfer of electrocardiographY), a FM that processes only the leads actually recorded, without any filling and without lead identity (Fig.~\ref{fig:arch}).

Like HuBERT-ECG~\cite{Coppola2024}, \relay is developed through an iterative process that combines (1) feature extraction, (2) clustering and (3) masked modelling (Fig.~\ref{fig:arch}a,b; Section~\ref{sec:methods-relay}).
In each stage, the 12 leads of every pretraining ECG are split into segments, each segment of each lead is described individually by a feature descriptor, and a clustering model fitted on these descriptors assigns every segment to a class.
The segments then become the nodes of a spatiotemporal graph, whose electrophysiologically motivated edges connect all segments within a lead and time-aligned segments across leads (Fig.~\ref{fig:arch}c).
During masked node modelling, a random subset of leads is sampled for each ECG, with the number of leads drawn uniformly from one to twelve; part of the nodes of each sampled lead is masked, and a graph attention network learns to predict the classes of the masked nodes from the rest of the graph.
In stage 1, segments are described by mel-frequency cepstral coefficients; in stage 2, by the latent representations of the corresponding nodes, extracted from the stage-1 network, and masked node modelling is repeated with the refined classes~\cite{Hsu2021, Coppola2024}.
Pretraining uses 9.2 million 12-lead ECGs from five countries and involves uniform random lead sampling to become robust to any input lead configuration (Sections~\ref{sec:method-data-and-preprocessing} and~\ref{sec:methods-representation}).
After pretraining, \relay is finetuned on 12-lead recordings, with a task head on the mean-pooled node embeddings, and, like the baselines, is evaluated with 12 leads, lead I, lead II, and one or two leads sampled randomly (Fig.~\ref{fig:arch}d).

\paragraph{At 12 leads}
Under the same protocol as the baselines, \relay enters a landscape in which no model is consistently best across benchmarks (Fig.~\ref{fig:twelve}b).
Its mean AUROC is 88.0\%, within the range of the baselines (86.0--89.7\%) and higher than that of ST-MEM, HuBERT-ECG and ECGFM-KED, and its mean MAE on the regression benchmarks is 0.594, within the range of the baselines (0.564--0.606) (Fig.~\ref{fig:twelve}a).
\relay reaches this level with just 7.1 million parameters, more than 12 times fewer than the largest baselines (87.2--97.2 million), and with the smallest informative input of all models: 500 samples for each recorded lead, against 500 to 5{,}000 samples per lead (Table~\ref{tab:compute}).

\paragraph{On real point-of-care recordings}
The macro AUROC of \relay is 87.6\% on CinC 2017 and 81.5\% on Icentia11k, against 80.8\% and 70.4\% (mean over lead slots) for ECG-JEPA, the baseline with the highest AUROC on both datasets, a difference of 6.8 and 11.1 AUROC points (Table~\ref{tab:poc}).
\relay requires no slot assignment: the single lead is processed as a small graph originated by one lead, and its AUROC is 81.5\% in every slot.
\relay has the highest AUROC in 10 of the 12 Icentia11k slots, in each of which all baselines are significantly worse (paired bootstrap), and ST-MEM has a higher AUROC in two (aVL, 81.8\%; V4, 82.0\%).
Per label, \relay has the highest AUROC for SR on CinC 2017 (89.0\%, against 83.4\% for ECGFounder) and for SR, AFIB, AFLT and PVC on Icentia11k, and ECG-JEPA has a higher AUROC for AFIB on CinC 2017 (87.6\% versus 86.1\%) and for PAC on Icentia11k (57.9\% versus 56.3\%).
At 90\% specificity on the test ROC curves, \relay has the highest sensitivity for SR on both datasets and for AFIB, AFLT and PVC on Icentia11k (Supplementary Table~\ref{tab:supp_poc_thresholds}).
ECG-JEPA has the highest for AFIB on CinC 2017, and ST-MEM for PAC on Icentia11k, where no model detects more than 17\% of windows.
When the threshold for 90\% specificity is instead chosen on the 12-lead PTB-XL All validation set
and applied unchanged, most models lose their operating point (Supplementary Table~\ref{tab:supp_threshold_transfer}):
four baselines flag almost every recording as AFIB (specificity 0.0--6.2\%), and three detect at most 11.0\% of AFIB
recordings on CinC 2017. \relay degrades the least, with a balanced accuracy significantly higher than that of every
baseline on both datasets for AFIB and SR (paired bootstrap): 70.1\% and 83.5\% for AFIB (sensitivity 98.7\% and
94.1\%, specificity 41.4\% and 72.9\%), against 48.6--51.0\% for the baselines except ST-MEM on Icentia11k (65.4\%),
and 61.7\% and 72.1\% for SR, against at most 53.7\%. The same thresholds give \relay no advantage on the 12-lead
PTB-XL All test set. With thresholds chosen on lead-I inputs, \relay is significantly better than six baselines for
AFIB on CinC 2017 but worse than ECG-JEPA (73.7\% vs 79.9\%), and worse than ECGFounder and MERL for SR
(57.3\% vs 65.1\% and 59.0\%). On Icentia11k, a single threshold gives \relay the same operating point in every lead
slot, whereas the specificity of the baselines depends on the slot (e.g., from 7.0\% to 94.9\% for HuBERT-ECG for AFIB).

\paragraph{On simulated point-of-care inputs}
\relay performs best on 26 of 29 benchmarks with one lead sampled randomly, 22 of 29 with two leads sampled randomly (23 of 29 considering ties), 15 of 29 with lead I and 15 of 29 with lead II (Fig.~\ref{fig:sim}e--h).
Its mean AUROC is 75.0\%, 80.3\%, 75.9\% and 77.7\% with one lead sampled randomly, two leads sampled randomly, lead I and lead II, against 69.1\%, 77.1\%, 74.7\% and 75.5\% for the baseline with the highest mean AUROC in each configuration (ECGFounder with random leads, ECG-JEPA with lead I and lead II; Fig.~\ref{fig:sim}e--h).
The mean difference from the best baseline on each benchmark is $+5.0$, $+2.5$, $+1.2$ and $+1.5$ AUROC points in the same order, reaching $+12.1$ (MIMIC-IV-ECG mortality, one random lead), $+11.0$ (Hefei, two random leads) and $+10.3$ (MIMIC-IV-ECG clinical deterioration, lead II).
\relay's mean drop from 12 leads is 13.0, 7.7, 12.1 and 10.3 AUROC points, against 19.9, 11.9, 14.5 and 13.6 AUROC points for the baseline with the smallest drop in each configuration (the same baselines; Fig.~\ref{fig:sim}c).

On the MIMIC-IV-ECG regression benchmarks, \relay reaches the lowest mean MAE with one and two leads sampled randomly (0.746, 0.685), and ECG-JEPA has the lowest with lead I (0.727 versus 0.761 for \relay) and lead II (0.731 versus 0.771).
The increase in the mean MAE of \relay from 12 leads is 0.151, 0.090, 0.166 and 0.177, against 0.167, 0.149, 0.157 and 0.161 for the baseline with the smallest increase in each configuration (HuBERT-ECG with random leads, ECG-JEPA with lead I and lead II; Fig.~\ref{fig:sim}b,d).
\relay has also the lowest MAE with one and two leads sampled randomly on age, ECG features and vital signs, with two leads sampled randomly on laboratory values, and with lead II on ECG features; on biometrics, its MAE is the highest of all models with lead I, lead II and one lead sampled randomly (Fig.~\ref{fig:sim}e--h).
Additional threshold-dependent metrics (sensitivity at 90\% and 95\% specificity, and specificity at 80\% and 90\% sensitivity) are reported in Supplementary Tables~\ref{tab:supp_sens90spec}--\ref{tab:supp_spec90sens}.

\paragraph{Computational efficiency}
The cost of all baselines but ST-MEM is independent of the number of leads,
because their 12-channel inputs are processed in full, whereas that of \relay and ST-MEM grows with it
(Table~\ref{tab:compute}). With one lead, the setting of point-of-care devices, \relay has the highest throughput
(9{,}438 samples s$^{-1}$) and the lowest latency (0.108\,ms per sample), and its memory occupation (384\,MB)
is second only to that of ECGFounder. With 12 leads, its memory occupation is higher, because graph frameworks
store edge indices explicitly.

\begin{tableorg}
\centering
\caption{
\textbf{Computational cost.}
Input size (samples $\times$ leads), trainable parameters, floating-point operations of the forward (F) and backward (B) pass at batch size 1, and peak GPU memory, throughput (Thr.) and latency (Lat.) at batch size 64, on one NVIDIA A100 GPU.
For \relay and ST-MEM, GFLOPS values are reported as \texttt{min}-\texttt{max}; for memory, throughput and latency values are reported as \texttt{median}(\texttt{min}-\texttt{max}); the minimum corresponds to one lead and the maximum to 12 leads.
$\uparrow$, higher is better; $\downarrow$, lower is better.}
\label{tab:compute}
\scriptsize
\setlength{\tabcolsep}{2.5pt}
\begin{lrbox}{\tabbox}%
\begin{tabular}{@{}lcrcrrr@{}}
\toprule
Model & Input size & Params $\downarrow$ & GFLOP (F/B) $\downarrow$ & Memory (MB) $\downarrow$ & Thr. (samples s$^{-1}$) $\uparrow$ & Lat. (ms) $\downarrow$ \\
\midrule
ECGFounder (CNN)         & 1250 $\times$ 12 & 33.8M & 0.602 / 5.066   & 202.3  & 2{,}423.7 & 0.450 \\
ECG-JEPA (Transformer)   & 2500 $\times$ 12 & 87.2M & 73.877 / 221.6  & 2{,}269.3 & 39.9   & 10.131 \\
ST-MEM (Transformer)     & 600 $\times$ [1, 12] & 90.3M & 1.703-20.926 / 5.110-62.779 & 833.8 (703.1-938.3) & 825.4 (410.8–3962.4) & 1.447 (0.252–2.435) \\
MERL (CNN)               & 1250 $\times$ 12 & 4.6M  & 0.863 / 2.582   & 903.2  & 9{,}274.5 & 0.367 \\
ECGFM-KED (CNN)          & 5000 $\times$ 12 & 9.7M  & 5.423 / 16.258  & 1{,}236.4 & 811.5  & 0.613 \\
HuBERT-ECG (Transformer) & 500 $\times$ 12  & 97.2M & 18.829 / 69.779 & 1{,}685.8 & 194.4  & 4.676 \\
ECG-CPC (SSM)            & 600 $\times$ 12  & 3.8M  & 1.741 / 5.213   & 481.3  & 1{,}384.0 & 2.260 \\
\midrule
\relay (GNN)             & 500 $\times$ [1, 12] & 7.1M & 0.572-6.867 / 1.003-13.735 & 2{,}913.5 (384.0-5{,}858.5) & 1{,}723.6 (9{,}438.2-771.2) & 0.687 (0.108-1.297) \\
\bottomrule
\end{tabular}%
\end{lrbox}%
\ifdim\wd\tabbox>\linewidth \resizebox{\linewidth}{!}{\usebox{\tabbox}}\else \usebox{\tabbox}\fi
\end{tableorg}

\subsection{The advantages of random lead sampling and modelling lead absence}
\label{sec:ablation}

Because \relay differs from the baselines in pretraining corpus, objective and size, the comparisons above do
not isolate the contribution of random lead sampling or of representing absent leads by omission. We therefore
run matched experiments within \relay, in which only the factor under study changes: each variant is pretrained for
60k steps and assessed by linear evaluation, a linear layer trained on frozen representations, so that differences
reflect pretraining alone. All ablations use Ribeiro-dev, a held-out subset of CODE disjoint from all evaluation
benchmarks and labelled for six rhythm and conduction diagnoses.
Ablations of architecture, hyperparameters, codebook sizes and pretraining design are in Supplementary Section~\ref{sec:supp_ablations}.

\begin{figure*}[htb!]
\centering
\IfFileExists{figures/out/fig5_ablations.pdf}{\includegraphics[width=\linewidth]{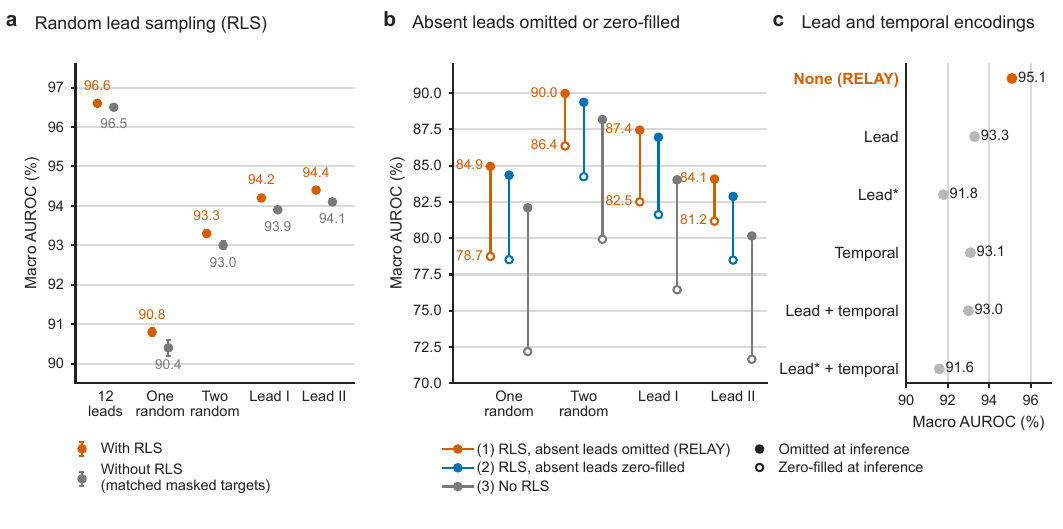}}{\figpending{60mm}{Fig.~5 pending}}
\caption{\textbf{Matched ablations within \relay.} Macro AUROC (\%) on the held-out Ribeiro-dev. \textbf{a,} Linear evaluation after pretraining with and without random lead sampling (RLS): a linear layer on frozen representations is trained and tested with the same input configuration. \textbf{b,} Zero-shot transfer: for each pretrained variant, one linear layer is trained on frozen 12-lead representations and tested without retraining with the absent leads omitted (filled circles) or filled with zeros (open circles); values are printed for variant 1 (\relay). \textbf{c,} Linear evaluation after pretraining with RLS, with temporal and lead encodings added under matched conditions. Temporal, sinusoidal temporal position embeddings; lead, learnable lead embeddings; lead*, learnable lead embeddings kept frozen during downstream evaluation. Random-lead values are means over five seeds; error bars in \textbf{a} are standard deviations, and standard deviations in \textbf{b} are at most 0.33 AUROC points. Values for every variant of \textbf{b}, including 12 leads, are in Supplementary Table~\ref{tab:supp_abl_transfer}.}
\label{fig:ablations}
\end{figure*}

First, \relay is pretrained with and without random lead sampling. Without it, every ECG contributes all
12 leads, so batch size and masking ratio are adjusted to give both models the same number of masked targets
per batch. For each model, a linear layer is trained and tested with each input configuration.
\emph{With random lead sampling, linear evaluation's AUROC is always higher than that without it} (Fig.~\ref{fig:ablations}a).

Second, we ask a different question: whether modelling lead absence is better than lead filling.
To do so, we compare three pretrained versions of \relay: (1) with random lead sampling and the non-sampled leads omitted; (2) with random lead sampling and the non-sampled leads filled with zeros; and (3) without random lead sampling.
In the second variant, the graph always contains 12 leads, with flat-signal nodes for the missing ones, mirroring random lead masking used in prior works~\cite{oh2022, mckeen2025ecg}; masking and pretraining targets are restricted to real leads, so its objective is identical to that of variant 1.
Each variant is linearly evaluated on 12 leads, that is, a linear layer is trained on its frozen 12-lead representations.
This single linear layer is then tested zero-shot, without retraining, on nine inputs: 12 leads, and one lead sampled randomly, two leads sampled randomly, lead I and lead II, each with the absent leads either omitted or filled with zeros.
With 12 leads, the three variants differ by at most 0.2 AUROC points (96.4--96.6\%; Supplementary Table~\ref{tab:supp_abl_transfer}).
\emph{For all three variants, omitting absent leads at inference gives a higher AUROC than filling them with zeros in every reduced configuration}: by 2.9 to 6.2 AUROC points for variant 1, 4.4 to 5.8 for variant 2 and 7.6 to 9.9 for variant 3 (Supplementary Table~\ref{tab:supp_abl_transfer}).
Filling absent leads with zeros both during pretraining and at inference (variant 2, zero-filled) instead of omitting them (variant 1, omitted) lowers the AUROC by 5.6 to 6.4 AUROC points, whereas zero-filling during pretraining alone (variant 2, omitted) lowers it by 0.5 to 1.2; pretraining without random lead sampling (variant 3, omitted) lowers it by 1.8 to 3.9.

\subsection{\relay uses no lead identity and does not model temporal order}
\label{sec:noidentity}

Unlike the previous section, this ablation, like all ablations in Supplementary Section~\ref{sec:supp_ablations}, compares variants by linear evaluation with random lead sampling, under matched and self-contained conditions: the same architecture, corpus, pretraining budget (60k steps), codebook and masking ratio, with only the encoding under study added to the node embeddings.
We add to the node embeddings learnable lead embeddings (providing lead identity), sinusoidal temporal position embeddings (proving temporal ordering), or both (Fig.~\ref{fig:ablations}c).
Without any encoding, the AUROC is 95.1\%. Although the ECG is a time series, adding the temporal order of
the segments lowers it by 2.0 AUROC points, and adding lead identity by 1.8 to 3.3 points (Fig.~\ref{fig:ablations}c).

\relay therefore does not model the temporal order of segments and does not use lead identity.
Order is implicitly retained only within each node embedding, whose receptive field spans about 400\,ms
and alignment across leads is retained by the inter-lead edges, which connect only time-aligned segments.
Without requiring neither the number nor the identity of the leads,
\relay can process recordings from leads that correspond to no standard lead,
such as the Icentia11k patch lead, without any assignment.
Moreover, without any encoding that breaks the symmetry between nodes,
every operation of \relay after the convolutional segment embedder is equivariant to
permutations of the leads and to permutations of the segment positions
applied identically to all leads, because both permutations leave the graph unchanged.
After the final mean pooling, the output is invariant to them as well.

\section{Discussion}
\label{sec:discussion}

Six of the seven published 12-lead ECG FMs require missing leads to be filled (e.g. with zeros)
to process reduced-lead inputs, while the seventh, ST-MEM, requires explicit lead identity.
Finetuned on 12-lead ECGs and transferred without retraining, these models exhibit substantial performance degradation
when only one or two leads are available, both on real handheld and patch recordings and on reduced inputs simulated across 29 benchmarks.
While they all reach strong performance on 12-lead ECGs, no baseline is consistently best across benchmarks.
More importantly, strong performance at full lead availability does not necessarily predict strong
performance with fewer leads (e.g., the baseline with the highest mean AUROC at 12 leads is not the
best under reduced-lead inference).
Operating points do not transfer either: on real point-of-care recordings, thresholds chosen on 12-lead data
do not preserve the specificity of most models, so a threshold chosen on hospital recordings cannot be assumed to hold on device recordings.
\relay, which processes only the recorded leads without binding them to lead identities or channel slots and
without filling the absent ones, is competitive at 12 leads with just 7.1 million parameters, fewer than five of the seven baselines
and more than 12 times fewer than the largest ones, and with the smallest input size of all models.
With one or two leads, it achieves the highest mean zero-shot AUROC across both real and simulated point-of-care ECGs.
Without recalibration, it also keeps the most informative operating point on both devices.
Moreover, because its predictions do not depend on assigning the recorded lead to a predefined input slot,
RELAY avoids the slot-dependent variability observed in the baselines.

Prior studies and challenges~\cite{Reyna2021, Reyna2022, lin2026bridging, monachino2025self} measure the performance that can be achieved when a model is developed for a target lead set known in advance, mostly on 12-lead recordings with leads removed rather than on real point-of-care ECGs.
Our study instead assesses zero-shot transfer to reduced-lead configurations without configuration-specific adaptation.
The two settings give complementary answers.
When models are developed for two leads, challenge scores change by less than 2\% from 12 leads~\cite{Reyna2021}.
Task-specific networks trained on lead I already lose accuracy on device recordings~\cite{lin2026bridging}.
When the lead set is not known in advance, the 12-lead FMs we evaluate lose 12--35 AUROC points on average with one or two of the 12 leads.
On real handheld and patch recordings, they reach a macro AUROC of only 41--81\%, extending to FMs the loss previously observed for task-specific networks.

The most plausible interpretation of the losses of the baselines is that how a model represents absent leads substantially influences its robustness under zero-shot transfer.
Zero-filling, the most common strategy~\cite{oh2022, mckeen2025ecg}, replaces a missing lead with a flat signal, the input a detached electrode produces, making lead absence indistinguishable from a recorded lead with no signal.
Binding each lead to a slot of a 12-lead template, as in ST-MEM~\cite{Na2024}, avoids filling absent leads but requires the
identity of every recorded lead and does not guarantee robust transfer, even when lead identities are known in advance.
Experiments within \relay support this interpretation: with architecture, data, budget and objective matched, omitting absent leads at inference gives
a higher AUROC than zero-filling for every pretrained variant, including the one pretrained with zero-filled leads,
with most of the loss arising at inference rather than during pretraining.
Random lead sampling during pretraining provides a complementary benefit, improving transfer whether absent leads are omitted or
zero-filled during pretraining. Explicit lead and temporal encodings, by contrast, do not improve \relay:
adding lead identity or the temporal order of the segments lowers AUROC in matched comparisons.
Local temporal information is still retained within each node representation, and temporal alignment across leads is preserved
by the inter-lead edges. Taken together, these results suggest that, for \relay, robustness to fewer leads comes from
random lead sampling during pretraining and from processing only the
recorded leads, without requiring explicit lead-identity information.
Because the 12 leads are largely redundant~\cite{kors1990reconstruction, nelwan2000minimal},
representations learned from varying lead subsets may retain much of the information available in the full lead set,
which would explain why \relay retains more of its performance when leads are absent.

Nevertheless, these interpretations have limits. The baselines differ from \relay and from each other in pretraining corpus, objective,
architecture, size and input length, so the comparison across models reflects their performance as released rather than isolating the effect
of how absent leads are represented. Within \relay, matched ablations isolate the effects of random lead sampling,
the representation of absent leads, and lead and temporal encodings by varying one component at a time,
thereby providing mechanistic evidence for its behaviour. These findings, however, were obtained on a single architecture and may
not generalize to other models.
Assessing the generalizability of these findings would require repeating the same ablations under matched conditions
for every FM, starting from its large-scale pretraining. The simulation varies only the number of leads, so other
sources of domain shift between hospital and point-of-care signals, such as noise, electrode placement,
acquisition settings and hardware~\cite{chan2020representing, an2022adaptive},
are captured in our evaluation only by the real recordings of CinC 2017 and Icentia11k.
These datasets are annotated only for rhythm and ectopy and come from two devices.
Consequently, for most diagnostic, structural, functional and prognostic targets, our findings are based exclusively on simulated lead reduction, in which, as in prior work~\cite{Reyna2021, lin2026bridging, monachino2025self}, the reduced-lead inputs keep the labels of the original 12-lead recordings. Whether pretraining on unlabelled point-of-care recordings further improves zero-shot transfer lies outside our scope,
but current public point-of-care datasets, far smaller than our pretraining corpus and each typically from a single device and site,
are unlikely to answer it.
Answering this question would require point-of-care recordings at the scale of millions, annotated beyond rhythm.
Finally, prospective validation on devices in clinical use remains essential before deployment.

This study therefore has two main implications. First, the versatility expected of an ECG FM should extend
beyond clinical tasks to acquisition settings. A model pretrained once to serve many tasks and devices should be
evaluated, and judged, across all ECG acquisition settings: full and reduced lead availability, hospital and point-of-care
recordings, and lead configurations known in advance as well as unknown or non-standard ones.
Performance on standard 12-lead benchmarks alone is insufficient to establish such versatility.
Second, the input interface of an ECG FM is a design decision with measurable consequences rather than a formatting detail:
a model that processes only the recorded leads avoids confusing absence with a flat signal, requires no explicit lead identity,
and avoids unnecessary computation on absent leads.
Given the limited availability of large-scale point-of-care data, zero-shot transfer remains a practical strategy
for extending the applicability of hospital-trained ECG FMs to point-of-care settings. How a model represents absent leads appears to play
an important role in the effectiveness of this transfer. \relay shows that both requirements can be met by a compact model.

\section{Methods}
\label{sec:methods}

\subsection{Ethics}
\label{sec:methods-ethics}

This study was conducted in accordance with the principles of the Declaration of Helsinki and with all applicable national, European, UK and institutional regulations governing research involving human data.
The work is a retrospective, secondary analysis of pre-existing electrocardiographic datasets.
No new human-subject data was collected, no participants were prospectively enrolled by the authors, and all data accessed by the authors had been fully de-identified by the originating institutions before release.
All datasets were used in compliance with their licences and data use agreements (listed in Data availability), including the formal request procedure of CODE, the access procedure of EchoNext and the credentialed access of PhysioNet for the MIMIC-IV clinical modules and MIMIC-IV-Ext-MDS-ED, for which the required CITI human-subjects research training was completed by the accessing author (E.C.).

The datasets were collected and released by their respective custodians, not by the authors, and each custodian obtained the relevant ethics approvals; because the present work is a secondary analysis of de-identified data already approved for research use, no additional ethics approval was required at the authors' institutions (University of Cambridge and University of Brescia), in line with the regulations on the secondary use of fully anonymised data in the European Union (Regulation (EU) 2016/679) and in the United Kingdom (UK General Data Protection Regulation and Data Protection Act 2018), under Article 4(1) and Recital 26 of both regulations.
The ethics approvals and informed-consent provisions originally granted for each dataset are reported below.

\paragraph{CODE, including CODE-15\%}
Approved by the Research Ethics Committee of the Universidade Federal de Minas Gerais (protocol 49368496317.7.0000.5149). Informed consent was obtained from all participants by the original investigators.

\paragraph{PTB-XL}
Approved by the Institutional Ethics Committee of the Physikalisch-Technische Bundesanstalt for publication of anonymised data in an open-access database (approval reference PTB-2020-1).

\paragraph{MIMIC-IV-ECG and MIMIC-IV}
Collected at the Beth Israel Deaconess Medical Center and approved by the Institutional Review Boards of the Beth Israel Deaconess Medical Center (Boston, MA, USA) and the Massachusetts Institute of Technology (Cambridge, MA, USA).
The requirement for individual patient consent was waived because the project did not affect clinical care and all protected health information was de-identified.
MIMIC-IV-ECG is openly accessible through PhysioNet, and the MIMIC-IV clinical modules and MIMIC-IV-Ext-MDS-ED were accessed through PhysioNet under the Credentialed Health Data Use Agreement.

\paragraph{Chapman and Ningbo}
Approved by the Institutional Review Boards of Shaoxing People's Hospital (Shaoxing Hospital, Zhejiang University School of Medicine) and Ningbo First Hospital of Zhejiang University, which waived the requirement for informed consent.

\paragraph{SPH}
Approved by the Institutional Review Board of Shandong Provincial Hospital, which waived the requirement for individual patient consent; the database is publicly released after full de-identification.

\paragraph{EchoNext}
Approved by the New York-Presbyterian/Columbia University Institutional Review Board, which waived the requirement for patient consent.

\paragraph{ZZU pECG}
Approved by the Ethics Committee for Scientific Research and Clinical Trials of the First Affiliated Hospital of Zhengzhou University (approvals 2024-KY-0221-003 and 2025-KY-0369-001), which waived the requirement to sign informed consent forms, including parental or legal-guardian consent for participants under 18 years of age, and approved the desensitisation of the ECG data for public scientific research.

\paragraph{IKEM}
Provided by the Institute for Clinical and Experimental Medicine (Prague, Czech Republic) and released publicly by~\cite{sejak2023} in anonymised form, with each recording paired only with an anonymised patient identifier and no human labels; no ethics approval or consent provision is reported in the original publication.

\paragraph{Icentia11k}
Approved, for both the study and the release of the data, by the ethics institutional review boards of the Universit\'e de Montr\'eal (\#CERSES-19-065-D).

\paragraph{Public challenge and benchmark releases}
The following datasets were assembled, fully de-identified and publicly released as research benchmarks by their originating institutions, and are used strictly under the terms of their public licences (listed in Data availability):
the PTB Diagnostic ECG Database, compiled by the Physikalisch-Technische Bundesanstalt from recordings collected at the Department of Cardiology of University Clinic Benjamin Franklin, Berlin;
CPSC2018 and CPSC-Extra, released by the organisers of the China Physiological Signal Challenge 2018;
Georgia, released through the PhysioNet/Computing in Cardiology Challenge 2020;
Hefei, released by the organisers of the Tianchi arrhythmia competition;
and CinC 2017, released through the PhysioNet/Computing in Cardiology Challenge 2017.
For these datasets, no ethics-related information beyond that reported by the original data custodians is available to the authors.

No identifiable individual-level information is presented in this manuscript.

\subsection{Data and preprocessing}
\label{sec:method-data-and-preprocessing}

\paragraph{Datasets and benchmarks}
The evaluation comprises 29 benchmarks derived from 13 datasets (Table~\ref{tab:datasets}).
Adult ECG interpretation is evaluated on CODE-15\%~\cite{Ribeiro2020}, PTB-XL~\cite{Wagner2020} (all statements, diagnostic, form, rhythm, subclass and superclass labels), CPSC2018 and CPSC-Extra~\cite{Liu2018}, Chapman~\cite{Zheng2020b} (all labels and rhythm labels), Georgia~\cite{Goldberger2000}, Hefei~\cite{hefei}, Ningbo~\cite{Zheng2020a}, PTB~\cite{Bousseljot2009} and SPH~\cite{Liu2022}.
Paediatric ECG interpretation is evaluated on ZZU pECG, and cardiac structure and function on EchoNext~\cite{poterucha2025detecting}, whose labels are echocardiographic findings.
MIMIC-IV-ECG~\cite{Gow2023} provides eleven benchmarks: six evaluated by AUROC (cardiac outcomes, non-cardiac outcomes, clinical deterioration, mortality, ICU admission and sex) and five regression benchmarks evaluated by MAE (age, biometrics, ECG features, laboratory values and vital signs).
Benchmarks marked with a dagger in Table~\ref{tab:datasets} contain a subset of the labels of another benchmark of the same dataset and are used for evaluation only: models are finetuned on the parent benchmark and evaluated on the label subset.
The two real point-of-care datasets, CinC 2017 and Icentia11k, are used for testing only.
All datasets are publicly available except the unlabelled partition and the 85\% labelled partition of CODE, which are available on request from the data owners.
We refer the reader to the original publication of each dataset, cited in Table~\ref{tab:datasets}, for details on acquisition, population and labelling.

\paragraph{Splits}
We adopt the training, validation and test partitions of the established benchmarking study of~\cite{almasud2026benchmarkingecgfmsreality}, which are split at the patient level.
For every dataset, test partitions are held out from every stage of pretraining and finetuning.
All models are finetuned, validated and tested on the same partitions under matched budgets and identical conditions.

\paragraph{Pretraining corpus}
\relay is pretrained on 9.2 million 12-lead ECGs from 11 datasets collected in 5 countries,
which include labelled and unlabelled collections (Table~\ref{tab:datasets}): CODE~\cite{Ribeiro2020},
IKEM~\cite{sejak2023}, PTB-XL~\cite{Wagner2020}, CPSC and CPSC-Extra~\cite{Liu2018}, Chapman~\cite{Zheng2020b},
Georgia~\cite{Goldberger2000}, MIMIC-IV-ECG~\cite{Gow2023}, Ningbo~\cite{Zheng2020a}, PTB~\cite{Bousseljot2009}
and SPH~\cite{Liu2022}.
Pretraining uses no clinical labels, but relies on pseudo-labels obtained through clustering.
This unsupervised step clusters feature descriptors (mel-frequency cepstral coefficients or latent representations) of segments of the ECGs that have never been extracted from or trained against downstream clinical labels, assigning each segment to a cluster.
This prevents any form of label leakage.
Moreover, using the same datasets for self-supervised pretraining and for supervised finetuning of FMs
is common practice in numerous domains~\cite{caron2021emerging,Hsu2021,Baevski2020,caron2018deep,azizi2021big,Pai2024,Mehari2022,haghighi2022dira, Richter2024, tang2025electrocardiogram},
and a recent survey has even documented no tangible benefit in such data reusage~\cite{VanBerlo2024}.
This is due to the fact that self-supervised pretraining does not use any labels, and therefore does not leak any information about the downstream tasks.

\paragraph{MIMIC-IV-ECG acute-care endpoints}
The acute-care targets of MIMIC-IV-ECG~\cite{Gow2023} follow the definitions of the MIMIC-IV-Ext-MDS-ED benchmark.
Each sample is an emergency department (ED) visit with an ECG recorded within 90\,min of arrival, and outcomes refer to events after this window: clinical deterioration and ICU admission within 24\,h, ICU admission during the stay, and in-stay, 1-, 7-, 28-, 90-, 180- and 365-day mortality (15 labels).
Mortality combines in-hospital deaths with out-of-hospital deaths from the Massachusetts state registry, ascertained up to one year after the last discharge.
All horizons ($\leq$365 days from ED arrival) therefore lie within the ascertainment window, and outcomes are fully observed rather than censored.
Fixed-horizon classification evaluated by AUROC is thus an appropriate formulation for these prognostic targets.
Clinical deterioration (6 labels), mortality (7 horizons) and ICU admission (2 labels) are evaluated as three separate benchmarks following~\citet{almasud2026benchmarkingecgfmsreality}.

\paragraph{Preprocessing}
Signals are preprocessed on the fly for the model in use.
Each baseline receives the sampling rate, window length and normalisation of its original implementation.
For \relay, signals are band-pass filtered between 0.05 and 47\,Hz, downsampled to 100\,Hz, and each individual lead is rescaled to $[-1, 1]$~\cite{Coppola2024,Natarajan2020}.

\begin{table*}
\centering
\caption{
\textbf{Datasets and benchmarks.}
Task, target type, number of targets, ECGs and patients, use in pretraining and in downstream evaluation (zero-shot transfer after finetuning on 12-lead ECGs), and country of origin.
Indented rows are benchmarks of the dataset above; a dagger marks benchmarks that contain a subset of the labels of another benchmark and are used for evaluation only.
$^\ddagger$10-s windows drawn from the recordings (Section~\ref{sec:methods-evaluation-protocol}).
n.a., not available.}
\label{tab:datasets}
\setlength{\tabcolsep}{3pt}
\renewcommand{\arraystretch}{0.9}
\resizebox{\textwidth}{!}{
\begin{tabular}{@{}lllrrrccl@{}}
\toprule
Dataset / benchmark & Task & Type & Targets & ECGs & Patients & Pretraining & Downstream evaluation & Country \\
\midrule
CODE~\cite{Ribeiro2020}              &                          &             &       & 8,200,069 & n.a.    & \checkmark &            & Brazil \\
\quad Unlabelled                     &                          &             & --    & 5,877,604 & n.a.    & \checkmark &            & Brazil \\
\quad Labelled                       & Adult ECG interpretation & Multi-label & 6     & 2,322,465 & n.a.    & \checkmark &            & Brazil \\
\quad\quad CODE 85\%                 & Adult ECG interpretation & Multi-label & 6     & 2,088,985 & n.a.    & \checkmark &            & Brazil \\
\quad\quad CODE-15\%                 & Adult ECG interpretation & Multi-label & 6     &   233,480 & 233,480 & \checkmark & \checkmark & Brazil \\
IKEM~\cite{sejak2023}                &                          &             & --    &    98,130 & n.a.    & \checkmark &            & Czech Republic \\
PTB-XL~\cite{Wagner2020}             & Adult ECG interpretation & Multi-label &       &    21,799 & 18,869  & \checkmark &            & Germany \\
\quad All                            &                          &             & 71    &           &         &            & \checkmark & \\
\quad Diagnostic$^\dagger$           &                          &             & 44    &           &         &            & \checkmark & \\
\quad Form$^\dagger$                 &                          &             & 19    &           &         &            & \checkmark & \\
\quad Rhythm$^\dagger$               &                          &             & 12    &           &         &            & \checkmark & \\
\quad Subclass                       &                          &             & 23    &           &         &            & \checkmark & \\
\quad Superclass                     &                          &             & 5     &           &         &            & \checkmark & \\
CPSC2018~\cite{Liu2018}              & Adult ECG interpretation & Multi-label & 9     &     6,867 & n.a.    & \checkmark & \checkmark & China \\
CPSC-Extra~\cite{Liu2018}            & Adult ECG interpretation & Multi-label & 33    &     3,441 & n.a.    & \checkmark & \checkmark & China \\
Chapman~\cite{Zheng2020b}            & Adult ECG interpretation & Multi-label &       &    10,646 & 10,646  & \checkmark &            & China \\
\quad All                            &                          &             & 42    &           &         &            & \checkmark & \\
\quad Rhythm$^\dagger$               &                          &             & 9     &           &         &            & \checkmark & \\
Georgia~\cite{Goldberger2000}        & Adult ECG interpretation & Multi-label & 50    &    10,286 & n.a.    & \checkmark & \checkmark & U.S.A. \\
Hefei~\cite{hefei}                   & Adult ECG interpretation & Multi-label & 29    &    20,036 & 20,036  &            & \checkmark & China \\
Ningbo~\cite{Zheng2020a}             & Adult ECG interpretation & Multi-label & 68    &    34,808 & n.a.    & \checkmark & \checkmark & China \\
PTB~\cite{Bousseljot2009}            & Adult ECG interpretation & Multi-label & 22    &       549 & 290     & \checkmark & \checkmark & Germany \\
SPH~\cite{Liu2022}                   & Adult ECG interpretation & Multi-label & 35    &    25,770 & 24,666  & \checkmark & \checkmark & China \\
ZZU pECG                             & Paediatric ECG interpretation & Multi-label & 58 &   12,328 & 10,350  &            & \checkmark & China \\
EchoNext~\cite{poterucha2025detecting} & Cardiac structure and function & Multi-label & 11 & 82,543 & 36,286 &          & \checkmark & U.S.A. \\
MIMIC-IV-ECG~\cite{Gow2023}          &                          &             &       &   800,036 & n.a.    & \checkmark &            & U.S.A. \\
\quad Cardiac outcomes               & Discharge diagnoses      & Multi-label & 158   &   114,355 & 49,400  &            & \checkmark & \\
\quad Non-cardiac outcomes           & Discharge diagnoses      & Multi-label & 918   &   178,163 & 81,930  &            & \checkmark & \\
\quad Acute care                     &                          &             & 15    &           &         &            &            & \\
\quad\quad Clinical deterioration    & Acute care prediction    & Multi-label & 6     &     5,577 & 4,595   &            & \checkmark & \\
\quad\quad Mortality                 & Acute care prediction    & Multi-label & 7     &    17,639 & 10,220  &            & \checkmark & \\
\quad\quad ICU admission             & Acute care prediction    & Multi-label & 2     &    18,690 & 13,868  &            & \checkmark & \\
\quad Patient characteristics: sex   & Patient characteristics  & Binary      & 1     &   182,076 & 83,736  &            & \checkmark & \\
\quad Patient characteristics: continuous &                     &             & 35    &           &         &            &            & \\
\quad\quad Age                       & Patient characteristics  & Regression  & 1     &   182,076 & 83,736  &            & \checkmark & \\
\quad\quad Biometrics                & Patient characteristics  & Regression  & 3     &   119,214 & 53,702  &            & \checkmark & \\
\quad\quad ECG features              & Patient characteristics  & Regression  & 7     &   181,989 & 83,721  &            & \checkmark & \\
\quad\quad Laboratory values         & Patient characteristics  & Regression  & 18    &    88,448 & 53,293  &            & \checkmark & \\
\quad\quad Vital signs               & Patient characteristics  & Regression  & 6     &   131,602 & 66,605  &            & \checkmark & \\
\midrule
CinC 2017~\cite{Clifford2017}        & Rhythm (handheld device) & Multi-class & 2     & 8,528 & n.a.   &            & Evaluation only & n.a. \\
Icentia11k~\cite{icentia11k}         & Rhythm and ectopy (patch) & Multi-label & 5    & 58,657$^\ddagger$ & 11,000 &         & Evaluation only & Canada \\
\bottomrule
\end{tabular}
}
\end{table*}

\subsection{Evaluation protocol}
\label{sec:methods-evaluation-protocol}

\paragraph{Finetuning}
We follow the unified benchmarking protocol of~\citet{almasud2026benchmarkingecgfmsreality} and re-run all experiments.
Each model is finetuned on the 12-lead training partition of each dataset for 100 epochs with a linear head, using the AdamW optimiser~\cite{Loshchilov2017} with a learning rate of $10^{-3}$ and a weight decay of $10^{-3}$.
No lead-aware augmentation is applied.Only the best checkpoint on the validation partition is kept, selected by macro AUROC or, for regression targets, by MAE.
Models are finetuned only on 12-lead ECGs, with no data augmentation and no removal of leads, and are evaluated without lead-specific adaptation.

\paragraph{Input configurations}
Each finetuned model is evaluated on the test partition with all 12 leads, lead I alone, lead II alone, and $L\in\{1,2\}$ leads drawn at random from a uniform distribution for each record.
Random draws are made without replacement, independently for each record, and are repeated for five seeds.
The seed is fixed during finetuning, and at inference all models receive the same lead subsets in the same order of samples, which makes the comparisons paired, reproducible and fair.

\paragraph{Reduced-lead inputs}
We use the public implementation and pretrained weights of each of the seven baselines (Table~\ref{tab:models}) and audit the input interface of each model in its code.
A model has a fixed 12-channel input when its first layer takes 12 input channels in a fixed order.
For these models, a recording with fewer leads is placed in the channels of the recorded leads, and the channels of the absent leads are set to zero after preprocessing and normalisation.
ECG-JEPA uses only the 8 independent leads (I, II, V1--V6) and derives no information from the channels of leads III, aVR, aVL and aVF.
ST-MEM accepts an arbitrary number of leads but requires their identities; it is given only the recorded leads, each with the learned embedding of its lead identity.
\relay builds a graph from the recorded leads only and receives no lead identity.

\paragraph{Transfer to real point-of-care recordings}
The models finetuned on 12-lead PTB-XL (all 71 labels) are applied to the device recordings without retraining.
The output heads are mapped to the labels of each dataset: AFIB and SR for CinC 2017, and SR, AFIB, AFLT, PVC and PAC for Icentia11k.
We report the macro AUROC over these labels and the AUROC of each label.
Icentia11k is evaluated on 10-s windows (2,500 samples at 250\,Hz), each lying entirely within one annotated rhythm episode (sinus rhythm, atrial fibrillation or atrial flutter), with no rhythm change.
Up to two windows per rhythm class and patient are drawn at random, giving 58,657 windows out of 158.5 million valid ones.
PVC and PAC labels are assigned from the beats annotated within each window, not from the rhythm class for which the window was drawn.
Device recordings are preprocessed on the fly depending on the model being used, analogously to what is done for other datasets.
For models with a fixed 12-channel input, the single recorded lead is placed in one lead slot of a zero-filled 12-lead input: the lead I slot for CinC 2017 (AliveCor recordings), and each of the 12 slots in turn for Icentia11k, whose patch lead has no standard identity.
For ECG-JEPA, a lead placed in slot III, aVR, aVL or aVF never reaches the model, so its Icentia11k mean is over 8 slots.
\relay uses no lead identity, so its output does not depend on the slot.

\paragraph{Metrics}
Over classification benchmarks, performance is primarily reported as macro-averaged AUROC and, for the five MIMIC-IV-ECG regression benchmarks, as mean absolute error (MAE) on z-scored targets, averaged over the targets of the benchmark without weighting.
As threshold-dependent metrics, we report sensitivity at 90\% and 95\% specificity, and specificity at 80\% and 90\% sensitivity, for the 18 benchmarks outside MIMIC-IV-ECG, macro-averaged over conditions (Supplementary Tables~\ref{tab:supp_sens90spec}--\ref{tab:supp_spec90sens}).
On the real point-of-care datasets, we report sensitivity and positive predictive value of each condition at 90\% specificity (Supplementary Table~\ref{tab:supp_poc_thresholds}).
Thresholds are set for every condition to compare models and not to propose clinical thresholds.
These values are deliberately read from the ROC curve of each test set, with the threshold re-derived in every bootstrap resample: like AUROC, they summarise discrimination, here in its high-specificity region, and are the highest values that any threshold can reach at that operating point.
To test whether operating points transfer, we also choose the threshold for 90\% specificity on the 12-lead PTB-XL All validation fold (2,183 records), following the standard protocol in which thresholds are fixed on validation data before testing, and apply it unchanged to the device recordings, as would be done without labelled device recordings, keeping it fixed in every bootstrap resample. On Icentia11k, the same threshold is used in every lead slot.
On CinC 2017, whose recordings correspond to lead I, we also choose it on lead-I inputs of the same partition, as a developer who knows the device lead would do.
We summarise each transferred operating point by its balanced accuracy, the mean of sensitivity and specificity, which is 50\% for a threshold that labels every recording as positive or every recording as negative.
We restrict this analysis to AFIB and SR, the only conditions labelled in both point-of-care datasets, with 151 and 1,670 positives in the validation set, and AF is the main target of rhythm screening with point-of-care devices~\cite{perez2019large, svennberg2021clinical, van20242024} (Supplementary Table~\ref{tab:supp_threshold_transfer}).

\paragraph{CIs and statistical tests}
In every input configuration, 95\% CIs are estimated by bootstrapping the test set (1000 resamples).
For $L\in\{1,2\}$, the five seeds are averaged within each resample before the percentiles are taken,
so that intervals refer to the reported seed mean.
For $L\in\{1,2\}$ we also report mean and standard deviation across the five seeds.
A model is marked as not significantly worse than the best model on a benchmark, or of a row of the point-of-care tables,
when the 95\% interval of the paired bootstrap distribution of the AUROC (or MAE) difference
(1000 resamples, the same resamples for every model and seed) includes zero.
All tests are two-sided.

\subsection{Building \relay}
\label{sec:methods-relay}

In this section, we describe the full development of \relay, from pretraining to finetuning and zero-shot transfer (Fig.~\ref{fig:arch}).

\subsubsection{Spatiotemporal graph topology}
\label{sec:methods-architecture}

Representing multichannel time series as graphs enables explicit modelling of the relationships (edges)
among signal segments (nodes)~\cite{Jin2024SurveyGNNTimeSeries}.
Existing ECG graph formulations either represent individual voltage samples as nodes~\cite{Qiang2024ConvRGNNECG, chen2025fusion},
which gives very large graphs for long, high-rate recordings,
or represent each lead as a single node~\cite{Moghaddam2024GraphBasedCardiac, Zeinalipour2023GNNTopologicalECG},
which discards the temporal structure within a lead.
We introduce a topology that retains physiological grounding, scalability, and both intra- and inter-lead dependencies.

Let $X\in\mathbb{R}^{L\times T}$ be an ECG with $L$ leads and $T$ samples.
Each lead $X_i$, $i\in\{1,\dots,L\}$, is partitioned into $S$ contiguous temporal segments $x_{ij}\in\mathbb{R}^{T/S}$, $j\in\{1,\dots,S\}$.
An embedding function $\psi$ maps each segment to a feature vector $e_{ij}=\psi(x_{ij})\in\mathbb{R}^{F}$, which gives lead-wise matrices $E_i\in\mathbb{R}^{S\times F}$ and, stacking the $L$ leads, the node feature matrix $E\in\mathbb{R}^{LS\times F}$.
The graph $\mathcal{G}=(\mathcal{V},\mathcal{E})$ has nodes $\mathcal{V}=\{v_{ij}\}$, with $|\mathcal{V}|=LS$, and edges $\mathcal{E}=\mathcal{E}_\mathrm{intra}\cup\mathcal{E}_\mathrm{inter}$, where
$\mathcal{E}_\mathrm{intra}=\bigcup_{i=1}^{L}\{(v_{ij},v_{ik})\mid j,k\in\{1,\dots,S\}\}$ connects all segments of the same lead, including each segment with itself (self-loops), and
$\mathcal{E}_\mathrm{inter}=\bigcup_{j=1}^{S}\{(v_{ij},v_{i'j})\mid i\neq i'\}$ connects time-aligned segments of different leads.
The adjacency matrix has all-ones blocks on the diagonal (within a lead, with self-loops) and identity blocks elsewhere (Fig.~\ref{fig:arch}c).
Graph construction requires no lead identity, and no lead embedding or temporal position embedding is added to the nodes.

Complete intra-lead connectivity allows message passing across the whole duration of a lead, analogously to self-attention within a lead,
so that both wave-to-wave and beat-to-beat relations can be modelled.
These edges define which segments communicate, not their temporal order, which is retained within the segment representations (\ref{sec:methods-implementation}).
Physiologically, intra-lead coupling reflects recurrent cardiac cycles and rate-dependent effects such as repolarisation hysteresis and restitution,
for which the interpretation of a segment benefits from preceding and following context.
Inter-lead edges encode the synchrony of leads, which are simultaneous projections of the
same cardiac electrical field, and allow cross-lead manifestations such as coordinated ST-segment
deviations or conduction asynchrony to be modelled without increasing the density of the graph further.

\subsubsection{Learning the codebook}

Predicting discrete, precomputed targets has been an effective self-supervised objective for speech~\cite{Hsu2021} and ECG~\cite{jin2025reading, Coppola2024}.
A codebook partitions the space of segment morphologies into $C$ classes of recurring patterns, whose assignments serve as targets for masked prediction.
Each lead of a 12-lead ECG is taken individually and divided into segments, and each segment is described by $h_{ij}=\phi(x_{ij})\in\mathbb{R}^{D}$,
where $\phi$ is a fixed feature extractor based on mel-frequency cepstral coefficients or latent representations.
Each descriptor therefore comes from a single lead, unlike in HuBERT-ECG~\cite{Coppola2024}, where the leads are concatenated before being segmented and described.
A single prototyping model $\mathcal{K}:\mathbb{R}^{D}\rightarrow\{1,\dots,C\}$ is
fitted on the descriptors of all leads and segments,
and assigns each segment to a class, $\tilde{y}_{ij}=\mathcal{K}(h_{ij})$.
The target of a segment therefore does not depend on which other leads are present.

\subsubsection{ECG representation learning}
\label{sec:methods-representation}

\paragraph{Random lead sampling}
The 12 leads of an ECG carry largely redundant information~\cite{kors1990reconstruction, nelwan2000minimal}, so a subset of a few leads retains much of the information.
We exploit this redundancy by sampling leads at random during pretraining, which encourages the model to learn representations of cardiac electrical activity that are robust to which leads are available.
For each 12-lead ECG in a mini-batch, we first draw a lead count $L\sim\mathcal{U}\{1,\dots,12\}$ and then a subset $\mathcal{L}\subseteq\{1,\dots,12\}$ of $L$ lead indices without replacement.
Only the corresponding rows of $X$ are retained, giving $X_\mathcal{L}\in\mathbb{R}^{L\times T}$.
Monachino et al.~\cite{monachino2025self} used random lead selection to generate positive pairs for their contrastive learning,
averaging the selected leads into a single representation and thereby discarding the cross-lead relationships that \relay models explicitly.

\paragraph{Graph construction}
Each retained lead is embedded by $\psi$, and the stacked embeddings $E\in\mathbb{R}^{LS\times F}$,
together with the topology above, give a graph $\mathcal{G}_\mathcal{L}$ whose $LS$ nodes change with the lead set.

\paragraph{Masked node modelling}
In each retained lead, a fraction $p_\mathrm{MASK} \in [0,1]$ of the nodes is masked,
so that every retained lead contributes masked targets and visible context.
Masked node embeddings are replaced by a learnable node embedding $e_\mathrm{MASK}$,
which corrupts $E$ to give $E'$.
A graph attention encoder $\varphi$~\cite{brody2021attentive} maps $(E',\mathcal{E})$ to
output node representations $O\in\mathbb{R}^{LS\times F}$,
and a linear head predicts a distribution over the $C$ classes of the prototyping model $\mathcal{K}$ for each masked node.
Training minimises $\mathcal{L}_\mathrm{MNM}=-\sum_{v_{ij}\in M}\log p(\tilde{y}_{ij}\mid E')$ over the set $M$ of masked nodes.
As in the graph topology above, $v_{ij}$ denotes the node of the $j$-th segment of the $i$-th retained lead, and $\tilde{y}_{ij}$ is the class assigned to that segment by $\mathcal{K}$.
The model therefore infers masked segments from the context within the lead and from time-aligned segments of other leads, under lead configurations that change at every step.
For downstream tasks, no node is masked, and the output node representations $O$ are mean-pooled and given to a linear task head.

\subsubsection{Transfer-guided codebook refinement}
\label{sec:methods-pretraining}

Following~\cite{Hsu2021, Coppola2024}, the pretraining targets are refined in a second pretraining stage.
After stage 1, all parameters of \relay are kept frozen and node representations are extracted
from a layer of the graph encoder (layer selection in Supplementary Fig.~\ref{fig:supp_oversmoothing_stage1}) and clustered into $C'$ classes by a second prototyping model $\mathcal{K}^{(2)}$
to produce the targets of stage 2 of masked node modelling (choice of $C'$ in Supplementary Fig.~\ref{fig:supp_sse} and Table~\ref{tab:supp_codebook_pdrop}).
Dense intra-lead connectivity and repeated aggregation can cause oversmoothing,
in which the variance between nodes collapses~\cite{chen2020measuring, zhao2020pairnorm},
degrading both target discovery and transfer. We counter it in two ways.
First, the checkpoint used for representation extraction is selected by transfer-guided
early stopping: a linear probe on a held-out development set (Ribeiro-dev,
a subset of CODE disjoint from all evaluation benchmarks)
is evaluated periodically on frozen representations, and the earliest checkpoint after
which the probe AUROC no longer improves materially is retained (Supplementary Fig.~\ref{fig:supp_oversmoothing_stage1}).
The probe is used only for checkpoint selection.
Second, in stage 2, intra-lead edges are dropped at random with probability $p_\mathrm{drop}$~\cite{rong2019dropedge},
which reduces temporal mixing while keeping inter-lead synchrony
(choice of $p_\mathrm{drop}$ and its interaction with $C'$ in Supplementary Fig.~\ref{fig:supp_pdrop} and Table~\ref{tab:supp_codebook_pdrop}).
No edges are dropped in stage 1, when maximal contextual aggregation benefits
learning of coarse semantic structures.
The stage-2 checkpoint is selected in the same way (Supplementary Fig.~\ref{fig:supp_oversmoothing_stage2}).

\subsubsection{Complexity analysis}

For an undirected graph, our topology has $|\mathcal{E}_\mathrm{intra}|=LS(S+1)/2=\mathcal{O}(LS^2)$, self-loops included,
and $|\mathcal{E}_\mathrm{inter}|=SL(L-1)/2=\mathcal{O}(SL^2)$ edges, for a total of $|\mathcal{E}|=\mathcal{O}(LS^2+SL^2)$.
Since $L \sim \mathcal{U}_{\mathbb{Z}}(1,12)$, we can derive best-, worst- and expected-case message passing costs underlying our graph encoding.
With one lead (best-case), $|\mathcal{E}|=\mathcal{O}(S^2)$; with 12 leads (worst-case), $|\mathcal{E}|=\mathcal{O}(12S^2+144S)$.
Under uniform lead sampling, $\mathbb{E}[|\mathcal{E}|]=\mathcal{O}(S^2\,\mathbb{E}[L]+S\,\mathbb{E}[L^2])$, so lead sampling reduces
the expected number of intra- and inter-lead edges by factors $\mathbb{E}[L]/12$ and $\mathbb{E}[L(L-1)]/132$, respectively.

Operating on a complete $LS$-node graph, which characterizes Transformer-based models with self-attention, per-layer computation costs $\mathcal{O}(L^2S^2)$, against our $|\mathcal{E}|=\mathcal{O}(LS^2+SL^2)$ (up to feature dimension $F$).
The per-layer ratio of the two costs is thus $\mathcal{O}(1/L+1/S)$, meaning that operating on our topology is asymptotically cheaper than on a complete graph whenever $L,S>1$.

\subsubsection{Implementation details}
\label{sec:methods-implementation}

\relay operates on 5-s windows at 100\,Hz ($T=500$); its modules are summarised in Table~\ref{tab:relay_architecture}.

\paragraph{Per-sampled-lead node count}
The kernel widths and strides of the four-layer convolutional embedder $\psi$ (Table~\ref{tab:relay_architecture}) are chosen on electrophysiological grounds, and they set the number of nodes contributed by each sampled lead.
The cumulative stride, which sets how often the embedder $\psi$ emits a node, is $3\times2\times2\times2=24$ samples (240\,ms); with boundary effects, it reduces the 500 samples of a lead to exactly $S=20$ nodes of dimension $F=768$, so a graph built from $L$ sampled leads has $20 \times L$ nodes.
The receptive field of each node is wider than this spacing: propagated layer by layer, it spans 40 samples (about 400\,ms), so adjacent nodes overlap by about 160\,ms and no P wave or QRS complex falls between nodes.
Three choices preserve wave-scale morphology within each node:
(i) the first convolution has a 10-sample (100\,ms) kernel, matching the duration of P waves and QRS complexes, and slides in 30\,ms steps, capturing fine-grained morphology before any reduction;
(ii) temporal compression is hierarchical, through four small-kernel layers rather than a single coarse pooling operation;
and (iii) each node is a 768-dimensional vector, as in BERT~\cite{Devlin2019}, wav2vec~2.0~\cite{Baevski2020} and HuBERT~\cite{Hsu2021}, which gives enough capacity to encode local shape, amplitude and timing (Supplementary Table~\ref{tab:supp_layers_channels}).
$S=20$ therefore reflects the temporal resolution of the embedder, while the receptive field and the dimension of each node keep it morphologically informative.

\paragraph{Encoder, codebooks and optimisation}
The encoder $\varphi$ is a two-layer GATv2~\cite{brody2021attentive} with 768 channels (graph encoder and topology in Supplementary Table~\ref{tab:supp_topology}; depth and width in Supplementary Table~\ref{tab:supp_layers_channels}).
The codebooks are fitted with mini-batch $k$-means, with $C=50$ classes in stage 1 (Supplementary Fig.~\ref{fig:supp_sse} and Table~\ref{tab:supp_codebook_stage1}) and $C'=500$ in stage 2 (Supplementary Table~\ref{tab:supp_codebook_pdrop}).
The stage-2 targets are obtained from the node representations of the first graph-attention layer, which retains more variance across nodes than the second (Supplementary Fig.~\ref{fig:supp_oversmoothing_stage1}).
The empirically optimum masking ratio is found for $p_\mathrm{MASK}=0.4$ (8 of the 20 nodes of each lead) (Supplementary Fig.~\ref{fig:supp_mask_ratio}), and the edge-dropping probability in stage 2 for $p_\mathrm{drop}=0.2$ (Supplementary Fig.~\ref{fig:supp_pdrop} and Table~\ref{tab:supp_codebook_pdrop}).
Both stages use the Adam optimiser ($\beta_1=0.9$, $\beta_2=0.98$) with a learning rate of $10^{-3}$ and a weight decay of $10^{-3}$. Stage-1 and stage-2 checkpoints selected by transfer-guided early stopping are those at 80k and 75k steps (Supplementary Figs.~\ref{fig:supp_oversmoothing_stage1} and~\ref{fig:supp_oversmoothing_stage2}).
These choices are made on the held-out Ribeiro-dev by linear evaluation after 60k pretraining steps (Supplementary Section~\ref{sec:supp_ablations}).

\begin{table}
\centering
\caption{
\textbf{\relay architecture.}
$C$ and $C'$ are the codebook sizes of pretraining stages 1 and 2.}
\label{tab:relay_architecture}
\footnotesize
\begin{tabular}{@{}ll@{}}
\toprule
Module & Specification \\
\midrule
\multirow{5}{*}{1D convolutional embedder $\psi$}
 & strides: 3, 2, 2, 2 \\
 & kernel widths: 10, 3, 3, 2 \\
 & channels: 768 \\
 & activation: GELU \\
 & normalisation: layer normalisation \\
\midrule
\multirow{8}{*}{Graph attention encoder $\varphi$}
 & attention: GATv2 \\
 & layers: 2 \\
 & hidden channels: 768 \\
 & attention heads: 8 \\
 & activation: GELU \\
 & normalisation: layer normalisation \\
 & dropout: 0.1 \\
 & skip connections: yes \\
\midrule
\multirow{2}{*}{Pretraining head}
 & input dimension: 768 \\
 & output dimension: $C$ or $C'$ \\
\midrule
\multirow{3}{*}{Downstream head}
 & readout: mean pooling \\
 & input dimension: 768 \\
 & output dimension: number of targets \\
\midrule
Total trainable parameters & 7.1 million \\
\bottomrule
\end{tabular}
\end{table}

\section*{Data availability}
\label{sec:data-availability}

\paragraph{Data}
This study is a secondary analysis of existing datasets, and no new data was generated. The values underlying every figure and table, for every model, benchmark and input configuration, are given in the main and Supplementary Tables.
All downstream benchmarks are built from public datasets.
Their sources and licences are listed below, and the datasets of the benchmark of~\cite{almasud2026benchmarkingecgfmsreality} are obtained and preprocessed following the instructions of its repository (\url{https://github.com/AI4HealthUOL/ecg-fm-benchmarking}).
For MIMIC-IV-ECG, the diagnostic labels (MIMIC-IV-ECG-ICD), the acute-care labels (MIMIC-IV-Ext-MDS-ED), the machine measurements of MIMIC-IV-ECG and the biometric, vital-sign and laboratory tables of MIMIC-IV are all obtained from PhysioNet.
The unlabelled part of CODE and CODE 85\%, used only for pretraining, are not public and are available on request, subject to a data use agreement; they are used because no public corpus reaches this scale and geographic diversity.

\paragraph{Model checkpoints}
The checkpoints of the seven baselines are the public ones released by their authors and listed below with their licences.

\paragraph{Sources and licences of the datasets}
\begin{itemize}
\item \textbf{CODE, unlabelled and CODE 85\%}: On request, subject to a data use agreement; Data use agreement.
\item \textbf{CODE-15\%}: \url{https://zenodo.org/records/4916206}; CC BY 4.0.
\item \textbf{IKEM}: \url{https://zenodo.org/records/8393007}; CC BY 4.0.
\item \textbf{PTB-XL}: \url{https://physionet.org/content/ptb-xl/1.0.3/}; CC BY 4.0.
\item \textbf{CPSC2018, CPSC-Extra, Georgia, Ningbo}: \url{https://physionet.org/content/challenge-2021/1.0.3/}; CC BY 4.0.
\item \textbf{Chapman}: \url{https://figshare.com/collections/ChapmanECG/4560497}; CC BY 4.0.
\item \textbf{Hefei}: \url{https://tianchi.aliyun.com/competition/entrance/231754/information}; CC BY-NC-SA 4.0.
\item \textbf{PTB}: \url{https://www.physionet.org/content/ptbdb/1.0.0/}; ODC-By 1.0.
\item \textbf{SPH}: \url{https://springernature.figshare.com/collections/A_large-scale_multi-label_12-lead_electrocardiogram_database_with_standardized_diagnostic_statements/5779802/1}; CC BY 4.0.
\item \textbf{ZZU pECG}: \url{https://doi.org/10.6084/m9.figshare.27078763}; CC BY 4.0.
\item \textbf{EchoNext}: \url{https://physionet.org/content/echonext/1.1.0/}; PhysioNet Restricted Health Data License 1.5.0.
\item \textbf{MIMIC-IV-ECG}: \url{https://physionet.org/content/mimic-iv-ecg/1.0/}; ODbL 1.0.
\item \textbf{CinC 2017}: \url{https://physionet.org/content/challenge-2017/1.0.0/}; ODC-By 1.0.
\item \textbf{Icentia11k}: \url{https://physionet.org/content/icentia11k-continuous-ecg/1.0/}; CC BY-NC-SA 4.0.
\end{itemize}

\paragraph{Sources and licences of the checkpoints}
\begin{itemize}
\item \textbf{ECGFounder}: \url{https://huggingface.co/PKUDigitalHealth/ECGFounder/tree/main}; MIT.
\item \textbf{ECG-JEPA}: \url{https://drive.google.com/file/d/1gMOT4xjQQg0GZkY1iE6NuDzua4ALw00l/view}; MIT.
\item \textbf{ST-MEM}: \url{https://drive.google.com/file/d/14nScwPk35sFi8wc-cuLJLqudVwynKS0n/view}; CC BY 4.0.
\item \textbf{MERL}: \url{https://drive.google.com/drive/folders/13wb4DppUciMn-Y_qC2JRWTbZdz3xX0w2}; MIT.
\item \textbf{ECGFM-KED}: \url{https://zenodo.org/records/14881564}; CC BY 4.0.
\item \textbf{HuBERT-ECG}: \url{https://huggingface.co/Edoardo-BS/hubert-ecg-base/tree/main}; CC BY-NC 4.0.
\item \textbf{ECG-CPC}: \url{https://figshare.com/articles/dataset/ECG-CPC_Checkpoint_zip/30192604?file=58173919}; CC BY 4.0.
\end{itemize}
\clearpage

\section*{Code availability}
The code of this study, to pretrain, finetune and evaluate \relay and to reproduce the evaluation of all models, including the lead subsets, seeds and statistical tests, is available at https://github.com/Edoar-do/RELAY/tree/main and archived on Zenodo (10.5281/zenodo.24285394), under the CC BY-NC 4.0 licence.

\bibliography{sn-bibliography}

\section*{Acknowledgements}
We thank the authors of the CODE study for giving us access to their dataset.

\section*{Funding}
This work has been partly funded by 1) Regione Lombardia, under the
``Programme of measures for economic recovery: development of new cooperation agreements
with universities for research, innovation and tech transfer'' DGR n. XI/4445/2021 and 2)
European Union’s Horizon Europe research and innovation programme under Grant Agreement No 101156266.
\section*{Author contributions}
E.C. and S.F. contributed equally to this work.
E.C.: conceptualization, methodology, software, validation, formal analysis, investigation, data curation, writing -- original draft, writing -- review and editing, visualization, supervision and project administration.
S.F.: conceptualization, methodology, software, validation, formal analysis, investigation, data curation, writing -- original draft, writing -- review and editing, visualization and supervision.
P.L.: conceptualization, funding and writing -- review and editing.
M.S.: visualization, resources and writing -- review and editing.
A.S.: resources and funding acquisition.
E.C. and S.F. jointly supervised this work.

\section*{Competing interests}
The authors declare no competing interests.

\clearpage

\clearpage
\setcounter{page}{1}%
\begin{center}
{\LARGE\bfseries \papertitle\par}\medskip
{\Large Supplementary Material}
\end{center}
\bigskip

\appendixtitleoff%
\begin{appendices}
\renewcommand{\appendixname}{Supplementary Section}
\renewcommand{\thesection}{S\arabic{section}}
\renewcommand{\tablename}{Supplementary Table}
\renewcommand{\thetable}{\arabic{table}}
\renewcommand{\figurename}{Supplementary Fig.}
\renewcommand{\thefigure}{\arabic{figure}}
\setcounter{figure}{0}

\section{Computational resources and experiment duration}\label{sec:supp_compute}

All experiments run on an NVIDIA DGX A100 system (eight 80\,GB GPUs; two 64-core AMD EPYC 7742 CPUs; 2\,TB of RAM) with Ubuntu 26.04, Python 3.13, CUDA 12.6 and PyTorch 2.7.1; training is monitored with Weights \& Biases, which is also used for hyperparameter tuning.
The computational cost of every model at inference is reported in Table~\ref{tab:compute}.

Pretraining \relay involves six steps:
(1) extraction of the feature descriptors of the ECG segments, whose cost grows linearly with the number and length of the pretraining ECGs;
(2) fitting of the stage-1 codebook by clustering these descriptors, whose cost is that of the clustering algorithm (mini-batch $k$-means);
(3) stage-1 masked node modelling;
(4) extraction of the latent node representations;
(5) fitting of the stage-2 codebook by clustering these representations;
and (6) stage-2 masked node modelling.
Step (1) takes about one month, step (2) about one hour, step (3) about three hours to converge, step (4) about one month, step (5) about 4.5 hours and step (6) about three hours to converge.
Feature extraction and codebook fitting are vectorisable and performed once, and, like the two stages of masked node modelling, never need to be repeated to use \relay on new data, which requires only finetuning.
Finetuning time, once fixed the training budget, depends on the size of the downstream dataset; on CODE-15\%, the largest, it takes about 30 minutes.

\section{Additional ablations}\label{sec:supp_ablations}

All ablations are run on Ribeiro-dev, a held-out development set that is a subset of CODE introduced in~\cite{s-Coppola2024} and disjoint from all evaluation benchmarks, so that design choices are not tuned on the benchmarks.
It is labelled for first-degree atrioventricular block, right bundle branch block, left bundle branch block, sinus bradycardia, atrial fibrillation and sinus tachycardia.
Unless stated otherwise, each variant is pretrained for 60k steps with the optimiser of the final model (Section~\ref{sec:methods-implementation}) and assessed by linear evaluation on frozen representations, and performance is the macro AUROC (\%) on the Ribeiro-dev internal validation partition.
Variants differ only in the factor under study.
Ablations already reported in the main text (Fig.~\ref{fig:ablations}) are not repeated here, except for the values of the zero-shot transfer of Fig.~\ref{fig:ablations}b, which are given in Supplementary Table~\ref{tab:supp_abl_transfer}.

\begin{table}[h]
\centering
\caption{\textbf{Zero-shot transfer of the three pretrained variants (Fig.~\ref{fig:ablations}b).} Macro AUROC (\%) on Ribeiro-dev of one linear layer trained on frozen 12-lead representations and tested without retraining, with the absent leads omitted or filled with zeros at inference. Variant (1): random lead sampling (RLS), non-sampled leads omitted during pretraining (\relay); (2): RLS, non-sampled leads filled with zeros; (3): no RLS. Random-lead values are means over five seeds (standard deviations at most 0.33 AUROC points).}
\label{tab:supp_abl_transfer}
\begin{tabular}{llccccc}
\toprule
Variant & Absent leads & 12 leads & One random & Two random & Lead I & Lead II \\
\midrule
\multirow{2}{*}{(1) RLS, omitted (\relay)} & Omitted     & \multirow{2}{*}{96.57} & 84.94 & 89.96 & 87.44 & 84.07 \\
                                           & Zero-filled &                        & 78.74 & 86.35 & 82.50 & 81.17 \\
\midrule
\multirow{2}{*}{(2) RLS, zero-filled}      & Omitted     & \multirow{2}{*}{96.41} & 84.34 & 89.36 & 86.95 & 82.87 \\
                                           & Zero-filled &                        & 78.52 & 84.23 & 81.62 & 78.49 \\
\midrule
\multirow{2}{*}{(3) No RLS}                & Omitted     & \multirow{2}{*}{96.51} & 82.09 & 88.17 & 84.02 & 80.15 \\
                                           & Zero-filled &                        & 72.19 & 79.92 & 76.45 & 71.66 \\
\bottomrule
\end{tabular}
\end{table}

\subsection{Graph encoders and topologies}\label{sec:supp_abl_topology}

When designing a graph topology not known a priori, one should compare different topologies to find the best one for the specific task at hand.
However, the set of possible topologies, and size of the space to search, grows exponentially with the number of possible edges.
For this reason, standard practice is to compare against the fully connected topology,
in which every node is connected to every other node to allow for maximal information exchange,
and with no graph, in which a multilayer perceptron processees each node independently to measure the
information content of node features alone. Additional exploration axes depend on the model processing such topologies and its ability to learn edge weigths or treat them statically.
Supplementary Table~\ref{tab:supp_topology} covers all these axes under a fixed budget
(60k pretraining steps, 4 million parameters) and compares the spatiotemporal topology
of \relay (Section~\ref{sec:methods-architecture}) with a fully connected graph and no graph,
for several graph encoders treating edge weights as both learnable and static values.
Using a topology is always beneficial, and our spatiotemporal topology is consistently better
than fully connectivity, peaking at 93.7\% with a GAT encoder.

\begin{table}[h]
\centering
\caption{\textbf{Graph encoders and topologies.} Macro AUROC (\%) on Ribeiro-dev by linear evaluation, after 60k pretraining steps with 4 million parameters. GAT, graph attention network~\cite{s-velickovic2017graph}; GIN, graph isomorphism network~\cite{s-xu2018powerful}; GT, graph transformer~\cite{s-shi2020masked}; SNN, inductive sheaf neural network~\cite{s-bodnar2022neural,s-fiorini2026benchmarking}; SAN v1 and SAN v2, inductive sheaf attention networks~\cite{s-barbero2022sheaf,s-fiorini2026benchmarking}; MLP, multilayer perceptron applied to each node without a graph. Best value in bold.}
\label{tab:supp_topology}
\begin{tabular}{lccc}
\toprule
Encoder & Spatiotemporal & Fully connected & No graph \\
\midrule
GAT     & \textbf{93.7}  & 85.4            & --       \\
GIN     & 90.7           & 85.3            & --       \\
GT      & 90.0           & 82.0            & --       \\
SNN     & 80.8           & 79.9            & --       \\
SAN v1  & 81.1           & 75.2            & --       \\
SAN v2  & 82.2           & 70.0            & --       \\
MLP     & --             & --              & 74.7     \\
\bottomrule
\end{tabular}
\end{table}

\subsection{GAT layers and channels}\label{sec:supp_abl_layers}

Supplementary Table~\ref{tab:supp_layers_channels} reports the linear evaluation results of GAT encoders with two to four layers and 256 to 768 channels.
A two-layer encoder with 768 channels has the highest AUROC (95.1\%).
Deeper encoders have lower AUROCs at every width, consistent with the onset of oversmoothing in the small-diameter subgraphs of each lead~\cite{s-chen2020measuring,s-zhao2020pairnorm}.
The resulting architecture is reported in Table~\ref{tab:relay_architecture}.

\begin{table}[h]
\centering
\caption{\textbf{GAT layers and channels.} Macro AUROC (\%) on Ribeiro-dev by linear evaluation, after 60k pretraining steps. Best value in bold.}
\label{tab:supp_layers_channels}
\begin{tabular}{lccc}
\toprule
 & \multicolumn{3}{c}{Layers} \\
\cmidrule{2-4}
Channels & 2             & 3    & 4    \\
\midrule
256      & 92.4          & 92.8 & 92.3 \\
512      & 93.4          & 93.7 & 92.9 \\
768      & \textbf{95.1} & 93.6 & 93.0 \\
\bottomrule
\end{tabular}
\end{table}

\subsection{Masking ratio}\label{sec:supp_abl_mask}

\begin{figure}
\centering
\includegraphics[width=0.7\textwidth]{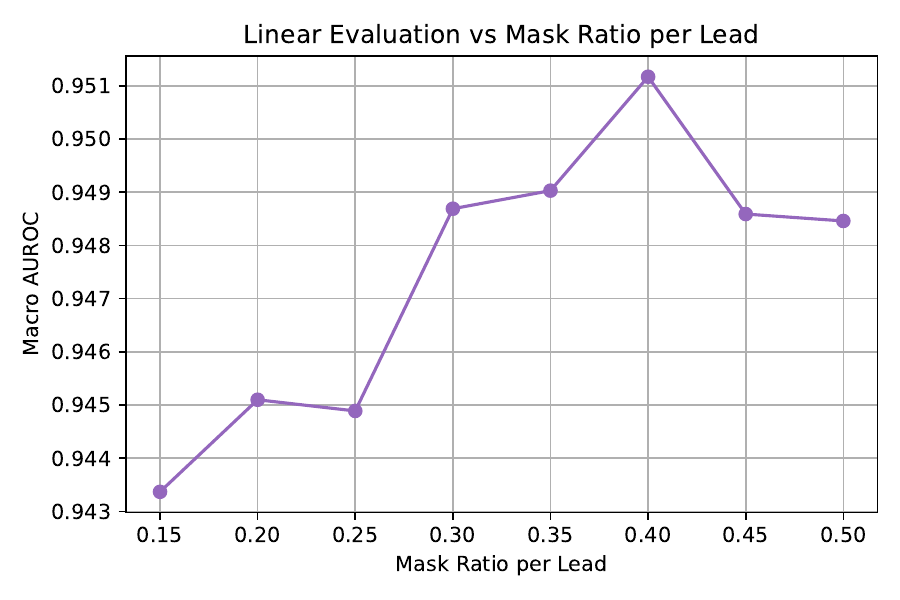}
\caption{\textbf{Masking ratio.} Macro AUROC on Ribeiro-dev by linear evaluation for different per-lead masking ratios $p_\mathrm{MASK}$, after 60k pretraining steps.}
\label{fig:supp_mask_ratio}
\end{figure}

The masking ratio sets the difficulty of masked node modelling.
$p_\mathrm{MASK}$ is varied over $\{0.15, 0.20, 0.25, 0.30, 0.35, 0.40, 0.45, 0.50\}$, with a two-layer, 768-channel GAT encoder and a stage-1 codebook of $C=50$ classes (Supplementary Fig.~\ref{fig:supp_mask_ratio}).
The AUROC is highest at $p_\mathrm{MASK}=0.40$ (95.1\%).
With 20 nodes per lead, this ratio masks 8 nodes of each lead, so no lead is ever masked entirely and every lead contributes both masked targets and visible context.

\subsection{Stage-1 pretraining optimisation}\label{sec:supp_abl_stage1}

\subsubsection{Codebook size and multi-task learning}\label{sec:supp_abl_codebook1}

\begin{figure}
\centering
\includegraphics[width=\textwidth]{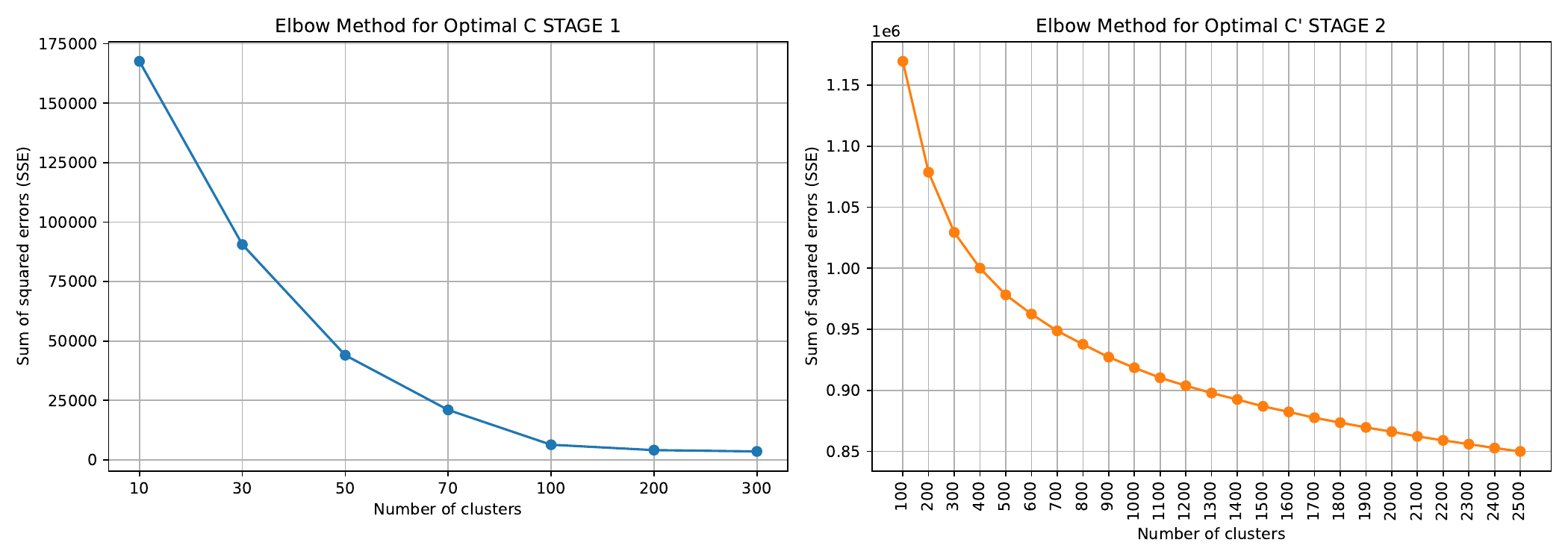}
\caption{\textbf{Sum of squared errors of the codebooks.} Left: clustering of the mel-frequency cepstral coefficient descriptors, before stage 1. Right: clustering of the node representations extracted from the first graph-attention layer of the stage-1 network, before stage 2.}
\label{fig:supp_sse}
\end{figure}

Stage-1 codebooks are fitted on mel-frequency cepstral coefficient descriptors with $C\in\{10, 30, 50, 70, 100, 200, 300\}$ classes.
The elbow of the sum of squared errors (Supplementary Fig.~\ref{fig:supp_sse}, left) restricts the search to $C\in\{50, 70, 100\}$.
\relay is then pretrained with $p_\mathrm{MASK}=0.40$ against each of these codebooks and against all their combinations, predicting the classes of several codebooks at once (multi-task learning), as in HuBERT~\cite{s-Hsu2021} and HuBERT-ECG~\cite{s-Coppola2024} (Supplementary Table~\ref{tab:supp_codebook_stage1}).
$C=50$, together with $C = [50, 100]$, reach the highest AUROC (95.1\%), and we choose the first its overall smaller size and lower computational cost.

\begin{table}
\centering
\caption{\textbf{Stage-1 codebook size and multi-task learning.} Macro AUROC (\%) on Ribeiro-dev by linear evaluation, after 60k pretraining steps. Columns with several sizes predict the classes of all the listed codebooks at once. Best value in bold.}
\label{tab:supp_codebook_stage1}
\begin{tabular}{lccccccc}
\toprule
Codebook size(s) $C$ & 50            & 70   & 100  & 50, 70 & 50, 100 & 70, 100 & 50, 70, 100 \\
\midrule
AUROC                & \textbf{95.1} & 94.3 & 94.8 & 94.6   & 95.1    & 94.9    & 94.7        \\
\bottomrule
\end{tabular}
\end{table}

\subsubsection{Training dynamics and checkpoint selection}\label{sec:supp_abl_ckpt1}

\begin{figure}[h]
\centering
\includegraphics[width=\textwidth]{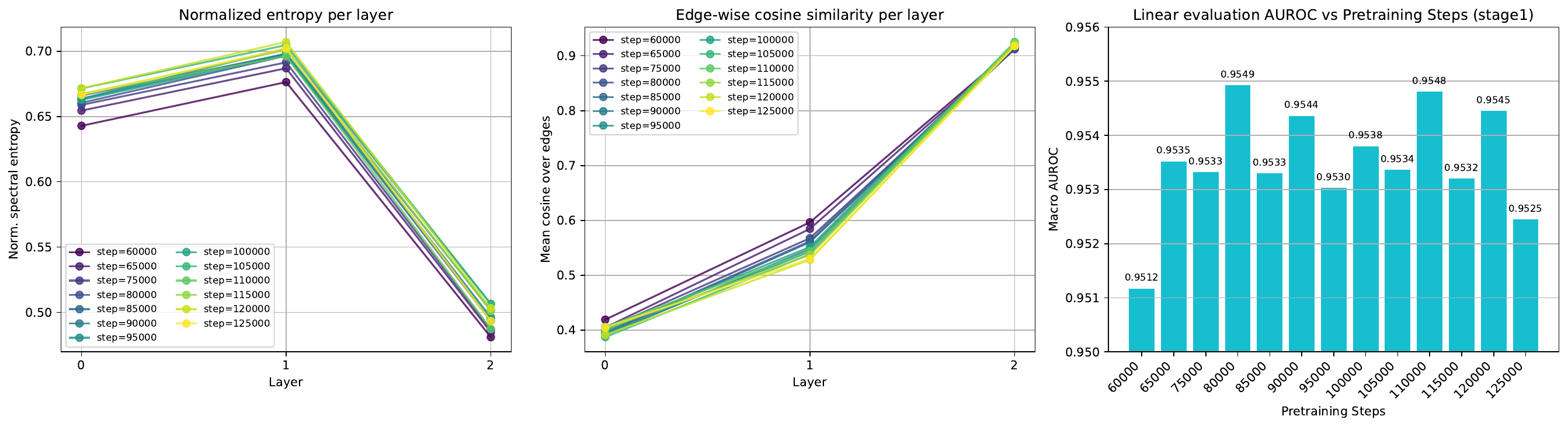}
\caption{\textbf{Stage-1 training dynamics.} Left: normalised spectral entropy of the node features of each graph-attention layer across stage-1 checkpoints. Centre: cosine similarity between the features of connected nodes of each layer across stage-1 checkpoints. Right: macro AUROC on Ribeiro-dev by linear evaluation of the stage-1 checkpoints.}
\label{fig:supp_oversmoothing_stage1}
\end{figure}

Stage-1 checkpoints are monitored with linear evaluation and with two unsupervised indicators of representation collapse in each graph-attention layer: the normalised spectral entropy of the node features, which measures their diversity, and the cosine similarity between the features of connected nodes, which measures oversmoothing~\cite{s-chen2020measuring,s-zhao2020pairnorm} (Supplementary Fig.~\ref{fig:supp_oversmoothing_stage1}).
The AUROC is highest at about 80k steps (95.5\%) and decreases with further training.
Over the same steps, the entropy decreases and the cosine similarity increases in the second layer, indicating the onset of oversmoothing.
The 80k-step checkpoint is therefore used to extract the node representations from which the stage-2 codebook is fitted.

\subsection{Stage 2: codebook refinement and structural regularisation}\label{sec:supp_abl_stage2}

In stage 2, the targets are no longer clusters of mel-frequency cepstral coefficient descriptors but clusters of the latent node representations learnt in stage 1, so that the targets are refined on learnt features rather than on fixed signal descriptors.

\subsubsection{Extraction layer selection and refined clustering}\label{sec:supp_abl_layer}

The analysis of Supplementary Fig.~\ref{fig:supp_oversmoothing_stage1} also selects in advance the layer from which the node representations are extracted without any extraction, clustering or training.
The first graph-attention layer keeps a higher entropy and a lower cosine similarity than the second, so its node representations are used.
The sum of squared errors of their clustering decreases more gradually than that of the stage-1 descriptors (Supplementary Fig.~\ref{fig:supp_sse}, right), and its rate of change restricts the candidate stage-2 codebook sizes to $C'\in\{500, 1000, 1500\}$.

\subsubsection{Edge dropout and codebook interaction}\label{sec:supp_abl_codebook2}

\begin{figure}[h]
\centering
\includegraphics[width=0.8\textwidth]{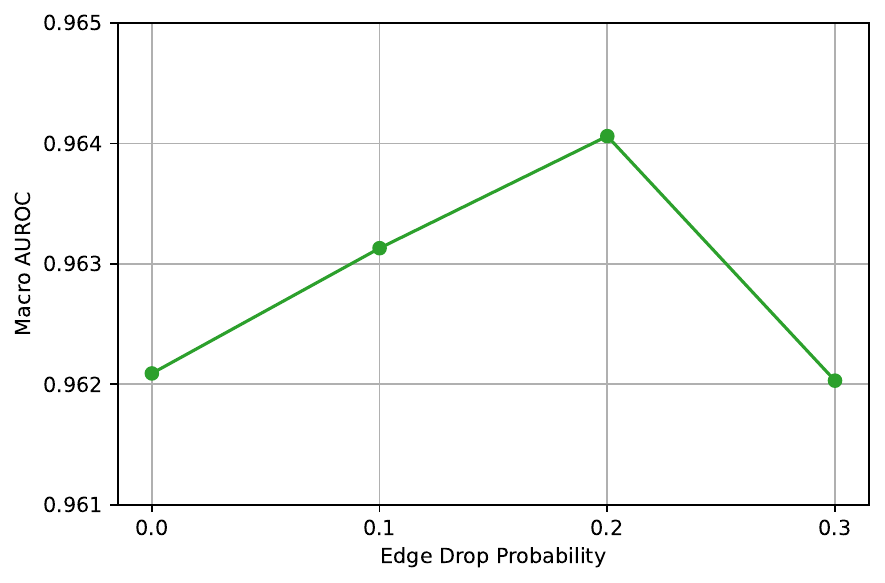}
\caption{\textbf{Edge-dropping probability.} Macro AUROC on Ribeiro-dev by linear evaluation after stage-2 pretraining with different edge-dropping probabilities $p_\mathrm{drop}$.}
\label{fig:supp_pdrop}
\end{figure}

In stage 2, intra-lead edges are dropped at random with probability $p_\mathrm{drop}$, which perturbs the topology during training and counters oversmoothing~\cite{s-rong2019dropedge}.
A grid search over $C'\in\{500, 1000, 1500\}$ and $p_\mathrm{drop}\in\{0.0, 0.1, 0.2\}$ (Supplementary Table~\ref{tab:supp_codebook_pdrop}) gives the highest AUROC with $C'=500$ and $p_\mathrm{drop}=0.2$ (96.4\%).
With $C'=500$, increasing $p_\mathrm{drop}$ to 0.3 decreases the AUROC to 96.2\% (Supplementary Fig.~\ref{fig:supp_pdrop}), so we keep the $C'=500$ and $p_\mathrm{drop}=0.2$ configuration for the final \relay model.

\begin{table}[h]
\centering
\caption{\textbf{Stage-2 codebook size and edge dropout.} Macro AUROC (\%) on Ribeiro-dev by linear evaluation for each combination of stage-2 codebook size $C'$ and edge-dropping probability $p_\mathrm{drop}$. Best value in bold.}
\label{tab:supp_codebook_pdrop}
\begin{tabular}{lccc}
\toprule
                      & $C'=500$      & $C'=1000$ & $C'=1500$ \\
\midrule
$p_\mathrm{drop}=0.0$ & 96.2          & 96.2      & 96.3      \\
$p_\mathrm{drop}=0.1$ & 96.3          & 96.1      & 96.1      \\
$p_\mathrm{drop}=0.2$ & \textbf{96.4} & 96.3      & 96.2      \\
\bottomrule
\end{tabular}
\end{table}

\subsubsection{Stage-2 dynamics analysis}\label{sec:supp_abl_ckpt2}

\begin{figure}[h]
\centering
\includegraphics[width=\textwidth]{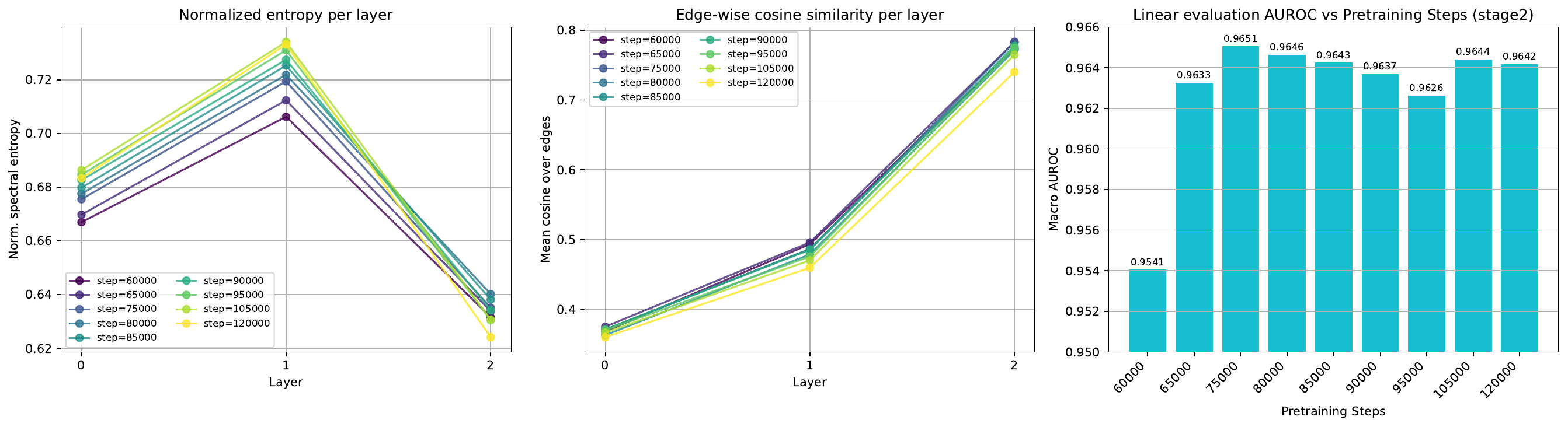}
\caption{\textbf{Stage-2 training dynamics.} Left: normalised spectral entropy of the node features of each graph-attention layer across stage-2 checkpoints. Centre: cosine similarity between the features of connected nodes of each layer across stage-2 checkpoints. Right: macro AUROC on Ribeiro-dev by linear evaluation of the stage-2 checkpoints.}
\label{fig:supp_oversmoothing_stage2}
\end{figure}

The stage-2 checkpoints are analysed in the same way (Supplementary Fig.~\ref{fig:supp_oversmoothing_stage2}).
The AUROC is highest earlier than in stage 1, at 75k steps (96.5\%), after which the entropy of the second layer drops and its cosine similarity rises.
The 75k-step checkpoint is therefore retained as the final \relay model.

\clearpage

\section{Additional tables}\label{sec:supp_tables}

\begin{table*}[!ht]
\centering
\caption{\textbf{Performance of every model on every benchmark with 12 leads.} Macro AUROC (\%) for the 24 classification benchmarks and mean absolute error (MAE, lower is better) for the five MIMIC-IV-ECG regression benchmarks, with 95\% CIs (bootstrap of the test set, $n=1000$). The best value of each row is in bold; *, not significantly worse than the best model of the row (paired bootstrap of the difference, 95\% CI includes 0). $^\dagger$Evaluation-only benchmark.}
\label{tab:supp_all12}
\setlength{\tabcolsep}{3pt}
\renewcommand{\arraystretch}{0.95}
\resizebox{\textwidth}{!}{%
\begin{tabular}{@{}lcccccccc@{}}
\toprule
Benchmark                     & ECGFounder                               & ECG-JEPA                                    & ST-MEM                             & MERL                                     & ECGFM-KED                          & HuBERT-ECG                               & ECG-CPC                                     & \textbf{RELAY}                           \\
\midrule
CODE 15\%                     & 99.0 {\scriptsize (98.5--99.4)}          & 99.2 {\scriptsize (98.8--99.5)}*            & 99.3 {\scriptsize (98.9--99.5)}*   & 98.7 {\scriptsize (98.1--99.1)}          & 98.5 {\scriptsize (97.7--99.2)}    & 98.9 {\scriptsize (97.8--99.5)}          & 99.2 {\scriptsize (99.0--99.5)}*            & \textbf{99.4} {\scriptsize (98.7--99.6)} \\
PTB-XL All                    & 93.4 {\scriptsize (92.5--94.3)}          & 93.9 {\scriptsize (93.0--94.6)}             & 90.8 {\scriptsize (89.6--91.9)}    & 92.5 {\scriptsize (91.4--93.7)}          & 89.9 {\scriptsize (89.4--91.3)}    & 91.1 {\scriptsize (89.8--92.1)}          & \textbf{94.6} {\scriptsize (93.9--95.2)}    & 91.5 {\scriptsize (89.7--92.1)}          \\
PTB-XL Diag$^\dagger$         & \textbf{94.9} {\scriptsize (94.0--95.8)} & 94.5 {\scriptsize (93.6--95.1)}             & 90.4 {\scriptsize (88.7--91.9)}    & 93.9 {\scriptsize (92.8--95.2)}          & 93.1 {\scriptsize (90.5--93.1)}    & 92.3 {\scriptsize (91.0--93.4)}          & 94.8 {\scriptsize (93.9--95.6)}             & 94.2 {\scriptsize (91.6--95.9)}          \\
PTB-XL Form$^\dagger$         & 87.3 {\scriptsize (85.5--90.7)}          & 89.8 {\scriptsize (87.8--92.9)}             & 88.2 {\scriptsize (86.4--90.8)}    & 84.6 {\scriptsize (82.3--86.2)}          & 78.0 {\scriptsize (76.6--82.5)}    & 83.6 {\scriptsize (80.9--87.1)}          & \textbf{92.3} {\scriptsize (90.8--94.1)}    & 86.0 {\scriptsize (85.0--89.7)}          \\
PTB-XL Rhythm$^\dagger$       & 96.3 {\scriptsize (95.4--97.1)}          & \textbf{97.3} {\scriptsize (96.4--97.9)}    & 95.1 {\scriptsize (92.8--96.6)}    & 91.2 {\scriptsize (87.5--94.4)}          & 88.4 {\scriptsize (85.4--91.1)}    & 95.9 {\scriptsize (94.5--96.9)}          & 96.0 {\scriptsize (94.9--96.8)}             & 95.6 {\scriptsize (94.3--96.7)}          \\
PTB-XL Sub                    & \textbf{94.1} {\scriptsize (92.9--95.0)} & 93.5 {\scriptsize (93.0--94.7)}*            & 91.5 {\scriptsize (90.3--92.4)}    & 93.4 {\scriptsize (92.4--94.6)}*         & 89.8 {\scriptsize (87.9--91.5)}    & 91.8 {\scriptsize (90.8--92.9)}          & 93.3 {\scriptsize (92.9--95.0)}             & 93.5 {\scriptsize (92.4--93.9)}*         \\
PTB-XL Super                  & \textbf{93.5} {\scriptsize (92.8--94.1)} & 92.0 {\scriptsize (91.6--93.1)}             & 89.7 {\scriptsize (88.7--90.4)}    & 92.8 {\scriptsize (92.3--93.6)}          & 90.4 {\scriptsize (89.8--91.4)}    & 90.6 {\scriptsize (89.9--91.6)}          & 93.1 {\scriptsize (92.7--94.0)}             & 92.3 {\scriptsize (90.8--92.6)}          \\
EchoNext                      & 81.1 {\scriptsize (80.3--82.5)}          & 82.1 {\scriptsize (80.9--83.1)}             & 81.3 {\scriptsize (80.2--82.7)}    & 81.6 {\scriptsize (80.8--83.1)}          & 80.7 {\scriptsize (79.7--82.2)}    & 79.1 {\scriptsize (77.8--80.5)}          & \textbf{83.0} {\scriptsize (82.1--84.4)}    & 79.7 {\scriptsize (77.7--80.5)}          \\
CPSC2018                      & 96.6 {\scriptsize (95.9--97.3)}          & 96.8 {\scriptsize (95.4--97.3)}*            & 94.3 {\scriptsize (93.4--95.1)}    & 93.6 {\scriptsize (92.7--94.6)}          & 93.0 {\scriptsize (92.1--93.7)}    & \textbf{97.2} {\scriptsize (96.5--97.8)} & 96.8 {\scriptsize (96.1--97.6)}*            & 96.8 {\scriptsize (95.7--97.4)}*         \\
CPSC-Extra                    & 87.2 {\scriptsize (85.5--89.3)}          & 85.4 {\scriptsize (83.5--86.7)}             & 85.3 {\scriptsize (84.3--86.9)}    & 86.5 {\scriptsize (84.5--88.3)}          & 80.9 {\scriptsize (78.9--82.3)}    & \textbf{88.1} {\scriptsize (85.5--90.4)} & 84.9 {\scriptsize (81.8--86.8)}             & 85.8 {\scriptsize (82.7--89.4)}          \\
Chapman                       & \textbf{96.8} {\scriptsize (96.4--97.7)} & 96.7 {\scriptsize (96.0--97.6)}*            & 94.8 {\scriptsize (94.1--96.0)}    & 94.5 {\scriptsize (93.8--95.8)}          & 91.7 {\scriptsize (90.7--93.6)}    & 93.6 {\scriptsize (92.6--95.1)}          & 95.6 {\scriptsize (95.4--97.2)}             & 95.2 {\scriptsize (93.7--95.6)}          \\
Chapman Rhythm$^\dagger$      & 99.1 {\scriptsize (98.7--99.5)}*         & \textbf{99.2} {\scriptsize (98.5--99.5)}    & 98.5 {\scriptsize (97.9--98.9)}    & 97.4 {\scriptsize (96.3--98.2)}          & 96.3 {\scriptsize (95.4--97.1)}    & 97.7 {\scriptsize (96.9--98.4)}          & 98.5 {\scriptsize (98.1--99.1)}             & 98.6 {\scriptsize (97.9--99.0)}          \\
Georgia                       & \textbf{92.0} {\scriptsize (90.4--94.2)} & 91.8 {\scriptsize (90.2--93.9)}*            & 88.7 {\scriptsize (86.2--90.8)}    & 91.0 {\scriptsize (89.9--92.6)}          & 85.6 {\scriptsize (84.2--87.2)}    & 86.3 {\scriptsize (84.9--88.7)}          & 91.2 {\scriptsize (89.6--92.1)}*            & 87.6 {\scriptsize (85.6--89.4)}          \\
Hefei                         & 94.8 {\scriptsize (92.8--96.3)}          & 94.5 {\scriptsize (92.3--95.8)}             & 95.3 {\scriptsize (93.3--96.7)}    & 96.8 {\scriptsize (95.2--97.7)}*         & 94.9 {\scriptsize (93.5--96.0)}    & 94.9 {\scriptsize (92.5--96.7)}          & \textbf{96.9} {\scriptsize (95.9--98.0)}    & 96.5 {\scriptsize (93.6--96.7)}          \\
Ningbo                        & \textbf{97.4} {\scriptsize (96.4--98.9)} & 97.3 {\scriptsize (96.0--98.6)}*            & 95.4 {\scriptsize (94.2--96.5)}    & 94.1 {\scriptsize (92.9--95.4)}          & 93.8 {\scriptsize (92.5--95.3)}    & 95.6 {\scriptsize (95.0--96.3)}          & 97.0 {\scriptsize (95.8--98.1)}             & 96.4 {\scriptsize (95.2--96.6)}          \\
PTB                           & 65.6 {\scriptsize (60.8--69.9)}*         & 67.9 {\scriptsize (64.7--77.2)}*            & 69.4 {\scriptsize (64.2--76.6)}*   & \textbf{71.7} {\scriptsize (69.3--81.1)} & 61.2 {\scriptsize (54.9--66.9)}    & 69.9 {\scriptsize (68.5--79.6)}          & 70.2 {\scriptsize (68.9--80.0)}             & 66.3 {\scriptsize (58.7--72.1)}*         \\
SPH                           & \textbf{98.2} {\scriptsize (97.8--98.5)} & 98.0 {\scriptsize (97.7--98.5)}*            & 96.4 {\scriptsize (96.1--97.2)}    & 94.4 {\scriptsize (93.9--96.7)}          & 93.1 {\scriptsize (92.0--95.2)}    & 95.7 {\scriptsize (94.7--97.4)}          & 97.8 {\scriptsize (97.7--98.3)}             & 96.3 {\scriptsize (95.2--97.3)}          \\
ZZU pECG                      & 89.7 {\scriptsize (88.9--93.5)}          & \textbf{90.8} {\scriptsize (89.8--92.7)}    & 89.3 {\scriptsize (88.7--92.1)}    & 88.4 {\scriptsize (87.8--91.8)}          & 86.8 {\scriptsize (85.9--89.3)}    & 87.2 {\scriptsize (86.7--89.1)}          & 89.1 {\scriptsize (88.6--92.8)}             & 88.7 {\scriptsize (88.2--92.0)}          \\
MIMIC Cardiac                 & 79.3 {\scriptsize (78.3--80.1)}          & 80.5 {\scriptsize (79.5--81.4)}*            & 77.3 {\scriptsize (76.3--78.2)}    & 80.2 {\scriptsize (79.3--81.1)}          & 79.6 {\scriptsize (78.5--80.5)}    & 77.9 {\scriptsize (76.8--78.9)}          & \textbf{80.8} {\scriptsize (79.9--81.7)}    & 78.3 {\scriptsize (77.3--79.3)}          \\
MIMIC Non-cardiac             & 71.4 {\scriptsize (70.6--72.1)}          & 72.9 {\scriptsize (72.0--73.6)}             & 69.7 {\scriptsize (69.1--70.3)}    & 72.5 {\scriptsize (71.8--73.2)}          & 71.9 {\scriptsize (71.0--72.6)}    & 70.3 {\scriptsize (69.5--71.1)}          & \textbf{73.4} {\scriptsize (72.7--74.1)}    & 70.5 {\scriptsize (69.7--71.2)}          \\
MIMIC Deterioration           & \textbf{79.4} {\scriptsize (75.7--82.5)} & 75.9 {\scriptsize (71.6--79.6)}             & 77.8 {\scriptsize (73.6--81.4)}*   & 78.3 {\scriptsize (74.1--81.8)}*         & 76.5 {\scriptsize (72.3--80.0)}    & 74.5 {\scriptsize (69.2--78.5)}          & 78.3 {\scriptsize (75.1--81.4)}*            & 76.3 {\scriptsize (72.7--80.0)}*         \\
MIMIC Mortality               & 82.8 {\scriptsize (80.1--84.9)}          & 83.2 {\scriptsize (80.5--85.2)}             & 82.5 {\scriptsize (79.8--84.6)}    & 83.3 {\scriptsize (80.7--85.4)}          & 82.9 {\scriptsize (80.4--84.9)}    & 82.2 {\scriptsize (79.2--84.4)}          & \textbf{84.8} {\scriptsize (81.9--86.7)}    & 82.9 {\scriptsize (80.3--84.9)}          \\
MIMIC ICU admission           & 75.3 {\scriptsize (73.4--77.1)}          & 75.4 {\scriptsize (73.4--77.4)}             & 74.2 {\scriptsize (72.2--76.1)}    & 75.9 {\scriptsize (74.0--77.7)}          & 75.2 {\scriptsize (73.1--77.1)}    & 73.8 {\scriptsize (71.9--75.7)}          & \textbf{76.7} {\scriptsize (74.7--78.6)}    & 74.8 {\scriptsize (73.0--76.6)}          \\
MIMIC Sex                     & 91.0 {\scriptsize (89.3--92.3)}          & 91.1 {\scriptsize (89.9--92.2)}             & 87.8 {\scriptsize (86.6--89.0)}    & 92.9 {\scriptsize (92.1--93.8)}*         & 91.1 {\scriptsize (90.1--92.0)}    & 88.3 {\scriptsize (86.9--89.7)}          & \textbf{93.4} {\scriptsize (92.2--94.4)}    & 89.0 {\scriptsize (87.0--90.5)}          \\
\midrule
MIMIC Age (MAE)               & 0.438 {\scriptsize (0.425--0.452)}       & 0.440 {\scriptsize (0.428--0.453)}          & 0.501 {\scriptsize (0.486--0.516)} & 0.440 {\scriptsize (0.427--0.454)}       & 0.453 {\scriptsize (0.439--0.467)} & 0.464 {\scriptsize (0.451--0.477)}       & \textbf{0.424} {\scriptsize (0.410--0.437)} & 0.480 {\scriptsize (0.466--0.493)}       \\
MIMIC Biometrics (MAE)        & 0.645 {\scriptsize (0.623--0.666)}       & 0.635 {\scriptsize (0.614--0.655)}          & 0.662 {\scriptsize (0.639--0.684)} & 0.627 {\scriptsize (0.606--0.649)}       & 0.646 {\scriptsize (0.626--0.666)} & 0.666 {\scriptsize (0.645--0.686)}       & \textbf{0.618} {\scriptsize (0.598--0.637)} & 0.669 {\scriptsize (0.645--0.691)}       \\
MIMIC ECG features (MAE)      & 0.465 {\scriptsize (0.453--0.478)}       & 0.473 {\scriptsize (0.460--0.485)}          & 0.497 {\scriptsize (0.484--0.509)} & 0.471 {\scriptsize (0.458--0.484)}       & 0.477 {\scriptsize (0.465--0.490)} & 0.482 {\scriptsize (0.469--0.494)}       & \textbf{0.460} {\scriptsize (0.447--0.472)} & 0.497 {\scriptsize (0.484--0.509)}       \\
MIMIC Laboratory values (MAE) & 0.638 {\scriptsize (0.619--0.659)}       & \textbf{0.609} {\scriptsize (0.592--0.650)} & 0.667 {\scriptsize (0.647--0.684)} & 0.636 {\scriptsize (0.593--0.679)}*      & 0.641 {\scriptsize (0.626--0.662)} & 0.656 {\scriptsize (0.632--0.680)}       & 0.626 {\scriptsize (0.587--0.664)}*         & 0.627 {\scriptsize (0.598--0.657)}*      \\
MIMIC Vital signs (MAE)       & 0.698 {\scriptsize (0.689--0.707)}       & 0.695 {\scriptsize (0.685--0.704)}          & 0.703 {\scriptsize (0.693--0.712)} & 0.695 {\scriptsize (0.686--0.704)}       & 0.696 {\scriptsize (0.687--0.705)} & 0.703 {\scriptsize (0.693--0.712)}       & \textbf{0.691} {\scriptsize (0.682--0.700)} & 0.700 {\scriptsize (0.690--0.709)}       \\
\midrule
Mean AUROC (24 benchmarks) & 89.0 & 89.2 & 87.6 & 88.3 & 86.0 & 87.4 & \textbf{89.7} & 88.0 \\
Mean MAE (5 benchmarks) & 0.577 & 0.570 & 0.606 & 0.574 & 0.583 & 0.594 & \textbf{0.564} & 0.594 \\
\bottomrule
\end{tabular}}
\end{table*}

\begin{table*}[p]
\centering
\caption{\textbf{Performance of every model on every benchmark with lead I.} Macro AUROC (\%) for the 24 classification benchmarks and mean absolute error (MAE, lower is better) for the five MIMIC-IV-ECG regression benchmarks, with 95\% CIs. The best value of each row is in bold; *, not significantly worse than the best model of the row (paired bootstrap of the difference, 95\% CI includes 0). $^\dagger$Evaluation-only benchmark.}
\label{tab:supp_leadI}
\setlength{\tabcolsep}{3pt}
\renewcommand{\arraystretch}{0.95}
\resizebox{\textwidth}{!}{%
\begin{tabular}{@{}lcccccccc@{}}
\toprule
Benchmark                     & ECGFounder                                  & ECG-JEPA                                    & ST-MEM                                      & MERL                                & ECGFM-KED                          & HuBERT-ECG                         & ECG-CPC                            & \textbf{RELAY}                           \\
\midrule
CODE 15\%                     & 95.2 {\scriptsize (93.5--97.0)}             & 94.2 {\scriptsize (91.1--96.7)}             & 51.7 {\scriptsize (45.8--57.2)}             & 71.4 {\scriptsize (65.8--76.6)}     & 71.0 {\scriptsize (66.4--75.1)}    & 58.0 {\scriptsize (52.5--63.9)}    & 58.7 {\scriptsize (53.8--63.9)}    & \textbf{95.5} {\scriptsize (92.7--97.3)} \\
PTB-XL All                    & 66.1 {\scriptsize (64.0--67.8)}             & 72.8 {\scriptsize (70.3--74.7)}             & 52.3 {\scriptsize (49.5--53.9)}             & 51.4 {\scriptsize (49.0--53.9)}     & 54.0 {\scriptsize (51.3--55.9)}    & 52.1 {\scriptsize (49.7--54.3)}    & 50.9 {\scriptsize (49.0--53.3)}    & \textbf{76.3} {\scriptsize (74.6--78.1)} \\
PTB-XL Diag$^\dagger$         & 65.7 {\scriptsize (62.9--68.0)}             & 71.1 {\scriptsize (68.0--73.9)}             & 52.4 {\scriptsize (49.2--54.9)}             & 52.2 {\scriptsize (49.0--55.5)}     & 55.8 {\scriptsize (52.1--58.0)}    & 49.1 {\scriptsize (46.3--52.2)}    & 48.3 {\scriptsize (46.0--51.7)}    & \textbf{75.8} {\scriptsize (73.4--78.1)} \\
PTB-XL Form$^\dagger$         & 59.7 {\scriptsize (56.3--62.8)}             & 70.7 {\scriptsize (68.5--74.4)}             & 50.7 {\scriptsize (47.2--56.0)}             & 47.7 {\scriptsize (45.0--51.7)}     & 49.4 {\scriptsize (45.5--52.7)}    & 52.4 {\scriptsize (49.2--58.1)}    & 53.5 {\scriptsize (50.5--57.0)}    & \textbf{75.1} {\scriptsize (72.2--77.8)} \\
PTB-XL Rhythm$^\dagger$       & 78.7 {\scriptsize (75.0--81.7)}             & \textbf{82.5} {\scriptsize (77.0--87.6)}    & 52.6 {\scriptsize (46.4--57.1)}             & 50.6 {\scriptsize (44.9--57.2)}     & 55.5 {\scriptsize (49.7--61.5)}    & 59.7 {\scriptsize (55.2--65.4)}    & 60.0 {\scriptsize (55.3--63.7)}    & 81.2 {\scriptsize (78.2--84.7)}          \\
PTB-XL Sub                    & 70.3 {\scriptsize (68.2--72.1)}             & 72.2 {\scriptsize (69.1--75.1)}             & 50.1 {\scriptsize (47.7--52.9)}             & 53.0 {\scriptsize (50.7--56.0)}     & 46.0 {\scriptsize (43.6--49.0)}    & 54.6 {\scriptsize (51.4--57.9)}    & 46.3 {\scriptsize (43.5--48.8)}    & \textbf{79.4} {\scriptsize (77.6--81.0)} \\
PTB-XL Super                  & 56.7 {\scriptsize (55.2--58.0)}             & 71.3 {\scriptsize (69.8--72.7)}             & 50.0 {\scriptsize (48.7--51.4)}             & 45.8 {\scriptsize (44.2--47.3)}     & 48.4 {\scriptsize (47.0--49.9)}    & 53.6 {\scriptsize (51.9--55.1)}    & 42.8 {\scriptsize (41.4--44.3)}    & \textbf{77.8} {\scriptsize (76.4--79.1)} \\
EchoNext                      & \textbf{72.9} {\scriptsize (71.5--74.2)}    & 72.7 {\scriptsize (71.0--74.2)}             & 52.3 {\scriptsize (50.5--54.3)}             & 57.5 {\scriptsize (55.7--59.2)}     & 56.9 {\scriptsize (55.3--58.7)}    & 65.2 {\scriptsize (63.5--66.9)}    & 59.8 {\scriptsize (58.1--61.4)}    & 67.4 {\scriptsize (65.1--68.7)}          \\
CPSC2018                      & 69.0 {\scriptsize (66.5--71.4)}             & \textbf{88.7} {\scriptsize (86.8--90.4)}    & 48.1 {\scriptsize (45.6--50.8)}             & 65.0 {\scriptsize (62.4--67.4)}     & 54.1 {\scriptsize (51.7--56.9)}    & 55.9 {\scriptsize (53.5--58.3)}    & 54.1 {\scriptsize (51.6--56.7)}    & 82.5 {\scriptsize (80.3--83.7)}          \\
CPSC-Extra                    & 65.7 {\scriptsize (60.5--70.3)}             & \textbf{80.1} {\scriptsize (76.0--82.8)}    & 51.0 {\scriptsize (46.1--55.4)}             & 60.0 {\scriptsize (54.3--63.7)}     & 51.6 {\scriptsize (46.5--55.5)}    & 55.0 {\scriptsize (49.9--58.9)}    & 63.8 {\scriptsize (59.5--67.6)}    & 73.1 {\scriptsize (69.0--75.9)}          \\
Chapman                       & 63.9 {\scriptsize (61.6--68.1)}             & 76.7 {\scriptsize (74.5--79.4)}             & 54.3 {\scriptsize (50.5--56.5)}             & 57.1 {\scriptsize (54.6--60.9)}     & 52.4 {\scriptsize (49.8--56.2)}    & 63.0 {\scriptsize (59.8--65.0)}    & 56.9 {\scriptsize (53.5--59.1)}    & \textbf{81.4} {\scriptsize (79.5--83.3)} \\
Chapman Rhythm$^\dagger$      & 78.6 {\scriptsize (76.1--82.5)}             & 91.2 {\scriptsize (89.4--92.9)}             & 52.0 {\scriptsize (47.0--56.8)}             & 63.1 {\scriptsize (60.8--71.8)}     & 61.4 {\scriptsize (56.6--63.7)}    & 60.3 {\scriptsize (56.9--62.9)}    & 56.9 {\scriptsize (53.8--62.0)}    & \textbf{93.0} {\scriptsize (91.3--94.4)} \\
Georgia                       & 69.4 {\scriptsize (67.2--71.4)}             & 75.4 {\scriptsize (72.6--76.9)}             & 49.4 {\scriptsize (46.9--52.5)}             & 59.7 {\scriptsize (56.3--62.2)}     & 57.7 {\scriptsize (54.4--60.2)}    & 49.1 {\scriptsize (46.6--52.1)}    & 62.2 {\scriptsize (58.4--64.0)}    & \textbf{75.9} {\scriptsize (73.5--77.8)} \\
Hefei                         & 71.5 {\scriptsize (68.0--75.2)}             & 75.7 {\scriptsize (72.6--79.2)}             & 53.7 {\scriptsize (49.2--57.0)}             & 58.1 {\scriptsize (54.4--63.5)}     & 53.9 {\scriptsize (51.4--59.1)}    & 59.5 {\scriptsize (54.3--64.0)}    & 58.0 {\scriptsize (54.4--61.2)}    & \textbf{75.9} {\scriptsize (72.4--80.4)} \\
Ningbo                        & 62.5 {\scriptsize (60.7--65.1)}             & \textbf{85.2} {\scriptsize (83.6--86.9)}    & 53.3 {\scriptsize (50.5--55.6)}             & 62.4 {\scriptsize (60.3--64.2)}     & 62.1 {\scriptsize (60.0--64.8)}    & 64.6 {\scriptsize (62.5--67.2)}    & 55.7 {\scriptsize (53.7--58.3)}    & 82.1 {\scriptsize (80.2--84.1)}          \\
PTB                           & 62.2 {\scriptsize (54.7--67.9)}*            & \textbf{62.4} {\scriptsize (55.8--72.5)}    & 49.7 {\scriptsize (43.6--55.7)}             & 57.4 {\scriptsize (49.8--63.4)}     & 51.5 {\scriptsize (45.1--57.5)}    & 60.4 {\scriptsize (53.4--66.1)}*   & 54.8 {\scriptsize (46.8--61.5)}    & 58.5 {\scriptsize (51.6--65.6)}*         \\
SPH                           & 68.2 {\scriptsize (64.0--70.4)}             & 81.5 {\scriptsize (78.8--85.5)}             & 50.8 {\scriptsize (46.7--54.8)}             & 60.5 {\scriptsize (55.9--63.4)}     & 53.1 {\scriptsize (48.9--56.6)}    & 65.3 {\scriptsize (61.7--69.1)}    & 60.9 {\scriptsize (56.9--65.5)}    & \textbf{84.6} {\scriptsize (82.3--86.3)} \\
ZZU pECG                      & 72.4 {\scriptsize (71.5--77.6)}             & \textbf{80.0} {\scriptsize (78.2--84.7)}    & 55.5 {\scriptsize (53.1--58.3)}             & 61.1 {\scriptsize (59.9--66.3)}     & 52.4 {\scriptsize (50.2--56.9)}    & 55.4 {\scriptsize (53.4--60.0)}    & 57.4 {\scriptsize (55.2--61.4)}    & 76.3 {\scriptsize (75.2--81.4)}          \\
MIMIC Cardiac                 & 60.0 {\scriptsize (58.9--61.0)}             & \textbf{69.7} {\scriptsize (68.4--70.9)}    & 54.9 {\scriptsize (53.8--56.0)}             & 53.1 {\scriptsize (51.9--54.3)}     & 56.9 {\scriptsize (55.6--58.1)}    & 58.7 {\scriptsize (57.0--60.3)}    & 57.1 {\scriptsize (55.9--58.4)}    & 68.1 {\scriptsize (66.9--69.1)}          \\
MIMIC Non-cardiac             & 52.8 {\scriptsize (52.2--53.3)}             & \textbf{63.0} {\scriptsize (62.2--63.7)}    & 53.1 {\scriptsize (52.5--53.7)}             & 54.3 {\scriptsize (53.6--55.1)}     & 55.3 {\scriptsize (54.5--56.1)}    & 56.4 {\scriptsize (55.4--57.3)}    & 55.0 {\scriptsize (54.3--55.6)}    & 62.3 {\scriptsize (61.6--62.9)}*         \\
MIMIC Deterioration           & 59.2 {\scriptsize (54.9--63.9)}             & 61.4 {\scriptsize (56.7--66.1)}             & 59.4 {\scriptsize (53.7--64.2)}             & 48.1 {\scriptsize (42.7--53.5)}     & 51.9 {\scriptsize (46.9--57.5)}    & 57.7 {\scriptsize (52.0--62.5)}    & 59.8 {\scriptsize (54.8--64.8)}    & \textbf{69.9} {\scriptsize (65.7--74.6)} \\
MIMIC Mortality               & 54.2 {\scriptsize (50.9--57.9)}             & 68.9 {\scriptsize (64.9--73.1)}             & 59.8 {\scriptsize (55.2--64.0)}             & 40.9 {\scriptsize (37.0--45.7)}     & 56.3 {\scriptsize (51.0--60.4)}    & 56.2 {\scriptsize (52.4--59.4)}    & 60.3 {\scriptsize (56.8--64.0)}    & \textbf{76.4} {\scriptsize (73.4--78.6)} \\
MIMIC ICU admission           & 56.4 {\scriptsize (54.2--58.4)}             & 61.2 {\scriptsize (58.7--63.6)}             & 54.8 {\scriptsize (52.5--57.0)}             & 47.5 {\scriptsize (45.0--49.9)}     & 52.5 {\scriptsize (50.1--54.7)}    & 56.1 {\scriptsize (53.9--58.4)}    & 57.3 {\scriptsize (55.2--59.4)}    & \textbf{67.2} {\scriptsize (65.0--69.2)} \\
MIMIC Sex                     & 59.1 {\scriptsize (57.1--61.1)}             & 64.1 {\scriptsize (61.4--66.2)}*            & 53.0 {\scriptsize (51.2--54.8)}             & 54.9 {\scriptsize (52.7--57.1)}     & 52.6 {\scriptsize (50.1--55.7)}    & 55.3 {\scriptsize (52.1--57.9)}    & 52.7 {\scriptsize (49.9--55.3)}    & \textbf{66.8} {\scriptsize (64.8--68.7)} \\
\midrule
MIMIC Age (MAE)               & 0.783 {\scriptsize (0.756--0.808)}          & \textbf{0.706} {\scriptsize (0.678--0.731)} & 0.870 {\scriptsize (0.838--0.901)}          & 0.904 {\scriptsize (0.870--0.937)}  & 0.864 {\scriptsize (0.831--0.895)} & 0.769 {\scriptsize (0.744--0.792)} & 0.800 {\scriptsize (0.775--0.825)} & 0.720 {\scriptsize (0.698--0.740)}*      \\
MIMIC Biometrics (MAE)        & 0.818 {\scriptsize (0.795--0.841)}          & 0.806 {\scriptsize (0.782--0.830)}          & \textbf{0.787} {\scriptsize (0.760--0.814)} & 0.789 {\scriptsize (0.759--0.819)}* & 0.867 {\scriptsize (0.834--0.900)} & 0.867 {\scriptsize (0.832--0.902)} & 0.804 {\scriptsize (0.777--0.829)} & 0.870 {\scriptsize (0.848--0.892)}       \\
MIMIC ECG features (MAE)      & \textbf{0.697} {\scriptsize (0.682--0.712)} & 0.710 {\scriptsize (0.696--0.724)}          & 0.779 {\scriptsize (0.762--0.795)}          & 0.806 {\scriptsize (0.790--0.821)}  & 0.736 {\scriptsize (0.721--0.751)} & 0.714 {\scriptsize (0.698--0.729)} & 0.743 {\scriptsize (0.727--0.758)} & 0.710 {\scriptsize (0.696--0.724)}       \\
MIMIC Laboratory values (MAE) & 0.803 {\scriptsize (0.780--0.843)}          & \textbf{0.676} {\scriptsize (0.657--0.700)} & 0.927 {\scriptsize (0.732--0.956)}          & 0.734 {\scriptsize (0.708--0.759)}  & 0.743 {\scriptsize (0.729--0.788)} & 0.703 {\scriptsize (0.673--0.733)} & 0.714 {\scriptsize (0.694--0.734)} & 0.736 {\scriptsize (0.698--0.753)}       \\
MIMIC Vital signs (MAE)       & 0.792 {\scriptsize (0.781--0.801)}          & \textbf{0.740} {\scriptsize (0.730--0.748)} & 0.761 {\scriptsize (0.751--0.771)}          & 0.777 {\scriptsize (0.766--0.785)}  & 0.775 {\scriptsize (0.765--0.783)} & 0.756 {\scriptsize (0.745--0.766)} & 0.763 {\scriptsize (0.752--0.772)} & 0.767 {\scriptsize (0.756--0.775)}       \\
\midrule
Mean AUROC (24 benchmarks) & 66.3 & 74.7 & 52.7 & 55.5 & 54.7 & 57.2 & 56.0 & \textbf{75.9} \\
Mean MAE (5 benchmarks) & 0.778 & \textbf{0.727} & 0.825 & 0.802 & 0.797 & 0.762 & 0.765 & 0.761 \\
\bottomrule
\end{tabular}}
\end{table*}

\begin{table*}[p]
\centering
\caption{\textbf{Performance of every model on every benchmark with lead II.} Macro AUROC (\%) for the 24 classification benchmarks and mean absolute error (MAE, lower is better) for the five MIMIC-IV-ECG regression benchmarks, with 95\% CIs. The best value of each row is in bold; *, not significantly worse than the best model of the row (paired bootstrap of the difference, 95\% CI includes 0). $^\dagger$Evaluation-only benchmark.}
\label{tab:supp_leadII}
\setlength{\tabcolsep}{3pt}
\renewcommand{\arraystretch}{0.95}
\resizebox{\textwidth}{!}{%
\begin{tabular}{@{}lcccccccc@{}}
\toprule
Benchmark                     & ECGFounder                               & ECG-JEPA                                    & ST-MEM                                   & MERL                                & ECGFM-KED                          & HuBERT-ECG                          & ECG-CPC                            & \textbf{RELAY}                              \\
\midrule
CODE 15\%                     & 94.5 {\scriptsize (92.4--96.2)}          & 94.4 {\scriptsize (92.2--96.0)}             & 90.9 {\scriptsize (87.9--93.8)}          & 70.5 {\scriptsize (64.4--75.5)}     & 70.6 {\scriptsize (65.2--75.4)}    & 58.4 {\scriptsize (52.0--64.8)}     & 57.7 {\scriptsize (52.6--62.9)}    & \textbf{96.4} {\scriptsize (94.8--97.6)}    \\
PTB-XL All                    & 62.9 {\scriptsize (60.4--64.7)}          & 74.0 {\scriptsize (71.6--75.9)}             & 74.3 {\scriptsize (72.2--75.7)}          & 52.6 {\scriptsize (50.5--54.7)}     & 51.0 {\scriptsize (48.2--53.5)}    & 53.2 {\scriptsize (50.9--55.0)}     & 58.8 {\scriptsize (56.7--61.5)}    & \textbf{78.4} {\scriptsize (76.4--79.9)}    \\
PTB-XL Diag$^\dagger$         & 60.2 {\scriptsize (57.2--62.6)}          & 72.4 {\scriptsize (69.3--74.7)}             & 73.2 {\scriptsize (70.5--75.2)}          & 55.4 {\scriptsize (52.1--57.8)}     & 54.3 {\scriptsize (50.1--57.1)}    & 51.1 {\scriptsize (48.5--53.7)}     & 59.0 {\scriptsize (56.1--61.9)}    & \textbf{78.2} {\scriptsize (75.8--80.0)}    \\
PTB-XL Form$^\dagger$         & 55.6 {\scriptsize (52.7--60.6)}          & 68.2 {\scriptsize (65.9--72.3)}             & 68.9 {\scriptsize (67.0--72.2)}          & 52.2 {\scriptsize (48.7--55.6)}     & 49.9 {\scriptsize (44.1--52.8)}    & 52.8 {\scriptsize (49.1--57.6)}     & 55.3 {\scriptsize (50.6--58.0)}    & \textbf{73.7} {\scriptsize (70.9--78.3)}    \\
PTB-XL Rhythm$^\dagger$       & 80.6 {\scriptsize (76.6--83.7)}          & \textbf{87.9} {\scriptsize (82.8--92.8)}    & 85.8 {\scriptsize (80.9--90.2)}          & 46.0 {\scriptsize (41.7--51.5)}     & 46.1 {\scriptsize (40.5--52.3)}    & 61.1 {\scriptsize (56.1--64.1)}     & 66.0 {\scriptsize (59.6--73.2)}    & 84.6 {\scriptsize (81.2--88.2)}             \\
PTB-XL Sub                    & 66.4 {\scriptsize (63.1--69.3)}          & 76.0 {\scriptsize (73.6--77.9)}             & 73.2 {\scriptsize (70.7--75.5)}          & 53.9 {\scriptsize (50.7--57.5)}     & 50.4 {\scriptsize (48.0--53.0)}    & 55.9 {\scriptsize (53.5--58.5)}     & 50.8 {\scriptsize (48.7--53.1)}    & \textbf{79.4} {\scriptsize (77.4--81.1)}    \\
PTB-XL Super                  & 60.9 {\scriptsize (59.4--62.4)}          & 72.2 {\scriptsize (70.9--73.6)}             & 68.0 {\scriptsize (66.7--69.3)}          & 51.4 {\scriptsize (49.7--53.1)}     & 50.5 {\scriptsize (49.1--52.0)}    & 52.7 {\scriptsize (51.5--54.2)}     & 51.5 {\scriptsize (49.9--53.0)}    & \textbf{77.8} {\scriptsize (76.5--79.1)}    \\
EchoNext                      & \textbf{72.1} {\scriptsize (70.7--73.6)} & 69.8 {\scriptsize (68.3--71.2)}             & 63.5 {\scriptsize (61.5--65.2)}          & 58.6 {\scriptsize (56.9--60.3)}     & 54.0 {\scriptsize (52.3--55.8)}    & 62.6 {\scriptsize (60.7--64.6)}     & 52.9 {\scriptsize (51.0--55.0)}    & 69.3 {\scriptsize (67.2--70.5)}             \\
CPSC2018                      & 71.1 {\scriptsize (68.7--73.4)}          & \textbf{89.4} {\scriptsize (87.3--91.5)}    & 85.5 {\scriptsize (83.2--87.5)}          & 62.4 {\scriptsize (59.7--64.9)}     & 49.9 {\scriptsize (47.2--52.8)}    & 56.6 {\scriptsize (54.3--58.9)}     & 59.9 {\scriptsize (57.2--62.8)}    & 83.6 {\scriptsize (81.2--85.3)}             \\
CPSC-Extra                    & 68.8 {\scriptsize (63.0--72.4)}          & 80.1 {\scriptsize (75.9--83.5)}             & \textbf{81.9} {\scriptsize (78.7--84.2)} & 59.2 {\scriptsize (55.1--62.7)}     & 53.8 {\scriptsize (50.0--58.9)}    & 58.3 {\scriptsize (53.4--62.0)}     & 60.2 {\scriptsize (56.0--65.1)}    & 75.7 {\scriptsize (71.7--78.6)}             \\
Chapman                       & 74.7 {\scriptsize (72.1--76.7)}          & 83.7 {\scriptsize (81.5--86.1)}             & 78.0 {\scriptsize (75.6--80.9)}          & 65.0 {\scriptsize (62.6--67.8)}     & 64.5 {\scriptsize (61.8--67.1)}    & 58.1 {\scriptsize (55.2--60.4)}     & 59.7 {\scriptsize (56.7--62.6)}    & \textbf{85.0} {\scriptsize (83.3--86.8)}    \\
Chapman Rhythm$^\dagger$      & 83.3 {\scriptsize (81.4--85.6)}          & \textbf{96.6} {\scriptsize (95.4--97.5)}    & 88.9 {\scriptsize (86.5--90.7)}          & 65.8 {\scriptsize (62.9--73.4)}     & 72.1 {\scriptsize (69.3--74.2)}    & 61.3 {\scriptsize (57.8--64.2)}     & 52.3 {\scriptsize (49.6--59.8)}    & 94.0 {\scriptsize (92.2--95.9)}             \\
Georgia                       & 66.5 {\scriptsize (64.5--69.0)}          & 76.0 {\scriptsize (73.5--77.5)}             & \textbf{77.9} {\scriptsize (75.7--79.9)} & 55.7 {\scriptsize (52.4--58.6)}     & 56.9 {\scriptsize (53.9--59.6)}    & 52.0 {\scriptsize (50.0--55.2)}     & 55.1 {\scriptsize (52.6--57.9)}    & 76.9 {\scriptsize (74.7--79.2)}             \\
Hefei                         & 78.7 {\scriptsize (74.4--82.0)}          & 79.9 {\scriptsize (76.5--82.6)}             & 83.0 {\scriptsize (80.1--85.3)}          & 64.4 {\scriptsize (60.3--69.2)}     & 57.4 {\scriptsize (54.8--62.2)}    & 65.8 {\scriptsize (60.3--69.4)}     & 58.4 {\scriptsize (54.5--62.6)}    & \textbf{85.0} {\scriptsize (82.0--86.7)}    \\
Ningbo                        & 73.8 {\scriptsize (71.6--75.0)}          & \textbf{89.2} {\scriptsize (87.8--90.2)}    & 79.4 {\scriptsize (77.4--80.8)}          & 68.7 {\scriptsize (66.6--70.6)}     & 58.6 {\scriptsize (56.3--61.0)}    & 65.2 {\scriptsize (63.3--67.5)}     & 64.0 {\scriptsize (61.7--65.7)}    & 85.6 {\scriptsize (83.7--86.9)}             \\
PTB                           & 57.2 {\scriptsize (48.3--63.9)}          & 63.4 {\scriptsize (56.1--69.6)}*            & \textbf{67.1} {\scriptsize (59.5--71.9)} & 55.8 {\scriptsize (47.0--62.6)}     & 50.6 {\scriptsize (43.8--57.8)}    & 61.2 {\scriptsize (53.3--67.6)}     & 49.4 {\scriptsize (41.8--57.3)}    & 65.1 {\scriptsize (58.0--70.4)}             \\
SPH                           & 66.3 {\scriptsize (62.0--69.4)}          & \textbf{90.2} {\scriptsize (88.2--91.5)}    & 82.7 {\scriptsize (80.6--85.4)}          & 65.2 {\scriptsize (62.2--69.4)}     & 53.4 {\scriptsize (48.8--56.9)}    & 62.2 {\scriptsize (58.8--66.3)}     & 63.3 {\scriptsize (59.8--67.8)}    & 87.1 {\scriptsize (84.8--89.5)}             \\
ZZU pECG                      & 78.8 {\scriptsize (77.7--83.2)}          & \textbf{81.7} {\scriptsize (80.8--87.0)}    & 78.2 {\scriptsize (77.4--83.4)}          & 67.5 {\scriptsize (66.3--72.0)}     & 72.7 {\scriptsize (71.4--77.4)}    & 54.4 {\scriptsize (51.5--59.5)}     & 64.7 {\scriptsize (62.3--68.4)}    & 81.1 {\scriptsize (80.3--86.7)}             \\
MIMIC Cardiac                 & 55.5 {\scriptsize (54.3--56.8)}          & 63.9 {\scriptsize (62.3--65.2)}             & 53.4 {\scriptsize (52.2--54.5)}          & 58.5 {\scriptsize (57.4--59.7)}     & 57.8 {\scriptsize (56.6--59.0)}    & 55.3 {\scriptsize (53.7--56.8)}     & 57.6 {\scriptsize (56.5--59.0)}    & \textbf{66.7} {\scriptsize (65.6--67.8)}    \\
MIMIC Non-cardiac             & 50.5 {\scriptsize (49.9--51.1)}          & 59.1 {\scriptsize (58.1--60.0)}             & 52.1 {\scriptsize (51.4--52.7)}          & 55.6 {\scriptsize (54.9--56.3)}     & 57.1 {\scriptsize (56.3--57.8)}    & 52.9 {\scriptsize (52.1--53.6)}     & 55.1 {\scriptsize (54.4--55.8)}    & \textbf{61.1} {\scriptsize (60.5--61.8)}    \\
MIMIC Deterioration           & 54.0 {\scriptsize (49.1--58.6)}          & 58.7 {\scriptsize (52.9--64.8)}             & 60.2 {\scriptsize (54.6--65.1)}          & 53.9 {\scriptsize (48.6--58.7)}     & 56.4 {\scriptsize (50.4--62.1)}    & 59.0 {\scriptsize (54.1--63.6)}     & 59.2 {\scriptsize (53.6--64.8)}    & \textbf{70.5} {\scriptsize (66.2--74.5)}    \\
MIMIC Mortality               & 51.6 {\scriptsize (46.6--57.4)}          & 61.9 {\scriptsize (55.1--67.1)}             & 58.6 {\scriptsize (54.0--62.4)}          & 51.0 {\scriptsize (46.1--56.0)}     & 62.7 {\scriptsize (59.1--65.9)}    & 56.0 {\scriptsize (49.7--60.8)}     & 62.3 {\scriptsize (58.0--67.8)}    & \textbf{71.6} {\scriptsize (68.5--74.4)}    \\
MIMIC ICU admission           & 51.4 {\scriptsize (49.0--53.9)}          & 60.3 {\scriptsize (57.6--62.8)}             & 53.9 {\scriptsize (51.5--56.3)}          & 52.9 {\scriptsize (50.6--55.3)}     & 57.0 {\scriptsize (54.6--59.5)}    & 57.5 {\scriptsize (55.2--59.6)}     & 54.7 {\scriptsize (52.4--57.2)}    & \textbf{68.5} {\scriptsize (66.6--70.4)}    \\
MIMIC Sex                     & 59.1 {\scriptsize (56.0--61.8)}          & 63.8 {\scriptsize (60.7--66.3)}             & 53.2 {\scriptsize (51.3--54.9)}          & 54.6 {\scriptsize (52.4--57.0)}     & 60.9 {\scriptsize (58.3--63.3)}    & 52.8 {\scriptsize (49.8--55.3)}     & 56.1 {\scriptsize (52.9--59.0)}    & \textbf{70.0} {\scriptsize (67.9--71.9)}    \\
\midrule
MIMIC Age (MAE)               & 0.773 {\scriptsize (0.744--0.800)}*      & \textbf{0.767} {\scriptsize (0.741--0.790)} & 0.814 {\scriptsize (0.783--0.844)}       & 0.785 {\scriptsize (0.756--0.814)}* & 0.835 {\scriptsize (0.804--0.866)} & 0.878 {\scriptsize (0.847--0.906)}  & 0.794 {\scriptsize (0.767--0.822)} & 0.799 {\scriptsize (0.775--0.820)}          \\
MIMIC Biometrics (MAE)        & 0.811 {\scriptsize (0.785--0.837)}       & \textbf{0.785} {\scriptsize (0.760--0.810)} & 0.806 {\scriptsize (0.781--0.830)}       & 0.824 {\scriptsize (0.789--0.856)}  & 0.842 {\scriptsize (0.809--0.876)} & 0.786 {\scriptsize (0.756--0.815)}* & 0.798 {\scriptsize (0.772--0.823)} & 0.926 {\scriptsize (0.902--0.948)}          \\
MIMIC ECG features (MAE)      & 0.699 {\scriptsize (0.685--0.712)}       & 0.676 {\scriptsize (0.662--0.690)}          & 0.769 {\scriptsize (0.753--0.785)}       & 0.824 {\scriptsize (0.809--0.840)}  & 0.722 {\scriptsize (0.710--0.735)} & 0.726 {\scriptsize (0.709--0.742)}  & 0.730 {\scriptsize (0.716--0.744)} & \textbf{0.655} {\scriptsize (0.642--0.667)} \\
MIMIC Laboratory values (MAE) & 0.755 {\scriptsize (0.738--0.785)}       & \textbf{0.674} {\scriptsize (0.650--0.700)} & 0.920 {\scriptsize (0.724--0.951)}       & 0.713 {\scriptsize (0.687--0.739)}  & 0.721 {\scriptsize (0.706--0.770)} & 0.706 {\scriptsize (0.672--0.740)}  & 0.713 {\scriptsize (0.684--0.741)} & 0.715 {\scriptsize (0.674--0.754)}*         \\
MIMIC Vital signs (MAE)       & 0.779 {\scriptsize (0.769--0.788)}       & \textbf{0.753} {\scriptsize (0.743--0.761)} & 0.768 {\scriptsize (0.758--0.778)}       & 0.798 {\scriptsize (0.786--0.808)}  & 0.784 {\scriptsize (0.773--0.793)} & 0.764 {\scriptsize (0.753--0.775)}  & 0.763 {\scriptsize (0.752--0.773)} & 0.764 {\scriptsize (0.754--0.772)}          \\
\midrule
Mean AUROC (24 benchmarks) & 66.4 & 75.5 & 72.2 & 58.2 & 57.0 & 57.4 & 57.7 & \textbf{77.7} \\
Mean MAE (5 benchmarks) & 0.763 & \textbf{0.731} & 0.816 & 0.789 & 0.781 & 0.772 & 0.760 & 0.771 \\
\bottomrule
\end{tabular}}
\end{table*}

\begin{table*}[p]
\centering
\caption{\textbf{Performance of every model on every benchmark with one lead sampled randomly.} Macro AUROC (\%) for the 24 classification benchmarks and mean absolute error (MAE, lower is better) for the five MIMIC-IV-ECG regression benchmarks, with 95\% CIs of the seed mean (the five seeds are averaged within each bootstrap resample). Values are means over five seeds. The best value of each row is in bold; *, not significantly worse than the best model of the row (paired bootstrap of the difference, 95\% CI includes 0). $^\dagger$Evaluation-only benchmark.}
\label{tab:supp_random1}
\setlength{\tabcolsep}{3pt}
\renewcommand{\arraystretch}{0.95}
\resizebox{\textwidth}{!}{%
\begin{tabular}{@{}lcccccccc@{}}
\toprule
Benchmark                     & ECGFounder                                      & ECG-JEPA                                        & ST-MEM                                          & MERL                                            & ECGFM-KED                                       & HuBERT-ECG                                      & ECG-CPC                                         & \textbf{RELAY}                                           \\
\midrule
CODE 15\%                     & 94.6 {\scriptsize (92.6--96.4)}                 & 77.2 {\scriptsize (74.4--79.5)}                 & 78.4 {\scriptsize (75.6--80.6)}                 & 80.1 {\scriptsize (76.6--82.9)}                 & 64.6 {\scriptsize (61.8--67.3)}                 & 51.7 {\scriptsize (48.3--54.8)}                 & 65.6 {\scriptsize (62.3--69.0)}                 & \textbf{95.9} {\scriptsize (94.5--97.1)}                 \\
PTB-XL All                    & 68.8 {\scriptsize (67.3--70.2)}                 & 63.8 {\scriptsize (62.9--65.0)}                 & 61.8 {\scriptsize (60.4--62.9)}                 & 57.1 {\scriptsize (55.7--58.8)}                 & 52.9 {\scriptsize (51.2--54.6)}                 & 51.9 {\scriptsize (50.7--53.3)}                 & 59.8 {\scriptsize (58.5--61.5)}                 & \textbf{74.4} {\scriptsize (73.2--75.6)}                 \\
PTB-XL Diag$^\dagger$         & 67.8 {\scriptsize (65.7--69.6)}                 & 63.3 {\scriptsize (61.9--64.8)}                 & 60.7 {\scriptsize (58.9--62.2)}                 & 70.4 {\scriptsize (68.5--71.2)}                 & 54.5 {\scriptsize (54.1--57.1)}                 & 50.5 {\scriptsize (49.3--52.5)}                 & 58.8 {\scriptsize (56.7--61.1)}                 & \textbf{72.9} {\scriptsize (71.4--74.5)}                 \\
PTB-XL Form$^\dagger$         & 62.6 {\scriptsize (59.4--65.4)}                 & 62.7 {\scriptsize (60.6--64.3)}                 & 58.1 {\scriptsize (56.4--60.3)}                 & 66.9 {\scriptsize (64.7--68.7)}                 & 46.3 {\scriptsize (45.5--48.7)}                 & 51.0 {\scriptsize (48.8--53.2)}                 & 57.3 {\scriptsize (54.9--60.0)}                 & \textbf{72.4} {\scriptsize (71.0--74.7)}                 \\
PTB-XL Rhythm$^\dagger$       & 81.5 {\scriptsize (78.9--84.0)}                 & 70.2 {\scriptsize (68.2--72.8)}                 & 71.5 {\scriptsize (68.3--74.2)}                 & 53.8 {\scriptsize (50.4--56.8)}                 & 55.3 {\scriptsize (52.8--58.6)}                 & 59.0 {\scriptsize (54.5--61.4)}                 & 67.1 {\scriptsize (64.6--70.7)}                 & \textbf{82.6} {\scriptsize (80.3--85.1)}                 \\
PTB-XL Sub                    & 70.5 {\scriptsize (68.9--71.9)}                 & 62.7 {\scriptsize (61.8--64.3)}                 & 62.0 {\scriptsize (60.3--63.4)}                 & 63.2 {\scriptsize (61.2--65.2)}                 & 49.8 {\scriptsize (47.6--52.3)}                 & 52.3 {\scriptsize (50.5--54.4)}                 & 54.3 {\scriptsize (53.2--56.4)}                 & \textbf{75.3} {\scriptsize (74.0--76.5)}                 \\
PTB-XL Super                  & 62.1 {\scriptsize (60.9--63.1)}                 & 59.9 {\scriptsize (59.5--61.0)}                 & 57.8 {\scriptsize (57.0--58.6)}                 & 56.5 {\scriptsize (55.6--57.7)}                 & 51.5 {\scriptsize (50.9--52.4)}                 & 49.9 {\scriptsize (49.0--51.0)}                 & 49.8 {\scriptsize (49.0--51.1)}                 & \textbf{72.3} {\scriptsize (71.4--73.2)}                 \\
EchoNext                      & \textbf{69.4} {\scriptsize (68.5--70.9)}        & 60.3 {\scriptsize (59.2--61.4)}                 & 56.9 {\scriptsize (56.0--57.8)}                 & 56.3 {\scriptsize (55.6--57.9)}                 & 52.7 {\scriptsize (52.2--53.9)}                 & 59.3 {\scriptsize (58.0--60.6)}                 & 56.6 {\scriptsize (55.6--58.0)}                 & 66.1 {\scriptsize (65.0--67.1)}                          \\
CPSC2018                      & 75.3 {\scriptsize (74.2--77.1)}                 & 75.4 {\scriptsize (74.3--77.0)}                 & 76.2 {\scriptsize (74.5--77.7)}                 & 64.9 {\scriptsize (63.5--66.3)}                 & 46.5 {\scriptsize (45.2--48.2)}                 & 55.9 {\scriptsize (54.2--57.6)}                 & 65.6 {\scriptsize (64.0--67.4)}                 & \textbf{84.7} {\scriptsize (83.3--86.0)}                 \\
CPSC-Extra                    & 70.6 {\scriptsize (68.0--72.6)}                 & 67.0 {\scriptsize (65.1--69.3)}                 & 66.5 {\scriptsize (64.1--69.0)}                 & 56.7 {\scriptsize (54.3--59.8)}                 & 45.6 {\scriptsize (44.6--47.6)}                 & 58.7 {\scriptsize (55.1--60.9)}                 & 60.3 {\scriptsize (58.7--62.1)}                 & \textbf{73.6} {\scriptsize (70.4--75.9)}                 \\
Chapman                       & 75.3 {\scriptsize (73.6--77.7)}                 & 70.7 {\scriptsize (69.2--72.3)}                 & 66.0 {\scriptsize (64.3--67.9)}                 & 64.6 {\scriptsize (63.1--67.4)}                 & 55.0 {\scriptsize (53.5--57.4)}                 & 58.6 {\scriptsize (55.9--60.2)}                 & 62.4 {\scriptsize (60.7--63.9)}                 & \textbf{80.7} {\scriptsize (79.3--81.9)}                 \\
Chapman Rhythm$^\dagger$      & 87.7 {\scriptsize (86.4--89.8)}                 & 81.0 {\scriptsize (79.2--82.6)}                 & 78.4 {\scriptsize (76.9--79.4)}                 & 67.8 {\scriptsize (66.3--75.4)}                 & 62.6 {\scriptsize (59.9--64.8)}                 & 59.2 {\scriptsize (55.4--61.0)}                 & 66.0 {\scriptsize (64.2--67.9)}                 & \textbf{91.7} {\scriptsize (90.5--92.5)}                 \\
Georgia                       & 71.5 {\scriptsize (69.1--73.0)}                 & 60.9 {\scriptsize (58.6--62.5)}                 & 65.2 {\scriptsize (63.8--67.2)}                 & 60.2 {\scriptsize (58.0--62.8)}                 & 50.8 {\scriptsize (48.4--52.7)}                 & 50.8 {\scriptsize (49.4--52.4)}                 & 58.3 {\scriptsize (57.2--61.0)}                 & \textbf{75.1} {\scriptsize (73.7--76.7)}                 \\
Hefei                         & 70.9 {\scriptsize (68.4--72.9)}                 & 69.7 {\scriptsize (67.1--71.7)}                 & 69.4 {\scriptsize (67.2--71.3)}                 & 62.4 {\scriptsize (59.2--66.5)}                 & 51.4 {\scriptsize (49.8--54.9)}                 & 59.9 {\scriptsize (56.7--62.4)}                 & 63.2 {\scriptsize (60.3--65.6)}                 & \textbf{81.5} {\scriptsize (78.8--83.6)}                 \\
Ningbo                        & 77.7 {\scriptsize (77.5--80.2)}                 & 67.5 {\scriptsize (67.3--70.1)}                 & 67.3 {\scriptsize (66.3--68.5)}                 & 64.2 {\scriptsize (63.1--65.2)}                 & 56.7 {\scriptsize (55.8--57.6)}                 & 65.7 {\scriptsize (64.5--67.4)}                 & 60.5 {\scriptsize (59.1--61.3)}                 & \textbf{82.2} {\scriptsize (81.2--83.2)}                 \\
PTB                           & 58.8 {\scriptsize (55.2--72.5)}                 & 60.1 {\scriptsize (56.8--70.2)}                 & 59.1 {\scriptsize (55.1--62.4)}                 & 53.5 {\scriptsize (50.0--63.2)}                 & 52.7 {\scriptsize (49.0--56.0)}                 & 54.6 {\scriptsize (53.1--63.7)}                 & 56.3 {\scriptsize (55.3--66.6)}                 & \textbf{60.4} {\scriptsize (53.3--67.0)}                 \\
SPH                           & 75.2 {\scriptsize (72.6--76.2)}                 & 74.0 {\scriptsize (72.0--75.8)}                 & 68.8 {\scriptsize (66.8--70.3)}                 & 66.3 {\scriptsize (63.9--68.4)}                 & 55.5 {\scriptsize (53.1--57.4)}                 & 58.6 {\scriptsize (56.2--60.7)}                 & 68.0 {\scriptsize (65.5--70.9)}                 & \textbf{81.9} {\scriptsize (80.3--83.6)}                 \\
ZZU pECG                      & 79.4 {\scriptsize (77.9--83.4)}                 & 78.3 {\scriptsize (76.4--82.1)}                 & 68.1 {\scriptsize (67.1--71.8)}                 & 68.0 {\scriptsize (66.8--72.8)}                 & 60.7 {\scriptsize (58.6--64.1)}                 & 57.8 {\scriptsize (56.0--61.9)} & 62.5 {\scriptsize (61.6--65.5)}                 & \textbf{79.6} {\scriptsize (78.9--83.3)} \\
MIMIC Cardiac                 & 59.1 {\scriptsize (58.3--59.9)}                 & 59.3 {\scriptsize (58.6--60.0)}                 & 54.2 {\scriptsize (53.3--55.0)}                 & 57.5 {\scriptsize (56.6--58.5)}                 & 55.0 {\scriptsize (54.2--55.7)}                 & 55.3 {\scriptsize (54.4--56.0)}                 & 56.9 {\scriptsize (56.2--57.7)}                 & \textbf{64.9} {\scriptsize (64.1--65.6)}                 \\
MIMIC Non-cardiac             & 53.4 {\scriptsize (53.0--53.8)}                 & 54.5 {\scriptsize (54.2--54.8)}                 & 53.2 {\scriptsize (52.7--53.7)}                 & 55.4 {\scriptsize (54.9--56.0)}                 & 53.8 {\scriptsize (53.4--54.3)}                 & 52.3 {\scriptsize (51.9--52.8)}                 & 54.1 {\scriptsize (53.6--54.5)}                 & \textbf{59.8} {\scriptsize (59.4--60.3)}                 \\
MIMIC Deterioration           & 55.1 {\scriptsize (51.6--59.1)}                 & 53.3 {\scriptsize (50.7--55.7)}                 & 58.7 {\scriptsize (54.7--62.3)}                 & 53.2 {\scriptsize (48.5--57.8)}                 & 51.0 {\scriptsize (47.3--55.0)}                 & 56.1 {\scriptsize (53.2--58.6)}                 & 56.2 {\scriptsize (53.4--59.1)}                 & \textbf{63.5} {\scriptsize (60.4--66.3)}                 \\
MIMIC Mortality               & 55.3 {\scriptsize (52.5--58.1)}                 & 56.4 {\scriptsize (54.3--58.1)}                 & 58.7 {\scriptsize (54.0--62.2)}                 & 52.7 {\scriptsize (49.3--55.7)}                 & 54.0 {\scriptsize (50.8--56.7)}                 & 53.9 {\scriptsize (50.1--56.9)}                 & 59.7 {\scriptsize (57.4--61.9)}                 & \textbf{71.8} {\scriptsize (69.6--73.9)}                 \\
MIMIC ICU admission           & 52.4 {\scriptsize (50.6--54.1)}                 & 54.4 {\scriptsize (53.3--55.4)}                 & 56.3 {\scriptsize (54.4--58.3)}                 & 53.6 {\scriptsize (51.6--55.4)}                 & 52.1 {\scriptsize (50.7--53.6)}                 & 55.6 {\scriptsize (54.2--57.0)}                 & 55.7 {\scriptsize (54.3--57.0)}                 & \textbf{66.1} {\scriptsize (64.6--67.6)}                 \\
MIMIC Sex                     & 62.5 {\scriptsize (61.0--64.0)}                 & 56.3 {\scriptsize (55.6--57.0)}                 & 55.3 {\scriptsize (53.8--56.6)}                 & 58.5 {\scriptsize (57.4--59.8)}                 & 53.8 {\scriptsize (52.8--54.8)}                 & 51.1 {\scriptsize (49.5--52.6)}                 & 54.2 {\scriptsize (53.0--55.6)}                 & \textbf{70.5} {\scriptsize (69.2--71.9)}                 \\
\midrule
MIMIC Age (MAE)               & 0.821 {\scriptsize (0.790--0.852)}              & 0.802 {\scriptsize (0.773--0.831)}              & 0.850 {\scriptsize (0.817--0.881)}              & 0.822 {\scriptsize (0.792--0.853)}              & 0.884 {\scriptsize (0.851--0.915)}              & 0.838 {\scriptsize (0.810--0.862)}              & 0.843 {\scriptsize (0.810--0.874)}              & \textbf{0.733} {\scriptsize (0.715--0.750)}              \\
MIMIC Biometrics (MAE)        & 0.782 {\scriptsize (0.754--0.812)}*             & 0.800 {\scriptsize (0.770--0.830)}              & 0.786 {\scriptsize (0.758--0.813)}*             & 0.808 {\scriptsize (0.774--0.841)}              & 0.833 {\scriptsize (0.802--0.863)}              & 0.797 {\scriptsize (0.766--0.825)}              & \textbf{0.782} {\scriptsize (0.755--0.808)}     & 0.900 {\scriptsize (0.880--0.918)}                       \\
MIMIC ECG features (MAE)      & 0.669 {\scriptsize (0.654--0.682)}              & 0.714 {\scriptsize (0.699--0.729)}              & 0.748 {\scriptsize (0.730--0.764)}              & 0.738 {\scriptsize (0.722--0.753)}              & 0.764 {\scriptsize (0.750--0.777)}              & 0.721 {\scriptsize (0.705--0.737)}              & 0.718 {\scriptsize (0.702--0.732)}              & \textbf{0.652} {\scriptsize (0.641--0.664)}              \\
MIMIC Laboratory values (MAE) & 0.771 {\scriptsize (0.755--0.808)}              & \textbf{0.692} {\scriptsize (0.674--0.717)}     & 0.914 {\scriptsize (0.724--0.947)}              & 0.706 {\scriptsize (0.683--0.732)}*             & 0.793 {\scriptsize (0.778--0.833)}              & 0.694 {\scriptsize (0.663--0.727)}*             & 0.733 {\scriptsize (0.705--0.759)}              & 0.701 {\scriptsize (0.684--0.720)}*                      \\
MIMIC Vital signs (MAE)       & 0.790 {\scriptsize (0.779--0.799)}              & 0.755 {\scriptsize (0.745--0.763)}              & 0.754 {\scriptsize (0.744--0.764)}              & 0.769 {\scriptsize (0.758--0.778)}              & 0.792 {\scriptsize (0.782--0.800)}              & 0.756 {\scriptsize (0.745--0.765)}              & 0.758 {\scriptsize (0.748--0.767)}              & \textbf{0.742} {\scriptsize (0.732--0.750)}              \\
\midrule
Mean AUROC (24 benchmarks) & 69.1 & 65.0 & 63.7 & 61.0 & 53.5 & 55.4 & 59.5 & \textbf{75.0} \\
Mean MAE (5 benchmarks) & 0.767 & 0.753 & 0.810 & 0.769 & 0.813 & 0.761 & 0.767 & \textbf{0.746} \\
\bottomrule
\end{tabular}}
\end{table*}

\begin{table*}[p]
\centering
\caption{\textbf{Performance of every model on every benchmark with two leads sampled randomly.} Macro AUROC (\%) for the 24 classification benchmarks and mean absolute error (MAE, lower is better) for the five MIMIC-IV-ECG regression benchmarks, with 95\% CIs of the seed mean (the five seeds are averaged within each bootstrap resample). Values are means over five seeds. The best value of each row is in bold; *, not significantly worse than the best model of the row (paired bootstrap of the difference, 95\% CI includes 0). $^\dagger$Evaluation-only benchmark.}
\label{tab:supp_random2}
\setlength{\tabcolsep}{3pt}
\renewcommand{\arraystretch}{0.95}
\resizebox{\textwidth}{!}{%
\begin{tabular}{@{}lcccccccc@{}}
\toprule
Benchmark                     & ECGFounder                               & ECG-JEPA                            & ST-MEM                                   & MERL                                            & ECGFM-KED                                       & HuBERT-ECG                          & ECG-CPC                                     & \textbf{RELAY}                              \\
\midrule
CODE 15\%                     & 98.0 {\scriptsize (97.2--98.7)}          & 88.4 {\scriptsize (86.4--90.1)}     & 85.5 {\scriptsize (83.6--87.5)}          & 88.6 {\scriptsize (85.6--91.1)}                 & 78.5 {\scriptsize (75.6--81.1)}                 & 62.2 {\scriptsize (59.0--64.8)}     & 71.5 {\scriptsize (68.7--74.4)}             & \textbf{98.2} {\scriptsize (97.5--98.7)}    \\
PTB-XL All                    & 80.2 {\scriptsize (78.9--81.3)}          & 71.1 {\scriptsize (69.8--72.4)}     & 66.1 {\scriptsize (64.8--67.3)}          & 65.9 {\scriptsize (64.3--67.7)}                 & 58.3 {\scriptsize (58.2--60.9)}                 & 57.7 {\scriptsize (56.2--59.2)}     & 68.6 {\scriptsize (67.4--70.5)}             & \textbf{80.9} {\scriptsize (79.8--82.1)}    \\
PTB-XL Diag$^\dagger$         & 80.8 {\scriptsize (79.2--82.2)}          & 69.9 {\scriptsize (68.4--71.4)}     & 65.4 {\scriptsize (63.7--66.7)}          & 73.6 {\scriptsize (71.8--74.7)}                 & 58.7 {\scriptsize (58.3--61.5)}                 & 56.4 {\scriptsize (54.9--58.5)}     & 68.4 {\scriptsize (66.4--70.6)}             & \textbf{81.7} {\scriptsize (80.4--82.9)}    \\
PTB-XL Form$^\dagger$         & 72.4 {\scriptsize (70.2--75.2)}          & 69.1 {\scriptsize (67.6--70.8)}     & 63.2 {\scriptsize (60.6--64.5)}          & 66.8 {\scriptsize (63.3--67.8)}                 & 54.4 {\scriptsize (52.5--57.9)}                 & 54.9 {\scriptsize (52.7--58.6)}     & 65.1 {\scriptsize (60.8--66.9)}             & \textbf{76.9} {\scriptsize (75.1--78.9)}    \\
PTB-XL Rhythm$^\dagger$       & \textbf{89.5} {\scriptsize (87.3--91.5)} & 80.1 {\scriptsize (76.1--83.8)}     & 74.7 {\scriptsize (71.4--76.9)}          & 67.7 {\scriptsize (63.7--71.8)}                 & 65.5 {\scriptsize (62.3--68.7)}                 & 64.8 {\scriptsize (60.0--67.7)}     & 78.2 {\scriptsize (75.7--79.9)}             & 85.2 {\scriptsize (81.6--89.0)}             \\
PTB-XL Sub                    & \textbf{83.5} {\scriptsize (82.3--85.1)} & 69.6 {\scriptsize (68.6--71.7)}     & 65.5 {\scriptsize (63.8--66.7)}          & 72.3 {\scriptsize (70.5--73.7)}                 & 53.9 {\scriptsize (51.8--56.0)}                 & 61.2 {\scriptsize (59.2--63.4)}     & 63.9 {\scriptsize (63.2--66.2)}             & 82.1 {\scriptsize (80.5--83.0)}             \\
PTB-XL Super                  & 74.4 {\scriptsize (73.4--75.4)}          & 67.4 {\scriptsize (67.0--68.7)}     & 64.0 {\scriptsize (63.3--64.8)}          & 64.9 {\scriptsize (64.1--66.1)}                 & 55.1 {\scriptsize (54.4--56.2)}                 & 55.9 {\scriptsize (55.0--57.1)}     & 55.3 {\scriptsize (54.5--56.6)}             & \textbf{78.6} {\scriptsize (77.7--79.5)}    \\
EchoNext                      & \textbf{73.6} {\scriptsize (72.6--75.1)} & 66.3 {\scriptsize (65.2--67.4)}     & 60.4 {\scriptsize (59.4--61.3)}          & 63.7 {\scriptsize (62.9--65.3)}                 & 54.0 {\scriptsize (53.4--55.2)}                 & 65.8 {\scriptsize (64.6--67.0)}     & 65.8 {\scriptsize (64.9--67.4)}             & 71.8 {\scriptsize (70.6--73.0)}             \\
CPSC2018                      & 85.9 {\scriptsize (83.4--87.2)}          & 85.6 {\scriptsize (84.4--86.9)}     & 82.3 {\scriptsize (80.9--83.6)}          & 74.7 {\scriptsize (72.8--77.7)}                 & 52.5 {\scriptsize (50.2--54.9)}                 & 66.4 {\scriptsize (64.7--68.1)}     & 78.8 {\scriptsize (77.7--82.3)}             & \textbf{90.0} {\scriptsize (88.7--91.2)}    \\
CPSC-Extra                    & 79.2 {\scriptsize (77.8--82.4)}          & 73.4 {\scriptsize (71.5--76.8)}     & 70.1 {\scriptsize (67.8--72.6)}          & 66.9 {\scriptsize (64.8--69.6)}                 & 56.6 {\scriptsize (53.3--59.2)}                 & 66.2 {\scriptsize (62.2--69.2)}     & 69.6 {\scriptsize (66.6--72.7)}             & \textbf{79.5} {\scriptsize (76.1--81.6)}    \\
Chapman                       & 85.9 {\scriptsize (84.6--87.9)}          & 80.6 {\scriptsize (79.0--81.8)}     & 68.5 {\scriptsize (67.2--70.3)}          & 72.9 {\scriptsize (71.3--75.1)}                 & 62.0 {\scriptsize (60.4--64.1)}                 & 65.7 {\scriptsize (63.8--67.4)}     & 72.9 {\scriptsize (71.3--75.0)}             & \textbf{86.1} {\scriptsize (84.9--87.5)}    \\
Chapman Rhythm$^\dagger$      & \textbf{95.4} {\scriptsize (94.3--96.3)} & 90.4 {\scriptsize (89.0--91.6)}     & 80.5 {\scriptsize (78.4--83.1)}          & 76.9 {\scriptsize (75.0--82.0)}                 & 69.8 {\scriptsize (67.3--71.9)}                 & 67.8 {\scriptsize (64.4--70.3)}     & 79.0 {\scriptsize (76.3--80.7)}             & \textbf{95.4} {\scriptsize (94.5--96.1)}    \\
Georgia                       & 80.3 {\scriptsize (78.9--82.8)}          & 67.6 {\scriptsize (65.2--69.2)}     & 67.8 {\scriptsize (66.5--69.4)}          & 68.2 {\scriptsize (67.0--69.7)}                 & 58.9 {\scriptsize (57.1--62.8)}                 & 55.4 {\scriptsize (53.7--57.8)}     & 68.4 {\scriptsize (66.9--71.3)}             & \textbf{80.4} {\scriptsize (78.2--81.7)}    \\
Hefei                         & 72.1 {\scriptsize (69.7--74.9)}          & 76.0 {\scriptsize (72.8--78.3)}     & 72.0 {\scriptsize (69.5--74.0)}          & 69.2 {\scriptsize (66.6--72.6)}                 & 60.3 {\scriptsize (58.7--63.5)}                 & 67.7 {\scriptsize (64.3--70.1)}     & 71.3 {\scriptsize (69.3--74.4)}             & \textbf{87.0} {\scriptsize (84.6--88.8)}    \\
Ningbo                        & 88.4 {\scriptsize (87.5--90.6)}          & 79.5 {\scriptsize (79.4--82.1)}     & 73.0 {\scriptsize (71.9--74.5)}          & 73.9 {\scriptsize (72.8--75.4)}                 & 67.2 {\scriptsize (66.3--69.1)}                 & 75.4 {\scriptsize (74.2--77.2)}     & 73.8 {\scriptsize (72.6--75.5)}             & \textbf{89.2} {\scriptsize (88.3--90.0)}    \\
PTB                           & 62.7 {\scriptsize (59.6--76.4)}          & 62.3 {\scriptsize (58.9--72.8)}     & \textbf{64.6} {\scriptsize (59.3--68.0)} & 56.2 {\scriptsize (52.2--66.4)}                 & 51.6 {\scriptsize (47.6--55.3)}                 & 62.8 {\scriptsize (60.7--71.6)}     & 56.8 {\scriptsize (55.3--66.9)}             & 62.9 {\scriptsize (55.8--68.7)}             \\
SPH                           & 84.4 {\scriptsize (82.6--86.3)}          & 84.0 {\scriptsize (82.4--85.5)}     & 71.8 {\scriptsize (70.3--73.5)}          & 72.9 {\scriptsize (71.2--75.4)}                 & 60.1 {\scriptsize (57.6--62.8)}                 & 66.5 {\scriptsize (64.5--69.4)}     & 75.1 {\scriptsize (73.0--77.9)}             & \textbf{88.0} {\scriptsize (86.6--89.3)}    \\
ZZU pECG                      & \textbf{84.7} {\scriptsize (82.3--86.2)} & 83.3 {\scriptsize (82.2--85.7)}     & 71.0 {\scriptsize (68.8--72.3)}          & 77.0 {\scriptsize (75.5--79.2)}                 & 69.8 {\scriptsize (67.6--72.4)}                 & 64.7 {\scriptsize (62.1--66.5)}     & 71.6 {\scriptsize (69.4--73.6)}             & 83.7 {\scriptsize (81.6--85.6)}             \\
MIMIC Cardiac                 & 66.3 {\scriptsize (65.4--67.1)}          & 65.4 {\scriptsize (64.6--66.2)}     & 62.8 {\scriptsize (61.8--63.8)}          & 64.7 {\scriptsize (63.6--65.6)}                 & 59.1 {\scriptsize (58.4--59.8)}                 & 63.1 {\scriptsize (62.1--63.9)}     & 62.7 {\scriptsize (61.9--63.6)}             & \textbf{71.2} {\scriptsize (70.3--72.0)}    \\
MIMIC Non-cardiac             & 57.3 {\scriptsize (56.9--57.8)}          & 58.1 {\scriptsize (57.6--58.5)}     & 58.4 {\scriptsize (57.8--58.9)}          & 59.2 {\scriptsize (58.6--59.8)}                 & 55.7 {\scriptsize (55.3--56.1)}                 & 56.5 {\scriptsize (56.0--56.9)}     & 56.1 {\scriptsize (55.7--56.5)}             & \textbf{63.7} {\scriptsize (63.1--64.2)}    \\
MIMIC Deterioration           & 62.1 {\scriptsize (58.4--66.0)}          & 58.1 {\scriptsize (54.9--61.2)}     & 64.3 {\scriptsize (59.3--69.3)}          & 59.2 {\scriptsize (54.7--63.2)}                 & 54.8 {\scriptsize (51.3--58.1)}                 & 62.3 {\scriptsize (59.1--65.5)}     & 59.6 {\scriptsize (56.4--63.3)}             & \textbf{70.1} {\scriptsize (66.7--73.2)}    \\
MIMIC Mortality               & 62.2 {\scriptsize (57.7--65.7)}          & 62.0 {\scriptsize (60.1--63.7)}     & 68.0 {\scriptsize (63.5--71.2)}          & 58.6 {\scriptsize (55.0--62.1)}                 & 56.9 {\scriptsize (54.4--59.3)}                 & 61.6 {\scriptsize (58.4--64.2)}     & 58.6 {\scriptsize (56.8--60.9)}             & \textbf{76.8} {\scriptsize (74.7--78.7)}    \\
MIMIC ICU admission           & 60.9 {\scriptsize (59.0--62.6)}          & 58.9 {\scriptsize (57.5--60.3)}     & 63.1 {\scriptsize (61.1--65.1)}          & 58.0 {\scriptsize (55.8--59.9)}                 & 54.4 {\scriptsize (52.7--56.0)}                 & 61.2 {\scriptsize (59.7--62.7)}     & 57.5 {\scriptsize (56.2--58.8)}             & \textbf{69.7} {\scriptsize (68.0--71.5)}    \\
MIMIC Sex                     & 69.2 {\scriptsize (67.7--70.7)}          & 62.1 {\scriptsize (61.1--63.0)}     & 63.6 {\scriptsize (61.9--65.1)}          & 66.1 {\scriptsize (65.1--67.2)}                 & 58.8 {\scriptsize (57.8--60.0)}                 & 58.6 {\scriptsize (56.9--60.2)}     & 62.2 {\scriptsize (60.9--63.6)}             & \textbf{77.3} {\scriptsize (75.9--78.5)}    \\
\midrule
MIMIC Age (MAE)               & 0.774 {\scriptsize (0.745--0.801)}       & 0.763 {\scriptsize (0.736--0.789)}  & 0.792 {\scriptsize (0.762--0.820)}       & 0.807 {\scriptsize (0.777--0.835)}              & 0.857 {\scriptsize (0.826--0.885)}              & 0.780 {\scriptsize (0.755--0.802)}  & 0.813 {\scriptsize (0.782--0.842)}          & \textbf{0.636} {\scriptsize (0.619--0.653)} \\
MIMIC Biometrics (MAE)        & 0.776 {\scriptsize (0.747--0.804)}*      & 0.794 {\scriptsize (0.766--0.823)}  & 0.777 {\scriptsize (0.750--0.803)}*      & 0.811 {\scriptsize (0.777--0.843)}              & 0.843 {\scriptsize (0.813--0.874)}              & 0.803 {\scriptsize (0.773--0.831)}  & \textbf{0.775} {\scriptsize (0.746--0.804)} & 0.798 {\scriptsize (0.777--0.818)}          \\
MIMIC ECG features (MAE)      & 0.611 {\scriptsize (0.598--0.624)}       & 0.691 {\scriptsize (0.677--0.704)}  & 0.733 {\scriptsize (0.718--0.747)}       & 0.675 {\scriptsize (0.660--0.689)}              & 0.708 {\scriptsize (0.694--0.721)}              & 0.698 {\scriptsize (0.684--0.712)}  & 0.685 {\scriptsize (0.670--0.698)}          & \textbf{0.590} {\scriptsize (0.578--0.602)} \\
MIMIC Laboratory values (MAE) & 0.786 {\scriptsize (0.770--0.830)}       & 0.685 {\scriptsize (0.668--0.717)}* & 0.930 {\scriptsize (0.739--0.966)}       & 0.739 {\scriptsize (0.720--0.756)}              & 0.828 {\scriptsize (0.814--0.875)}              & 0.682 {\scriptsize (0.655--0.714)}* & 0.781 {\scriptsize (0.756--0.796)}          & \textbf{0.677} {\scriptsize (0.646--0.708)} \\
MIMIC Vital signs (MAE)       & 0.787 {\scriptsize (0.776--0.796)}       & 0.749 {\scriptsize (0.739--0.757)}  & 0.746 {\scriptsize (0.736--0.755)}       & 0.764 {\scriptsize (0.753--0.773)}              & 0.791 {\scriptsize (0.781--0.799)}              & 0.754 {\scriptsize (0.743--0.763)}  & 0.766 {\scriptsize (0.755--0.775)}          & \textbf{0.721} {\scriptsize (0.712--0.730)} \\
\midrule
Mean AUROC (24 benchmarks) & 77.1 & 72.0 & 68.6 & 68.3 & 59.5 & 62.5 & 67.1 & \textbf{80.3} \\
Mean MAE (5 benchmarks) & 0.747 & 0.737 & 0.795 & 0.759 & 0.805 & 0.743 & 0.764 & \textbf{0.685} \\
\bottomrule
\end{tabular}}
\end{table*}

\begin{table*}[p]
\centering
\caption{\textbf{Per-condition performance on real point-of-care recordings at 90\% specificity.}
Zero-shot: models finetuned on 12-lead PTB-XL All, no retraining;
sensitivity and PPV are deliberately read from the ROC curve of the test set,
so they are the highest reachable at 90\% specificity rather than the result of
a deployable threshold (thresholds chosen on PTB-XL All: Supplementary Table~\ref{tab:supp_threshold_transfer}).
Conditions: all heads with a counterpart in each dataset, ordered by number of positives
($n_+$); Prev.: prevalence. Max.\ PPV: highest PPV reachable at 90\% specificity at this
prevalence, by a model with sensitivity 100\%, Prev./(Prev.\ + 0.1 $\times$ (100 $-$ Prev.));
it applies to this operating point and rises with stricter specificity.
CinC 2017: AF vs Normal records only, prevalence as observed in that subset.
Icentia11k: windows were sampled per class and patient, so prevalence is the rate over all valid
windows and PPV is derived from sensitivity and specificity at that rate;
mean over the 12 lead slots (ECG-JEPA: its 8 applicable slots), since the patch lead
has no standard identity. Brackets: 95\% bootstrap CI, 1000 resamples, thresholds re-derived.
Bold, underlined and italic values are the best, second-best and third-best of each row.}
\label{tab:supp_poc_thresholds}
\setlength{\tabcolsep}{2pt}
\renewcommand{\arraystretch}{0.9}
\resizebox{\textwidth}{!}{
\begin{tabular}{lrrrl cccccccc}
\toprule
Condition & $n_+$  & Prev. (\%) & Max.\ PPV (\%) & Metric & ECGFounder                                    & ECG-JEPA                                     & ST-MEM                                       & MERL                                   & ECGFM-KED                              & HuBERT-ECG                             & ECG-CPC                                      & \textbf{RELAY}                               \\
\midrule
\multicolumn{13}{l}{\textbf{CinC 2017 (AliveCor, lead I slot)}} \\
\midrule
SR        & 5,076  & 87.0       & 98.5           & Sens.  & \shortstack{\due{60.9} \\{\tiny[54.4, 63.0]}} & \shortstack{\tre{34.2}\\{\tiny[28.5, 42.4]}} & \shortstack{8.5\\{\tiny[6.5, 10.6]}}         & \shortstack{17.9\\{\tiny[12.7, 22.9]}} & \shortstack{3.9\\{\tiny[2.9, 5.4]}}    & \shortstack{19.5\\{\tiny[15.7, 25.6]}} & \shortstack{18.0\\{\tiny[13.6, 20.6]}}       & \shortstack{\uno{70.1}\\{\tiny[66.3, 73.6]}} \\
          &        &            &                & PPV    & \shortstack{\due{97.6} \\{\tiny[97.4, 97.7]}} & \shortstack{\tre{95.9}\\{\tiny[95.1, 96.6]}} & \shortstack{85.2\\{\tiny[81.4, 87.7]}}       & \shortstack{92.4\\{\tiny[89.6, 94.0]}} & \shortstack{72.4\\{\tiny[66.2, 78.6]}} & \shortstack{93.0\\{\tiny[91.4, 94.5]}} & \shortstack{92.4\\{\tiny[90.2, 93.4]}}       & \shortstack{\uno{97.9}\\{\tiny[97.8, 98.0]}} \\
\addlinespace
AFIB      & 758    & 13.0       & 59.9           & Sens.  & \shortstack{23.9       \\{\tiny[19.3, 28.2]}} & \shortstack{\uno{59.5}\\{\tiny[55.6, 63.2]}} & \shortstack{8.2\\{\tiny[5.7, 10.2]}}         & \shortstack{23.2\\{\tiny[19.9, 26.7]}} & \shortstack{9.0\\{\tiny[6.8, 10.9]}}   & \shortstack{9.5\\{\tiny[7.3, 11.7]}}   & \shortstack{\tre{26.3}\\{\tiny[23.0, 30.1]}} & \shortstack{\due{49.1}\\{\tiny[44.3, 53.3]}} \\
          &        &            &                & PPV    & \shortstack{26.3       \\{\tiny[22.4, 29.7]}} & \shortstack{\uno{47.1}\\{\tiny[45.4, 48.6]}} & \shortstack{10.9\\{\tiny[7.9, 13.3]}}        & \shortstack{25.8\\{\tiny[22.9, 28.6]}} & \shortstack{11.8\\{\tiny[9.2, 14.0]}}  & \shortstack{12.4\\{\tiny[9.8, 14.8]}}  & \shortstack{\tre{28.2}\\{\tiny[25.6, 31.0]}} & \shortstack{\due{42.3}\\{\tiny[39.9, 44.4]}} \\
\midrule
\multicolumn{13}{l}{\textbf{Icentia11k (patch, mean over lead slots)}} \\
\midrule
SR        & 56,179 & 93.6       & 99.3           & Sens.  & \shortstack{34.4       \\{\tiny[33.3, 35.7]}} & \shortstack{\tre{35.0}\\{\tiny[33.4, 36.3]}} & \shortstack{\due{42.8}\\{\tiny[41.7, 44.1]}} & \shortstack{10.8\\{\tiny[10.0, 11.7]}} & \shortstack{4.5\\{\tiny[3.9, 5.2]}}    & \shortstack{22.2\\{\tiny[20.3, 24.4]}} & \shortstack{13.8\\{\tiny[12.8, 14.7]}}       & \shortstack{\uno{70.7}\\{\tiny[68.5, 72.6]}} \\
          &        &            &                & PPV    & \shortstack{\tre{97.7} \\{\tiny[97.6, 97.8]}} & \shortstack{\due{97.9}\\{\tiny[97.8, 98.0]}} & \shortstack{97.3\\{\tiny[97.2, 97.4]}}       & \shortstack{92.6\\{\tiny[91.9, 93.3]}} & \shortstack{86.4\\{\tiny[84.4, 88.2]}} & \shortstack{97.0\\{\tiny[96.7, 97.2]}} & \shortstack{94.8\\{\tiny[94.4, 95.1]}}       & \shortstack{\uno{99.0}\\{\tiny[99.0, 99.1]}} \\
\addlinespace
PAC       & 20,837 & 5.1        & 34.8           & Sens.  & \shortstack{\tre{14.9} \\{\tiny[14.6, 15.2]}} & \shortstack{\due{15.6}\\{\tiny[15.2, 16.1]}} & \shortstack{\uno{16.8}\\{\tiny[16.5, 17.1]}} & \shortstack{10.9\\{\tiny[10.5, 11.3]}} & \shortstack{10.6\\{\tiny[10.4, 10.8]}} & \shortstack{10.9\\{\tiny[10.5, 11.2]}} & \shortstack{11.2\\{\tiny[10.8, 11.5]}}       & \shortstack{14.7\\{\tiny[14.1, 15.3]}}       \\
          &        &            &                & PPV    & \shortstack{\tre{7.3}  \\{\tiny[7.2, 7.5]}}   & \shortstack{\due{7.7}\\{\tiny[7.5, 7.9]}}    & \shortstack{\uno{8.1}\\{\tiny[8.0, 8.3]}}    & \shortstack{5.5\\{\tiny[5.3, 5.7]}}    & \shortstack{5.4\\{\tiny[5.3, 5.5]}}    & \shortstack{5.5\\{\tiny[5.3, 5.6]}}    & \shortstack{5.6\\{\tiny[5.4, 5.8]}}          & \shortstack{\tre{7.3}\\{\tiny[7.0, 7.5]}}    \\
\addlinespace
PVC       & 17,414 & 4.9        & 34.0           & Sens.  & \shortstack{13.7       \\{\tiny[13.2, 14.2]}} & \shortstack{\due{51.3}\\{\tiny[50.6, 52.1]}} & \shortstack{\tre{28.0}\\{\tiny[27.6, 28.3]}} & \shortstack{7.3\\{\tiny[7.0, 7.6]}}    & \shortstack{4.7\\{\tiny[4.6, 5.0]}}    & \shortstack{14.0\\{\tiny[13.6, 14.4]}} & \shortstack{7.0\\{\tiny[6.7, 7.3]}}          & \shortstack{\uno{62.0}\\{\tiny[61.0, 63.2]}} \\
          &        &            &                & PPV    & \shortstack{6.4        \\{\tiny[6.2, 6.7]}}   & \shortstack{\due{21.1}\\{\tiny[20.8, 21.3]}} & \shortstack{\tre{12.0}\\{\tiny[11.9, 12.2]}} & \shortstack{3.6\\{\tiny[3.5, 3.8]}}    & \shortstack{2.4\\{\tiny[2.3, 2.5]}}    & \shortstack{6.7\\{\tiny[6.6, 6.9]}}    & \shortstack{3.5\\{\tiny[3.4, 3.6]}}          & \shortstack{\uno{24.2}\\{\tiny[23.9, 24.6]}} \\
\addlinespace
AFIB      & 1,468  & 4.8        & 33.6           & Sens.  & \shortstack{34.7       \\{\tiny[32.6, 36.7]}} & \shortstack{\due{47.7}\\{\tiny[45.1, 50.1]}} & \shortstack{\tre{47.6}\\{\tiny[46.3, 48.7]}} & \shortstack{20.9\\{\tiny[19.4, 22.6]}} & \shortstack{11.5\\{\tiny[10.3, 12.9]}} & \shortstack{11.0\\{\tiny[9.4, 12.7]}}  & \shortstack{25.4\\{\tiny[23.8, 26.9]}}       & \shortstack{\uno{69.1}\\{\tiny[66.1, 71.9]}} \\
          &        &            &                & PPV    & \shortstack{14.7       \\{\tiny[13.9, 15.4]}} & \shortstack{\due{19.4}\\{\tiny[18.5, 20.1]}} & \shortstack{\tre{18.1}\\{\tiny[17.7, 18.4]}} & \shortstack{9.5\\{\tiny[8.8, 10.1]}}   & \shortstack{5.5\\{\tiny[4.9, 6.1]}}    & \shortstack{5.2\\{\tiny[4.5, 6.0]}}    & \shortstack{11.3\\{\tiny[10.7, 11.9]}}       & \shortstack{\uno{25.9}\\{\tiny[25.0, 26.6]}} \\
\addlinespace
AFLT      & 1,010  & 1.6        & 14.3           & Sens.  & \shortstack{26.0       \\{\tiny[23.8, 28.4]}} & \shortstack{\due{45.5}\\{\tiny[42.5, 48.3]}} & \shortstack{\tre{34.0}\\{\tiny[32.2, 35.8]}} & \shortstack{6.4\\{\tiny[5.3, 7.7]}}    & \shortstack{22.6\\{\tiny[20.1, 25.4]}} & \shortstack{17.2\\{\tiny[14.7, 19.7]}} & \shortstack{26.3\\{\tiny[23.6, 29.3]}}       & \shortstack{\uno{60.3}\\{\tiny[56.9, 63.9]}} \\
          &        &            &                & PPV    & \shortstack{4.1        \\{\tiny[3.8, 4.5]}}   & \shortstack{\due{7.1}\\{\tiny[6.7, 7.5]}}    & \shortstack{\tre{5.3}\\{\tiny[5.0, 5.5]}}    & \shortstack{1.1\\{\tiny[0.9, 1.3]}}    & \shortstack{3.6\\{\tiny[3.3, 4.1]}}    & \shortstack{2.8\\{\tiny[2.4, 3.2]}}    & \shortstack{4.2\\{\tiny[3.8, 4.7]}}          & \shortstack{\uno{9.1}\\{\tiny[8.7, 9.6]}}    \\
\bottomrule
\end{tabular}
}
\end{table*}

\begin{table*}[p]
\centering
\caption{\textbf{Transfer of the 90\% specificity operating point to real point-of-care recordings.} Sensitivity/specificity and balanced accuracy (BA), in \%, for AFIB and SR of the models finetuned on 12-lead PTB-XL All, with bootstrap 95\% CIs in brackets. Thresholds are deliberately read from the test ROC curve, as in Supplementary Table~\ref{tab:supp_poc_thresholds}, or chosen at 90\% specificity on the PTB-XL All validation fold with 12-lead or lead-I inputs and applied unchanged (Section~\ref{sec:methods-evaluation-protocol}). Icentia11k: means over lead slots; $^\ddagger$range of the specificity across slots. ST-MEM is at chance level on lead I of PTB-XL All, so its lead-I threshold is uninformative. The SR threshold rests on 513 negative validation records, so its specificity is below 90\% even on the PTB-XL All test set.}
\label{tab:supp_threshold_transfer}
\setlength{\tabcolsep}{3pt}
\resizebox{\textwidth}{!}{
\begin{tabular}{ll cccccccc}
\toprule
Test set                                      & Threshold chosen on                & ECGFounder                                                                                            & ECG-JEPA                                                                                             & ST-MEM                                                                                              & MERL                                                                                                & ECGFM-KED                                                                                           & HuBERT-ECG                                                                                          & ECG-CPC                                                                                             & \textbf{RELAY}                                                                                         \\
\midrule
\multicolumn{10}{l}{\textbf{AFIB}} \\
\multirow{2}{*}{PTB-XL All test (12 leads)}   & Test ROC curve                     & \shortstack{98.0/90.0 \\{\tiny[95.6, 100.0]/[90.0, 90.1]}\\{\scriptsize BA 94.0 {\tiny[92.8, 95.0]}}} & \shortstack{98.0/90.1\\{\tiny[95.5, 100.0]/[90.0, 90.2]}\\{\scriptsize BA 94.1 {\tiny[92.7, 95.0]}}} & \shortstack{96.7/90.0\\{\tiny[93.6, 99.3]/[90.0, 90.2]}\\{\scriptsize BA 93.4 {\tiny[91.8, 94.7]}}} & \shortstack{94.1/90.0\\{\tiny[89.5, 97.6]/[90.0, 90.2]}\\{\scriptsize BA 92.1 {\tiny[89.8, 93.8]}}} & \shortstack{94.7/90.1\\{\tiny[90.1, 97.8]/[90.0, 90.2]}\\{\scriptsize BA 92.4 {\tiny[90.1, 94.0]}}} & \shortstack{96.1/90.0\\{\tiny[92.4, 98.8]/[90.0, 90.2]}\\{\scriptsize BA 93.0 {\tiny[91.2, 94.4]}}} & \shortstack{96.7/90.0\\{\tiny[93.6, 99.4]/[90.0, 90.2]}\\{\scriptsize BA 93.4 {\tiny[91.8, 94.7]}}} & \shortstack{95.4/90.0\\{\tiny[91.6, 98.6]/[90.0, 90.2]}\\{\scriptsize BA 92.7 {\tiny[90.8, 94.3]}}}    \\
                                              & PTB-XL, 12 leads                   & \shortstack{98.0/90.0 \\{\tiny[95.6, 100.0]/[88.6, 91.3]}\\{\scriptsize BA 94.0 {\tiny[92.6, 95.2]}}} & \shortstack{98.0/90.2\\{\tiny[95.5, 100.0]/[88.8, 91.5]}\\{\scriptsize BA 94.1 {\tiny[92.7, 95.3]}}} & \shortstack{96.7/90.8\\{\tiny[93.6, 99.3]/[89.5, 92.1]}\\{\scriptsize BA 93.8 {\tiny[92.1, 95.2]}}} & \shortstack{94.1/90.0\\{\tiny[89.9, 97.5]/[88.7, 91.3]}\\{\scriptsize BA 92.1 {\tiny[90.0, 93.8]}}} & \shortstack{94.7/89.7\\{\tiny[90.9, 98.0]/[88.3, 91.0]}\\{\scriptsize BA 92.2 {\tiny[90.1, 94.0]}}} & \shortstack{96.1/89.4\\{\tiny[92.4, 98.8]/[87.9, 90.7]}\\{\scriptsize BA 92.7 {\tiny[90.7, 94.3]}}} & \shortstack{97.4/89.1\\{\tiny[94.5, 99.5]/[87.7, 90.4]}\\{\scriptsize BA 93.2 {\tiny[91.6, 94.6]}}} & \shortstack{95.4/90.9\\{\tiny[91.7, 98.6]/[89.6, 92.2]}\\{\scriptsize BA 93.2 {\tiny[91.1, 94.8]}}}    \\
\addlinespace
\multirow{3}{*}{CinC 2017 (lead I slot)}      & Test ROC curve                     & \shortstack{23.9/90.0 \\{\tiny[19.3, 28.2]/[90.0, 90.1]}\\{\scriptsize BA 57.0 {\tiny[54.7, 59.1]}}}  & \shortstack{59.5/90.0\\{\tiny[55.6, 63.2]/[90.0, 90.1]}\\{\scriptsize BA 74.8 {\tiny[72.8, 76.6]}}}  & \shortstack{8.2/90.0\\{\tiny[5.7, 10.2]/[90.0, 90.1]}\\{\scriptsize BA 49.1 {\tiny[47.9, 50.1]}}}   & \shortstack{23.2/90.0\\{\tiny[19.9, 26.7]/[90.0, 90.1]}\\{\scriptsize BA 56.6 {\tiny[55.0, 58.4]}}} & \shortstack{9.0/90.0\\{\tiny[6.8, 10.9]/[90.0, 90.1]}\\{\scriptsize BA 49.5 {\tiny[48.4, 50.5]}}}   & \shortstack{9.5/90.0\\{\tiny[7.3, 11.7]/[90.0, 90.1]}\\{\scriptsize BA 49.8 {\tiny[48.7, 50.8]}}}   & \shortstack{26.3/90.0\\{\tiny[23.0, 30.1]/[90.0, 90.1]}\\{\scriptsize BA 58.1 {\tiny[56.5, 60.0]}}} & \shortstack{49.1/90.0\\{\tiny[44.3, 53.3]/[90.0, 90.1]}\\{\scriptsize BA 69.5 {\tiny[67.2, 71.7]}}}    \\
                                              & PTB-XL, 12 leads                   & \shortstack{100.0/0.9 \\{\tiny[100.0, 100.0]/[0.6, 1.1]}\\{\scriptsize BA 50.4 {\tiny[50.3, 50.6]}}}  & \shortstack{100.0/0.0\\{\tiny[100.0, 100.0]/[0.0, 0.0]}\\{\scriptsize BA 50.0 {\tiny[50.0, 50.0]}}}  & \shortstack{7.5/90.2\\{\tiny[5.6, 9.6]/[89.4, 91.0]}\\{\scriptsize BA 48.9 {\tiny[47.8, 50.0]}}}    & \shortstack{97.2/3.4\\{\tiny[96.1, 98.3]/[2.9, 3.9]}\\{\scriptsize BA 50.3 {\tiny[49.7, 50.9]}}}    & \shortstack{11.0/86.2\\{\tiny[8.7, 13.3]/[85.3, 87.2]}\\{\scriptsize BA 48.6 {\tiny[47.4, 49.8]}}}  & \shortstack{3.6/95.6\\{\tiny[2.4, 4.9]/[95.1, 96.2]}\\{\scriptsize BA 49.6 {\tiny[48.9, 50.3]}}}    & \shortstack{98.4/1.7\\{\tiny[97.4, 99.2]/[1.4, 2.1]}\\{\scriptsize BA 50.1 {\tiny[49.6, 50.5]}}}    & \shortstack{98.7/41.4\\{\tiny[97.8, 99.5]/[40.1, 42.9]}\\{\scriptsize BA 70.1 {\tiny[69.3, 70.8]}}}    \\
                                              & PTB-XL, lead I                     & \shortstack{65.3/75.7 \\{\tiny[61.8, 68.6]/[74.5, 76.7]}\\{\scriptsize BA 70.5 {\tiny[68.7, 72.2]}}}  & \shortstack{78.9/81.0\\{\tiny[76.0, 81.8]/[79.8, 81.9]}\\{\scriptsize BA 79.9 {\tiny[78.4, 81.5]}}}  & \shortstack{0.4/99.5\\{\tiny[0.0, 0.9]/[99.3, 99.7]}\\{\scriptsize BA 50.0 {\tiny[49.7, 50.2]}}}    & \shortstack{33.9/82.1\\{\tiny[30.5, 37.4]/[80.9, 83.1]}\\{\scriptsize BA 58.0 {\tiny[56.2, 59.8]}}} & \shortstack{5.0/93.8\\{\tiny[3.4, 6.7]/[93.1, 94.5]}\\{\scriptsize BA 49.4 {\tiny[48.5, 50.3]}}}    & \shortstack{5.7/93.2\\{\tiny[4.1, 7.5]/[92.5, 93.9]}\\{\scriptsize BA 49.5 {\tiny[48.6, 50.4]}}}    & \shortstack{8.8/98.0\\{\tiny[6.8, 11.0]/[97.6, 98.4]}\\{\scriptsize BA 53.4 {\tiny[52.4, 54.6]}}}   & \shortstack{97.6/49.8\\{\tiny[96.5, 98.7]/[48.5, 51.2]}\\{\scriptsize BA 73.7 {\tiny[72.9, 74.7]}}}    \\
\addlinespace
\multirow{3}{*}{Icentia11k (mean over slots)} & Test ROC curve                     & \shortstack{34.7/90.0 \\{\tiny[32.6, 36.7]/[90.0, 90.0]}\\{\scriptsize BA 62.4 {\tiny[61.3, 63.4]}}}  & \shortstack{47.7/90.0\\{\tiny[45.1, 50.1]/[90.0, 90.0]}\\{\scriptsize BA 68.8 {\tiny[67.6, 70.0]}}}  & \shortstack{47.6/90.0\\{\tiny[46.3, 48.7]/[90.0, 90.0]}\\{\scriptsize BA 68.8 {\tiny[68.2, 69.4]}}} & \shortstack{20.9/90.0\\{\tiny[19.4, 22.6]/[90.0, 90.0]}\\{\scriptsize BA 55.5 {\tiny[54.7, 56.3]}}} & \shortstack{11.5/90.0\\{\tiny[10.3, 12.9]/[90.0, 90.0]}\\{\scriptsize BA 50.8 {\tiny[50.2, 51.5]}}} & \shortstack{11.0/90.0\\{\tiny[9.4, 12.7]/[90.0, 90.0]}\\{\scriptsize BA 50.5 {\tiny[49.7, 51.4]}}}  & \shortstack{25.4/90.0\\{\tiny[23.8, 27.0]/[90.0, 90.0]}\\{\scriptsize BA 57.7 {\tiny[56.9, 58.5]}}} & \shortstack{69.1/90.0\\{\tiny[66.1, 71.9]/[90.0, 90.0]}\\{\scriptsize BA 79.6 {\tiny[78.0, 81.0]}}}    \\
                                              & PTB-XL, 12 leads                   & \shortstack{100.0/0.2 \\{\tiny[100.0, 100.0]/[0.1, 0.2]}\\{\scriptsize BA 50.1 {\tiny[50.1, 50.1]}}}  & \shortstack{100.0/0.0\\{\tiny[100.0, 100.0]/[0.0, 0.0]}\\{\scriptsize BA 50.0 {\tiny[50.0, 50.0]}}}  & \shortstack{39.0/91.7\\{\tiny[38.0, 40.0]/[91.6, 91.9]}\\{\scriptsize BA 65.4 {\tiny[64.9, 65.9]}}} & \shortstack{95.6/6.2\\{\tiny[95.1, 96.2]/[6.0, 6.3]}\\{\scriptsize BA 50.9 {\tiny[50.6, 51.2]}}}    & \shortstack{77.0/20.7\\{\tiny[75.6, 78.5]/[20.4, 21.0]}\\{\scriptsize BA 48.9 {\tiny[48.1, 49.6]}}} & \shortstack{31.7/70.3\\{\tiny[30.2, 33.2]/[69.9, 70.7]}\\{\scriptsize BA 51.0 {\tiny[50.2, 51.8]}}} & \shortstack{99.8/0.2\\{\tiny[99.7, 99.9]/[0.2, 0.2]}\\{\scriptsize BA 50.0 {\tiny[50.0, 50.1]}}}    & \shortstack{94.1/72.9\\{\tiny[92.7, 95.5]/[72.4, 73.5]}\\{\scriptsize BA 83.5 {\tiny[82.8, 84.2]}}}    \\
                                              & Spec.\ range over slots$^\ddagger$ & 0.0--0.7                                                                                              & 0.0                                                                                                  & 82.7--99.9                                                                                          & 0.0--18.6                                                                                           & 1.5--68.7                                                                                           & 7.0--94.9                                                                                           & 0.0--1.8                                                                                            & 72.9                                                                                                   \\
\addlinespace
\midrule
\multicolumn{10}{l}{\textbf{SR}} \\
\multirow{2}{*}{PTB-XL All test (12 leads)}   & Test ROC curve                     & \shortstack{63.9/90.1 \\{\tiny[52.8, 72.1]/[90.0, 90.6]}\\{\scriptsize BA 77.0 {\tiny[71.6, 81.1]}}}  & \shortstack{69.2/90.1\\{\tiny[60.5, 76.4]/[90.0, 90.6]}\\{\scriptsize BA 79.7 {\tiny[75.4, 83.3]}}}  & \shortstack{56.9/90.1\\{\tiny[48.8, 61.8]/[90.0, 90.6]}\\{\scriptsize BA 73.5 {\tiny[69.6, 76.1]}}} & \shortstack{50.8/90.1\\{\tiny[44.5, 56.0]/[90.0, 90.6]}\\{\scriptsize BA 70.5 {\tiny[67.4, 73.1]}}} & \shortstack{45.8/90.1\\{\tiny[37.1, 55.2]/[90.0, 90.6]}\\{\scriptsize BA 67.9 {\tiny[63.7, 72.6]}}} & \shortstack{68.1/90.1\\{\tiny[55.5, 73.2]/[90.0, 90.6]}\\{\scriptsize BA 79.1 {\tiny[72.9, 81.7]}}} & \shortstack{64.3/90.1\\{\tiny[55.7, 71.5]/[90.0, 90.6]}\\{\scriptsize BA 77.2 {\tiny[72.9, 80.8]}}} & \shortstack{49.3/90.1\\{\tiny[37.5, 59.1]/[90.0, 90.6]}\\{\scriptsize BA 69.7 {\tiny[63.9, 74.6]}}}    \\
                                              & PTB-XL, 12 leads                   & \shortstack{76.5/84.9 \\{\tiny[74.6, 78.7]/[81.6, 88.0]}\\{\scriptsize BA 80.7 {\tiny[78.9, 82.7]}}}  & \shortstack{76.5/87.0\\{\tiny[74.5, 78.6]/[84.0, 89.9]}\\{\scriptsize BA 81.8 {\tiny[80.0, 83.6]}}}  & \shortstack{67.7/84.7\\{\tiny[65.4, 69.8]/[81.2, 87.8]}\\{\scriptsize BA 76.2 {\tiny[74.2, 78.1]}}} & \shortstack{59.6/85.7\\{\tiny[57.3, 61.8]/[82.3, 88.5]}\\{\scriptsize BA 72.7 {\tiny[70.7, 74.5]}}} & \shortstack{58.2/86.1\\{\tiny[55.7, 60.4]/[82.8, 89.1]}\\{\scriptsize BA 72.1 {\tiny[70.1, 74.0]}}} & \shortstack{78.6/83.2\\{\tiny[76.5, 80.6]/[80.0, 86.4]}\\{\scriptsize BA 80.9 {\tiny[78.9, 82.7]}}} & \shortstack{75.5/85.3\\{\tiny[73.5, 77.6]/[82.1, 88.3]}\\{\scriptsize BA 80.4 {\tiny[78.5, 82.3]}}} & \shortstack{61.6/85.9\\{\tiny[59.1, 63.9]/[83.0, 88.9]}\\{\scriptsize BA 73.7 {\tiny[71.7, 75.7]}}}    \\
\addlinespace
\multirow{3}{*}{CinC 2017 (lead I slot)}      & Test ROC curve                     & \shortstack{60.9/90.1 \\{\tiny[54.4, 63.0]/[90.0, 90.4]}\\{\scriptsize BA 75.5 {\tiny[72.3, 76.6]}}}  & \shortstack{34.2/90.1\\{\tiny[28.5, 42.4]/[90.0, 90.4]}\\{\scriptsize BA 62.2 {\tiny[59.4, 66.3]}}}  & \shortstack{8.5/90.1\\{\tiny[6.5, 10.6]/[90.0, 90.4]}\\{\scriptsize BA 49.3 {\tiny[48.3, 50.3]}}}   & \shortstack{17.9/90.1\\{\tiny[12.7, 23.0]/[90.0, 90.4]}\\{\scriptsize BA 54.0 {\tiny[51.4, 56.5]}}} & \shortstack{3.9/90.1\\{\tiny[2.9, 5.4]/[90.0, 90.4]}\\{\scriptsize BA 47.0 {\tiny[46.5, 47.8]}}}    & \shortstack{19.5/90.1\\{\tiny[15.7, 25.6]/[90.0, 90.5]}\\{\scriptsize BA 54.8 {\tiny[53.0, 57.9]}}} & \shortstack{18.0/90.1\\{\tiny[13.6, 20.6]/[90.0, 90.4]}\\{\scriptsize BA 54.1 {\tiny[51.8, 55.4]}}} & \shortstack{70.1/90.1\\{\tiny[66.3, 73.6]/[90.0, 90.4]}\\{\scriptsize BA 80.1 {\tiny[78.3, 81.9]}}}    \\
                                              & PTB-XL, 12 leads                   & \shortstack{0.0/100.0 \\{\tiny[0.0, 0.1]/[100.0, 100.0]}\\{\scriptsize BA 50.0 {\tiny[50.0, 50.1]}}}  & \shortstack{0.0/100.0\\{\tiny[0.0, 0.0]/[100.0, 100.0]}\\{\scriptsize BA 50.0 {\tiny[50.0, 50.0]}}}  & \shortstack{0.0/100.0\\{\tiny[0.0, 0.0]/[100.0, 100.0]}\\{\scriptsize BA 50.0 {\tiny[50.0, 50.0]}}} & \shortstack{0.0/100.0\\{\tiny[0.0, 0.0]/[100.0, 100.0]}\\{\scriptsize BA 50.0 {\tiny[50.0, 50.0]}}} & \shortstack{4.2/89.5\\{\tiny[3.7, 4.8]/[87.3, 91.6]}\\{\scriptsize BA 46.8 {\tiny[45.7, 47.9]}}}    & \shortstack{0.0/100.0\\{\tiny[0.0, 0.0]/[100.0, 100.0]}\\{\scriptsize BA 50.0 {\tiny[50.0, 50.0]}}} & \shortstack{0.0/100.0\\{\tiny[0.0, 0.0]/[100.0, 100.0]}\\{\scriptsize BA 50.0 {\tiny[50.0, 50.0]}}} & \shortstack{24.0/99.5\\{\tiny[22.8, 25.2]/[98.9, 99.9]}\\{\scriptsize BA 61.7 {\tiny[61.1, 62.4]}}}    \\
                                              & PTB-XL, lead I                     & \shortstack{31.2/99.1 \\{\tiny[30.0, 32.5]/[98.3, 99.7]}\\{\scriptsize BA 65.1 {\tiny[64.4, 65.9]}}}  & \shortstack{18.1/94.7\\{\tiny[17.0, 19.1]/[93.1, 96.3]}\\{\scriptsize BA 56.4 {\tiny[55.4, 57.3]}}}  & \shortstack{1.7/96.4\\{\tiny[1.3, 2.1]/[95.1, 97.7]}\\{\scriptsize BA 49.1 {\tiny[48.4, 49.7]}}}    & \shortstack{42.5/75.5\\{\tiny[41.1, 43.8]/[72.3, 78.5]}\\{\scriptsize BA 59.0 {\tiny[57.4, 60.6]}}} & \shortstack{8.4/83.1\\{\tiny[7.7, 9.1]/[80.4, 85.8]}\\{\scriptsize BA 45.8 {\tiny[44.4, 47.1]}}}    & \shortstack{8.5/94.9\\{\tiny[7.8, 9.3]/[93.3, 96.4]}\\{\scriptsize BA 51.7 {\tiny[50.8, 52.6]}}}    & \shortstack{37.2/73.8\\{\tiny[35.8, 38.6]/[70.6, 76.8]}\\{\scriptsize BA 55.5 {\tiny[53.7, 57.2]}}} & \shortstack{14.6/100.0\\{\tiny[13.5, 15.6]/[100.0, 100.0]}\\{\scriptsize BA 57.3 {\tiny[56.8, 57.8]}}} \\
\addlinespace
\multirow{3}{*}{Icentia11k (mean over slots)} & Test ROC curve                     & \shortstack{34.4/90.0 \\{\tiny[33.3, 35.7]/[90.0, 90.1]}\\{\scriptsize BA 62.2 {\tiny[61.7, 62.8]}}}  & \shortstack{35.0/90.1\\{\tiny[33.4, 36.3]/[90.1, 90.2]}\\{\scriptsize BA 62.5 {\tiny[61.8, 63.2]}}}  & \shortstack{42.8/90.0\\{\tiny[41.7, 44.2]/[90.0, 90.1]}\\{\scriptsize BA 66.4 {\tiny[65.9, 67.1]}}} & \shortstack{10.8/90.0\\{\tiny[10.0, 11.7]/[90.0, 90.1]}\\{\scriptsize BA 50.4 {\tiny[50.0, 50.9]}}} & \shortstack{4.5/90.0\\{\tiny[3.9, 5.2]/[90.0, 90.1]}\\{\scriptsize BA 47.3 {\tiny[47.0, 47.6]}}}    & \shortstack{22.3/90.0\\{\tiny[20.3, 24.4]/[90.0, 90.1]}\\{\scriptsize BA 56.1 {\tiny[55.2, 57.2]}}} & \shortstack{13.8/90.0\\{\tiny[12.8, 14.7]/[90.0, 90.1]}\\{\scriptsize BA 51.9 {\tiny[51.4, 52.4]}}} & \shortstack{70.7/90.0\\{\tiny[68.5, 72.6]/[90.0, 90.1]}\\{\scriptsize BA 80.4 {\tiny[79.3, 81.3]}}}    \\
                                              & PTB-XL, 12 leads                   & \shortstack{0.2/100.0 \\{\tiny[0.1, 0.2]/[100.0, 100.0]}\\{\scriptsize BA 50.1 {\tiny[50.1, 50.1]}}}  & \shortstack{0.0/100.0\\{\tiny[0.0, 0.0]/[100.0, 100.0]}\\{\scriptsize BA 50.0 {\tiny[50.0, 50.0]}}}  & \shortstack{7.6/99.9\\{\tiny[7.4, 7.7]/[99.8, 99.9]}\\{\scriptsize BA 53.7 {\tiny[53.7, 53.8]}}}    & \shortstack{0.0/100.0\\{\tiny[0.0, 0.0]/[100.0, 100.0]}\\{\scriptsize BA 50.0 {\tiny[50.0, 50.0]}}} & \shortstack{1.3/96.8\\{\tiny[1.2, 1.4]/[96.3, 97.2]}\\{\scriptsize BA 49.0 {\tiny[48.8, 49.3]}}}    & \shortstack{0.0/100.0\\{\tiny[0.0, 0.0]/[100.0, 100.0]}\\{\scriptsize BA 50.0 {\tiny[50.0, 50.0]}}} & \shortstack{0.0/100.0\\{\tiny[0.0, 0.0]/[100.0, 100.0]}\\{\scriptsize BA 50.0 {\tiny[50.0, 50.0]}}} & \shortstack{45.5/98.7\\{\tiny[44.9, 46.1]/[98.2, 99.1]}\\{\scriptsize BA 72.1 {\tiny[71.7, 72.5]}}}    \\
                                              & Spec.\ range over slots$^\ddagger$ & 100.0                                                                                                 & 100.0                                                                                                & 99.5--100.0                                                                                         & 100.0                                                                                               & 91.2--99.9                                                                                          & 100.0                                                                                               & 100.0                                                                                               & 98.7                                                                                                   \\
\bottomrule
\end{tabular}
}
\end{table*}

\begin{table*}[p]
\centering
\caption{\textbf{Sensitivity at 90\% specificity.} For every condition, sensitivity is deliberately read from the ROC curve of the test set at the threshold giving a specificity above 90\%, then the macro average over conditions $\times 100$ is reported (conditions with no positive test sample are left out). Like AUROC, these values summarise the ROC curve: they are the highest reachable by any threshold at this operating point, not the result of a deployable threshold (Supplementary Table~\ref{tab:supp_threshold_transfer}). Achieved: specificity actually reached by the thresholds, averaged over conditions and models. When leads are sampled randomly, mean(std) over 5 seeds is reported. ST-MEM runs on the available leads only, with their true lead embeddings. *: not significantly worse than the best model of the row (paired bootstrap of the difference, 1000 resamples with thresholds re-derived, 95\% CI includes 0). Bold, underlined and italic values are the best, second-best and third-best of each row. The MIMIC-IV-ECG benchmarks are not included. $^\dagger$Evaluation-only benchmark.}
\label{tab:supp_sens90spec}
\setlength{\tabcolsep}{3pt}
\renewcommand{\arraystretch}{0.9}
\resizebox{\textwidth}{!}{
\begin{tabular}{llc cccccccc}
\toprule
Benchmark & Input      & Achieved & ECGFounder       & ECG-JEPA        & ST-MEM          & MERL        & ECGFM-KED   & HuBERT-ECG & ECG-CPC         & \textbf{RELAY}   \\
\midrule
\multirow{5}{*}{CODE 15\%}
          & 12 leads   & 90.0     & \due{98.6}*      & \uno{99.2}      & \uno{99.2}*     & \tre{97.8}* & 97.1*       & 97.1*      & \uno{99.2}*     & \uno{99.2}*      \\
          & Lead I     & 90.1     & \due{84.6}*      & \tre{84.3}*     & 13.0            & 40.0        & 21.9        & 20.0       & 11.6            & \uno{89.0}       \\
          & Lead II    & 90.1     & \tre{81.0}       & \due{85.8}*     & 77.1            & 35.0        & 27.2        & 21.8       & 13.9            & \uno{91.9}       \\
          & One random & 90.0     & \due{86.3(2.7)}* & \tre{55.8(6.9)} & 45.8(4.2)       & 52.1(1.2)   & 21.6(2.4)   & 14.5(2.6)  & 18.3(3.9)       & \uno{89.9(2.3)}  \\
          & Two random & 90.0     & \due{94.7(1.3)}* & \tre{73.1(2.0)} & 57.5(5.4)       & 68.8(1.2)   & 42.5(2.3)   & 26.8(4.3)  & 28.8(3.1)       & \uno{97.3(1.0)}  \\
\midrule
\multirow{5}{*}{PTB-XL All}
          & 12 leads   & 90.0     & \due{83.6}*      & \tre{82.8}*     & 75.5            & 82.0        & 72.0        & 75.7       & \uno{84.9}      & 74.0             \\
          & Lead I     & 90.0     & \tre{27.4}       & \uno{45.2}      & 12.2            & 17.6        & 13.8        & 13.7       & 13.2            & \due{40.4}*      \\
          & Lead II    & 90.1     & 27.1             & \due{44.3}*     & \tre{41.6}      & 13.5        & 16.3        & 15.8       & 18.8            & \uno{46.4}       \\
          & One random & 90.1     & \due{31.6(1.7)}  & \tre{25.3(2.0)} & 22.9(2.2)       & 18.8(1.9)   & 13.9(1.5)   & 12.8(0.7)  & 20.8(1.0)       & \uno{38.2(1.8)}  \\
          & Two random & 90.1     & \uno{50.0(1.8)}  & \due{34.9(3.3)} & 26.7(1.1)       & 29.2(1.6)   & 20.9(2.1)   & 19.6(1.2)  & \tre{30.4(1.6)} & \uno{50.0(0.4)}* \\
\midrule
\multirow{5}{*}{PTB-XL Diag$^\dagger$}
          & 12 leads   & 90.0     & \uno{87.0}       & 82.3            & 74.2            & \due{86.1}* & 77.5        & 78.7       & \tre{85.9}*     & 78.5             \\
          & Lead I     & 90.0     & \tre{24.1}       & \uno{44.3}      & 14.3            & 19.6        & 15.6        & 11.3       & 12.3            & \due{40.0}*      \\
          & Lead II    & 90.0     & 23.9             & \due{39.4}      & \tre{38.4}      & 15.7        & 21.1        & 14.2       & 18.9            & \uno{44.5}       \\
          & One random & 90.0     & \due{28.5(2.8)}  & \tre{25.7(2.1)} & 21.7(1.8)       & 20.3(3.1)   & 14.8(1.9)   & 11.3(1.0)  & 19.2(1.2)       & \uno{35.1(1.4)}  \\
          & Two random & 90.0     & \due{49.8(2.0)}* & \tre{34.6(4.3)} & 25.1(0.7)       & 32.1(1.7)   & 20.0(2.9)   & 17.0(1.2)  & 29.6(1.1)       & \uno{50.4(1.3)}  \\
\midrule
\multirow{5}{*}{PTB-XL Form$^\dagger$}
          & 12 leads   & 90.0     & 72.1             & \due{76.6}*     & 69.6            & \tre{72.9}  & 60.3        & 58.4       & \uno{78.4}      & 61.9             \\
          & Lead I     & 90.0     & \tre{17.8}       & \uno{38.0}      & 11.9            & 10.8        & 9.7         & 16.3       & 11.2            & \due{35.9}*      \\
          & Lead II    & 90.0     & 15.1             & \due{36.7}*     & \tre{32.3}      & 10.7        & 10.3        & 15.5       & 11.1            & \uno{40.5}       \\
          & One random & 90.0     & \due{21.3(2.3)}  & \tre{21.1(2.2)} & 17.3(1.2)       & 13.9(2.4)   & 11.5(1.2)   & 10.6(1.3)  & 17.4(1.6)       & \uno{36.4(3.4)}  \\
          & Two random & 90.0     & \due{33.8(4.1)}  & \tre{27.6(2.9)} & 21.7(0.9)       & 18.1(2.3)   & 16.8(2.8)   & 16.6(2.6)  & 21.3(1.6)       & \uno{39.7(3.5)}  \\
\midrule
\multirow{5}{*}{PTB-XL Rhythm$^\dagger$}
          & 12 leads   & 90.0     & 87.2             & \uno{92.0}      & \tre{87.3}      & 78.6        & 69.8        & \tre{87.3} & \due{88.7}      & 73.5             \\
          & Lead I     & 90.1     & \due{52.1}*      & \uno{61.7}      & 6.9             & 18.2        & 13.6        & 17.4       & 20.6            & \tre{51.2}*      \\
          & Lead II    & 90.1     & 54.7             & \uno{74.9}      & \due{66.7}      & 11.0        & 8.9         & 24.1       & 27.4            & \tre{59.5}       \\
          & One random & 90.3     & \uno{57.0(2.1)}  & 32.6(4.8)       & \tre{36.0(6.2)} & 20.3(3.0)   & 13.3(3.9)   & 21.5(5.9)  & 30.3(2.8)       & \due{52.7(4.8)}* \\
          & Two random & 90.2     & \uno{71.8(2.4)}  & \tre{48.6(3.5)} & 38.9(4.1)       & 34.3(6.7)   & 29.6(5.5)   & 32.7(5.3)  & 44.8(5.0)       & \due{64.3(3.9)}* \\
\midrule
\multirow{5}{*}{PTB-XL Sub}
          & 12 leads   & 90.0     & \tre{83.2}*      & 82.7*           & 76.7            & \due{83.9}* & 71.7        & 76.7       & \uno{84.7}      & 79.2             \\
          & Lead I     & 90.0     & \tre{28.5}       & \due{41.8}*     & 9.3             & 15.0        & 9.2         & 15.2       & 9.5             & \uno{45.4}       \\
          & Lead II    & 90.0     & 28.3             & \due{46.4}*     & \tre{38.4}      & 16.7        & 11.3        & 13.8       & 11.5            & \uno{47.7}       \\
          & One random & 90.0     & \due{34.0(3.2)}  & \tre{28.7(4.7)} & 21.7(3.1)       & 20.0(2.3)   & 11.7(2.2)   & 12.7(1.2)  & 18.3(1.9)       & \uno{38.2(0.6)}  \\
          & Two random & 90.0     & \uno{55.6(2.7)}  & \tre{36.1(3.8)} & 27.3(1.1)       & 31.7(3.3)   & 14.1(3.0)   & 23.0(1.7)  & 26.4(2.5)       & \due{52.9(3.3)}* \\
\midrule
\multirow{5}{*}{PTB-XL Super}
          & 12 leads   & 90.0     & \uno{80.7}       & 78.1            & 69.3            & \tre{79.6}* & 71.3        & 74.8       & \due{80.6}*     & 75.3             \\
          & Lead I     & 90.0     & \tre{16.8}       & \due{34.0}      & 9.9             & 9.6         & 11.1        & 13.6       & 9.1             & \uno{45.6}       \\
          & Lead II    & 90.0     & 22.1             & \due{42.9}*     & \tre{29.5}      & 13.4        & 11.4        & 14.1       & 13.0            & \uno{43.3}       \\
          & One random & 90.0     & \due{23.8(0.8)}  & \tre{23.1(1.1)} & 15.6(1.1)       & 17.0(0.8)   & 12.0(1.2)   & 9.7(0.9)   & 11.9(0.3)       & \uno{35.3(1.2)}  \\
          & Two random & 90.0     & \due{36.6(0.6)}  & \tre{32.1(0.4)} & 21.9(1.3)       & 26.8(0.7)   & 14.8(1.3)   & 14.3(1.2)  & 16.1(0.9)       & \uno{44.9(1.9)}  \\
\midrule
\multirow{5}{*}{EchoNext}
          & 12 leads   & 90.0     & 47.0             & \tre{49.4}      & \tre{49.4}      & \due{49.6}  & 48.9        & 45.1       & \uno{52.1}      & 44.2             \\
          & Lead I     & 90.0     & \due{29.1}       & \uno{33.3}      & 10.8            & 18.1        & 16.5        & 21.0       & 20.6            & \tre{25.5}       \\
          & Lead II    & 90.0     & \uno{29.8}       & \tre{27.5}*     & 21.7            & 17.9        & 13.2        & 20.1       & 14.0            & \due{28.7}*      \\
          & One random & 90.0     & \uno{28.1(0.4)}  & \tre{21.3(1.0)} & 15.0(0.5)       & 16.4(0.9)   & 13.4(0.6)   & 16.8(1.1)  & 15.9(1.1)       & \due{25.5(1.1)}  \\
          & Two random & 90.0     & \uno{34.5(1.2)}  & \tre{28.1(1.2)} & 18.1(1.2)       & 23.8(0.9)   & 14.1(0.8)   & 23.0(1.2)  & 25.5(1.5)       & \due{33.6(1.1)}* \\
\midrule
\multirow{5}{*}{CPSC2018}
          & 12 leads   & 90.1     & \uno{97.7}       & 96.9*           & 90.9            & 94.3        & \due{97.6}* & 92.2       & \tre{97.4}*     & 90.6             \\
          & Lead I     & 90.1     & 30.2             & \uno{69.8}      & 8.3             & \tre{30.6}  & 15.2        & 13.1       & 13.6            & \due{53.2}       \\
          & Lead II    & 90.1     & 31.1             & \uno{75.6}      & \due{63.6}      & 22.8        & 10.5        & 13.7       & 21.4            & \tre{56.6}       \\
          & One random & 90.4     & \due{47.4(1.0)}  & \tre{46.9(1.6)} & 43.6(2.4)       & 33.1(2.2)   & 14.1(2.8)   & 14.3(1.0)  & 30.3(1.5)       & \uno{62.1(1.1)}  \\
          & Two random & 90.3     & \due{68.1(1.6)}  & \tre{64.8(3.0)} & 57.4(1.3)       & 49.4(1.7)   & 18.4(0.9)   & 28.6(1.0)  & 50.7(2.3)       & \uno{73.7(1.2)}  \\
\bottomrule
\end{tabular}
}
\end{table*}

\begin{table*}[p]
\centering
{\small\textbf{Supplementary Table~\ref{tab:supp_sens90spec} (continued).}}\par\medskip
\setlength{\tabcolsep}{3pt}
\renewcommand{\arraystretch}{0.9}
\resizebox{\textwidth}{!}{
\begin{tabular}{llc cccccccc}
\toprule
Benchmark & Input      & Achieved & ECGFounder       & ECG-JEPA         & ST-MEM          & MERL        & ECGFM-KED  & HuBERT-ECG       & ECG-CPC         & \textbf{RELAY}   \\
\midrule\multirow{5}{*}{CPSC-Extra}
          & 12 leads   & 90.2     & \uno{88.8}       & 79.5             & 81.3            & \tre{82.6}  & 78.9       & 70.9             & \due{86.1}*     & 66.4             \\
          & Lead I     & 90.2     & \due{36.8}       & \uno{55.4}       & 12.9            & 26.7        & 13.6       & 14.2             & 25.0            & \tre{36.1}       \\
          & Lead II    & 90.2     & 39.3             & \uno{54.6}       & \due{47.1}*     & 21.1        & 14.8       & 18.7             & 23.5            & \tre{41.9}       \\
          & One random & 90.2     & \uno{52.8(3.4)}  & \tre{40.9(6.1)}  & 26.3(1.3)       & 27.2(1.1)   & 13.3(2.4)  & 15.6(2.2)        & 34.9(3.4)       & \due{42.7(3.2)}  \\
          & Two random & 90.2     & \uno{68.0(2.2)}  & 50.1(2.2)        & 33.8(1.6)       & 39.9(2.8)   & 18.3(3.1)  & 28.4(2.8)        & \tre{51.0(2.7)} & \due{51.9(1.9)}  \\
\midrule
\multirow{5}{*}{Chapman}
          & 12 leads   & 90.0     & \uno{93.6}       & \due{93.3}*      & 83.8            & 81.5        & 73.5       & 82.4             & \tre{88.4}      & 83.5             \\
          & Lead I     & 90.1     & \tre{26.0}       & \due{49.1}*      & 14.9            & 22.4        & 12.9       & 21.1             & 18.4            & \uno{51.8}       \\
          & Lead II    & 90.1     & 36.1             & \uno{62.2}       & \tre{47.9}      & 29.5        & 21.7       & 17.5             & 21.6            & \due{60.6}*      \\
          & One random & 90.2     & \due{42.1(3.2)}  & \tre{37.9(3.2)}  & 26.9(2.2)       & 27.0(1.3)   & 14.9(1.4)  & 17.3(1.9)        & 23.1(1.6)       & \uno{49.9(2.9)}  \\
          & Two random & 90.1     & \uno{61.0(2.3)}  & \tre{54.8(4.6)}  & 31.4(4.6)       & 37.9(1.6)   & 20.8(3.4)  & 25.9(1.0)        & 36.7(2.2)       & \due{59.8(3.1)}* \\
\midrule
\multirow{5}{*}{Chapman Rhythm$^\dagger$}
          & 12 leads   & 90.1     & \uno{99.0}       & \due{98.0}*      & 97.3*           & 92.5        & 85.9       & 94.7             & 97.0            & \tre{97.6}*      \\
          & Lead I     & 90.2     & \tre{49.2}       & \due{78.9}*      & 12.3            & 28.4        & 20.0       & 16.7             & 18.1            & \uno{80.5}       \\
          & Lead II    & 90.1     & 56.8             & \uno{91.6}       & \tre{72.5}      & 33.0        & 26.6       & 18.3             & 20.7            & \due{81.8}       \\
          & One random & 90.4     & \due{65.3(1.7)}* & \tre{49.8(4.7)}  & 46.2(3.1)       & 34.9(2.2)   & 20.5(4.2)  & 18.1(4.4)        & 28.8(7.2)       & \uno{76.5(3.5)}  \\
          & Two random & 90.2     & \uno{86.9(1.4)}  & \tre{68.4(5.3)}  & 51.5(3.7)       & 50.4(2.0)   & 34.0(7.0)  & 31.4(1.6)        & 50.0(2.8)       & \due{86.7(2.8)}* \\
\midrule
\multirow{5}{*}{Georgia}
          & 12 leads   & 90.1     & 87.5             & \uno{92.3}       & 79.1            & \tre{87.8}  & 75.6       & 63.3             & \due{91.0}*     & 69.0             \\
          & Lead I     & 90.1     & \tre{33.8}       & \uno{48.1}       & 9.9             & 22.2        & 17.6       & 8.4              & 21.6            & \due{40.7}*      \\
          & Lead II    & 90.1     & 27.8             & \uno{47.2}       & \tre{43.3}*     & 17.2        & 18.2       & 12.8             & 14.2            & \due{44.4}*      \\
          & One random & 90.1     & \uno{44.0(2.0)}  & \tre{30.5(2.8)}  & 24.8(2.8)       & 28.0(2.6)   & 16.3(0.8)  & 9.6(1.2)         & 26.4(2.0)       & \due{40.5(1.8)}* \\
          & Two random & 90.1     & \uno{58.7(1.4)}  & 39.8(3.8)        & 29.7(1.4)       & 38.0(2.7)   & 24.5(1.3)  & 16.0(1.6)        & \tre{40.2(2.2)} & \due{46.4(1.5)}  \\
\midrule
\multirow{5}{*}{Hefei}
          & 12 leads   & 90.0     & 86.3             & 85.7             & \due{88.2}      & \uno{93.5}* & \tre{87.2} & 86.5             & \uno{93.5}      & 86.8             \\
          & Lead I     & 90.2     & \tre{41.2}       & \due{47.7}*      & 14.3            & 25.5        & 11.6       & 19.0             & 16.3            & \uno{51.8}       \\
          & Lead II    & 90.1     & 54.0             & \tre{54.6}       & \due{56.7}      & 30.1        & 17.5       & 29.2             & 27.9            & \uno{65.3}       \\
          & One random & 90.2     & \due{36.9(4.1)}  & \tre{36.0(4.6)}  & 32.0(5.1)       & 25.7(2.5)   & 11.9(3.5)  & 18.0(3.6)        & 23.2(2.6)       & \uno{52.9(3.0)}  \\
          & Two random & 90.2     & \tre{40.8(3.9)}  & \due{46.3(4.6)}  & 36.2(3.0)       & 35.2(1.1)   & 19.6(3.5)  & 28.6(4.8)        & 34.4(2.9)       & \uno{67.0(2.4)}  \\
\midrule
\multirow{5}{*}{Ningbo}
          & 12 leads   & 90.0     & \due{97.1}       & \tre{96.5}       & 91.7            & \tre{96.5}  & 92.9       & 86.0             & \uno{98.1}      & 89.4             \\
          & Lead I     & 90.0     & \tre{28.4}       & \uno{64.1}       & 11.7            & 24.2        & 23.8       & 28.1             & 17.8            & \due{56.7}       \\
          & Lead II    & 90.0     & 40.0             & \uno{72.8}       & \tre{48.7}      & 30.6        & 21.0       & 28.3             & 26.6            & \due{62.1}       \\
          & One random & 90.0     & \due{50.3(2.2)}* & \tre{43.1(2.1)}  & 28.1(1.2)       & 30.6(1.4)   & 21.5(2.6)  & 27.9(1.7)        & 25.5(1.3)       & \uno{52.7(2.2)}  \\
          & Two random & 90.0     & \uno{72.9(0.6)}  & \tre{58.4(3.0)}  & 38.2(1.6)       & 48.0(2.6)   & 35.4(1.9)  & 43.9(2.1)        & 43.2(1.1)       & \due{69.0(2.2)}  \\
\midrule
\multirow{5}{*}{PTB}
          & 12 leads   & 90.7     & \due{34.8}*      & \tre{33.9}*      & 31.6*           & \uno{40.0}  & 15.7       & 27.3*            & 32.2*           & 23.8             \\
          & Lead I     & 90.7     & \uno{21.4}       & \due{19.5}*      & 12.3*           & 17.3*       & 12.5*      & \tre{18.5}*      & 17.3*           & 11.2*            \\
          & Lead II    & 90.7     & 11.6             & \due{25.8}*      & \uno{28.0}      & \tre{20.1}* & 10.7       & 12.5*            & 15.2            & 15.9*            \\
          & One random & 90.7     & \uno{27.4(3.0)}  & \due{23.9(2.8)}* & \tre{19.0(2.8)} & 15.5(1.4)   & 11.2(3.5)  & 15.4(3.3)        & 14.8(2.9)*      & 18.4(2.4)        \\
          & Two random & 90.7     & \due{30.7(1.2)}* & 22.3(0.6)*       & \uno{31.2(6.3)} & 19.8(1.7)   & 11.2(2.6)  & \tre{23.1(3.0)}* & 14.5(2.4)       & 20.3(0.6)*       \\
\midrule
\multirow{5}{*}{SPH}
          & 12 leads   & 90.0     & \due{96.0}*      & \uno{96.4}*      & 87.3            & 89.2        & 84.6       & 88.6             & \uno{96.4}      & \tre{89.9}       \\
          & Lead I     & 90.1     & \tre{33.0}       & \due{60.1}*      & 12.5            & 22.3        & 19.3       & 25.8             & 28.6            & \uno{63.6}       \\
          & Lead II    & 90.1     & 37.6             & \uno{75.3}       & \tre{59.9}      & 32.8        & 15.8       & 21.1             & 28.7            & \due{69.7}*      \\
          & One random & 90.1     & \tre{47.1(2.5)}  & \due{49.1(3.1)}  & 29.1(1.7)       & 31.5(5.3)   & 16.0(1.7)  & 17.7(1.8)        & 30.9(2.7)       & \uno{54.2(3.3)}  \\
          & Two random & 90.1     & \tre{62.3(1.7)}* & \due{64.9(1.8)}* & 36.2(1.6)       & 38.9(1.6)   & 21.1(1.1)  & 31.8(4.8)        & 41.3(1.1)       & \uno{66.5(2.2)}  \\
\midrule
\multirow{5}{*}{ZZU pECG}
          & 12 leads   & 90.0     & 79.0*            & \uno{81.5}       & \tre{79.2}*     & 75.0        & 73.2       & 69.3             & \due{79.7}*     & 70.0             \\
          & Lead I     & 90.0     & \tre{41.1}       & \uno{57.5}       & 12.4            & 27.7        & 14.0       & 18.1             & 14.9            & \due{48.0}       \\
          & Lead II    & 90.0     & 46.4             & \uno{63.3}       & \tre{52.2}      & 31.1        & 41.4       & 15.3             & 27.8            & \due{57.5}*      \\
          & One random & 90.1     & \uno{53.7(1.0)}  & \due{53.2(3.4)}* & 31.6(2.4)       & 34.0(2.2)   & 22.0(2.5)  & 20.3(1.1)        & 25.7(2.4)       & \tre{51.6(4.0)}* \\
          & Two random & 90.0     & \uno{67.1(1.3)}  & \due{63.6(1.9)}* & 35.4(1.2)       & 44.9(2.5)   & 31.6(2.7)  & 27.1(1.1)        & 36.9(1.1)       & \tre{60.9(2.7)}  \\

\bottomrule
\end{tabular}
}
\end{table*}

\begin{table*}[p]
\centering
\caption{\textbf{Specificity at 80\% sensitivity.} For every condition, specificity is deliberately read from the ROC curve of the test set at the threshold giving a sensitivity above 80\%, then the macro average over conditions $\times 100$ is reported (conditions with no positive test sample are left out). Like AUROC, these values summarise the ROC curve: they are the highest reachable by any threshold at this operating point, not the result of a deployable threshold (Supplementary Table~\ref{tab:supp_threshold_transfer}). Achieved: sensitivity actually reached by the thresholds, averaged over conditions and models. It exceeds the target because sensitivity moves in large steps when a condition has few positives, which makes the reported specificities conservative. When leads are sampled randomly, mean(std) over 5 seeds is reported. ST-MEM runs on the available leads only, with their true lead embeddings. *: not significantly worse than the best model of the row (paired bootstrap of the difference, 1000 resamples with thresholds re-derived, 95\% CI includes 0). Bold, underlined and italic values are the best, second-best and third-best of each row. The MIMIC-IV-ECG benchmarks are not included. $^\dagger$Evaluation-only benchmark.}
\label{tab:supp_spec80sens}
\setlength{\tabcolsep}{3pt}
\renewcommand{\arraystretch}{0.9}
\resizebox{\textwidth}{!}{
\begin{tabular}{llc cccccccc}
\toprule
Benchmark & Input      & Achieved & ECGFounder       & ECG-JEPA        & ST-MEM          & MERL            & ECGFM-KED  & HuBERT-ECG      & ECG-CPC         & \textbf{RELAY}  \\
\midrule
\multirow{5}{*}{CODE 15\%}
          & 12 leads   & 82.8     & \tre{98.9}*      & \uno{99.1}      & \tre{98.9}*     & 97.8            & 97.9       & \uno{99.1}*     & \due{99.0}*     & 98.6*           \\
          & Lead I     & 82.9     & \uno{93.9}       & \tre{90.8}*     & 24.5            & 48.7            & 55.4       & 30.3            & 33.9            & \due{92.8}*     \\
          & Lead II    & 82.8     & \tre{89.2}*      & \due{91.0}*     & 80.4            & 47.2            & 51.6       & 27.3            & 31.0            & \uno{95.4}      \\
          & One random & 84.8     & \due{90.7(2.1)}* & 27.6(5.8)       & 61.3(3.2)       & \tre{64.3(3.0)} & 40.5(1.1)  & 21.0(0.7)       & 44.0(4.2)       & \uno{94.3(0.6)} \\
          & Two random & 82.9     & \due{97.2(0.9)}* & 79.4(5.8)       & 75.8(4.6)       & \tre{80.3(3.0)} & 58.0(2.7)  & 30.7(3.4)       & 51.2(2.5)       & \uno{97.4(0.3)} \\
\midrule
\multirow{5}{*}{PTB-XL All}
          & 12 leads   & 86.5     & \tre{89.5}*      & \due{89.6}*     & 84.1            & 86.7            & 83.4       & 85.4            & \uno{91.0}      & 84.6            \\
          & Lead I     & 86.5     & \tre{43.6}       & \due{50.3}      & 24.5            & 25.5            & 26.8       & 27.6            & 24.9            & \uno{59.5}      \\
          & Lead II    & 86.5     & 39.5             & \tre{53.9}      & \due{56.2}      & 26.1            & 23.1       & 27.7            & 33.6            & \uno{62.3}      \\
          & One random & 87.0     & \due{47.5(0.6)}  & 30.9(1.0)       & \tre{37.7(1.0)} & 30.4(1.3)       & 27.6(1.2)  & 26.2(1.5)       & 33.9(1.5)       & \uno{56.3(2.3)} \\
          & Two random & 86.6     & \due{65.5(0.7)}* & \tre{48.3(3.3)} & 43.4(0.6)       & 41.8(1.2)       & 34.2(3.0)  & 32.4(1.2)       & 46.6(1.5)       & \uno{67.4(0.9)} \\
\midrule
\multirow{5}{*}{PTB-XL Diag$^\dagger$}
          & 12 leads   & 86.7     & \due{91.4}*      & \tre{90.1}*     & 83.0            & 89.7*           & 85.2       & 87.2            & \uno{91.6}      & 87.9            \\
          & Lead I     & 86.7     & \tre{42.7}       & \due{47.2}      & 25.1            & 26.3            & 27.6       & 24.4            & 22.3            & \uno{58.7}      \\
          & Lead II    & 86.7     & 36.9             & \tre{51.0}      & \due{53.8}      & 28.6            & 26.0       & 25.7            & 33.6            & \uno{62.1}      \\
          & One random & 87.2     & \due{45.0(1.4)}  & 30.4(1.3)       & \tre{35.9(2.0)} & 32.0(1.3)       & 28.7(1.5)  & 24.8(2.9)       & 32.4(1.4)       & \uno{53.2(3.8)} \\
          & Two random & 86.8     & \due{65.5(0.9)}* & \tre{46.2(4.6)} & 41.9(0.6)       & 43.0(1.4)       & 34.4(3.9)  & 31.3(2.1)       & 45.1(2.1)       & \uno{67.9(1.0)} \\
\midrule
\multirow{5}{*}{PTB-XL Form$^\dagger$}
          & 12 leads   & 84.5     & \tre{82.0}       & \due{84.0}      & 80.4            & 81.8            & 81.1       & 75.1            & \uno{87.2}      & 79.2            \\
          & Lead I     & 84.6     & \tre{33.9}       & \due{49.8}      & 22.6            & 21.1            & 22.4       & 27.6            & 28.7            & \uno{56.2}      \\
          & Lead II    & 84.6     & 29.6             & \tre{46.4}      & \due{52.3}*     & 25.6            & 19.4       & 24.2            & 29.1            & \uno{54.8}      \\
          & One random & 85.4     & \due{38.0(1.2)}  & 27.4(3.0)       & \tre{32.3(2.4)} & 26.0(2.4)       & 21.8(3.9)  & 23.5(2.2)       & 30.1(3.2)       & \uno{54.2(2.7)} \\
          & Two random & 84.7     & \due{54.8(3.0)}  & \tre{45.6(2.7)} & 37.8(1.6)       & 35.2(2.3)       & 27.7(1.7)  & 27.5(1.7)       & 38.8(4.0)       & \uno{61.5(3.0)} \\
\midrule
\multirow{5}{*}{PTB-XL Rhythm$^\dagger$}
          & 12 leads   & 87.0     & \tre{93.6}*      & \uno{96.2}      & \tre{93.6}      & 83.6            & 80.1       & \tre{93.6}*     & \due{94.0}*     & 81.1            \\
          & Lead I     & 87.0     & \tre{63.6}*      & \due{66.3}*     & 26.2            & 25.6            & 31.4       & 39.7            & 34.5            & \uno{67.8}      \\
          & Lead II    & 87.0     & 62.4             & \uno{77.9}      & \due{74.3}*     & 20.3            & 20.7       & 40.3            & 40.6            & \tre{74.2}*     \\
          & One random & 87.5     & \due{71.0(1.5)}* & 39.1(1.9)       & \tre{53.2(2.8)} & 29.1(4.2)       & 32.0(6.1)  & 35.1(1.9)       & 44.2(3.1)       & \uno{71.6(3.2)} \\
          & Two random & 87.0     & \uno{82.3(1.9)}  & 62.7(1.3)       & 57.2(1.9)       & 46.1(5.4)       & 42.0(6.8)  & 43.3(2.2)       & \tre{63.8(2.2)} & \due{75.5(1.8)} \\
\midrule
\multirow{5}{*}{PTB-XL Sub}
          & 12 leads   & 83.7     & \uno{91.1}       & 89.4*           & 86.3            & \tre{90.0}*     & 80.7       & 86.3            & \due{90.5}*     & 88.8*           \\
          & Lead I     & 83.7     & \due{50.5}       & \tre{49.3}      & 19.7            & 27.6            & 18.6       & 28.0            & 20.0            & \uno{65.8}      \\
          & Lead II    & 83.7     & 44.4             & \due{56.1}      & \tre{51.4}      & 25.0            & 24.9       & 30.0            & 23.2            & \uno{63.8}      \\
          & One random & 84.7     & \due{50.3(2.4)}  & 23.9(4.0)       & \tre{36.9(3.8)} & 29.2(1.7)       & 21.1(2.0)  & 23.8(2.0)       & 26.0(1.1)       & \uno{56.5(0.3)} \\
          & Two random & 83.8     & \uno{70.9(1.4)}  & \tre{43.8(4.3)} & 41.8(2.1)       & 41.9(2.7)       & 25.5(2.5)  & 34.2(2.1)       & 37.0(3.2)       & \due{67.1(1.6)} \\
\midrule
\multirow{5}{*}{PTB-XL Super}
          & 12 leads   & 80.1     & \uno{90.0}       & 88.0            & 82.1            & \tre{89.7}*     & 84.9       & 84.1            & \due{89.8}*     & 86.3            \\
          & Lead I     & 80.1     & \tre{27.7}       & \due{46.9}      & 21.7            & 16.1            & 18.3       & 24.5            & 11.5            & \uno{59.5}      \\
          & Lead II    & 80.1     & 32.3             & \tre{44.3}      & \due{44.7}      & 21.2            & 20.9       & 26.9            & 21.5            & \uno{59.9}      \\
          & One random & 81.2     & \due{34.3(0.7)}  & 20.3(0.8)       & \tre{31.1(0.6)} & 27.0(0.7)       & 21.7(0.2)  & 21.9(0.3)       & 18.8(0.6)       & \uno{49.3(1.6)} \\
          & Two random & 80.1     & \due{54.1(0.9)}  & 38.8(1.5)       & \tre{38.9(1.2)} & 37.3(0.7)       & 26.0(0.7)  & 27.6(0.7)       & 24.3(1.4)       & \uno{61.4(1.0)} \\
\midrule
\multirow{5}{*}{EchoNext}
          & 12 leads   & 80.7     & \due{68.6}       & \tre{68.1}      & 66.8            & 67.8            & 66.4       & 63.8            & \uno{72.2}      & 65.3            \\
          & Lead I     & 80.7     & \uno{54.0}       & \due{50.8}*     & 23.4            & 26.3            & 28.6       & 40.0            & 29.7            & \tre{41.8}      \\
          & Lead II    & 80.7     & \uno{52.1}       & \due{46.0}      & 35.5            & 29.7            & 23.7       & 34.3            & 20.6            & \tre{45.9}      \\
          & One random & 81.6     & \uno{46.4(1.5)}  & 23.2(1.1)       & 28.6(0.6)       & 26.9(1.7)       & 21.9(1.0)  & \tre{30.6(0.5)} & 26.6(1.6)       & \due{40.8(1.1)} \\
          & Two random & 80.8     & \uno{54.8(0.8)}  & 38.1(2.6)       & 32.9(1.3)       & 36.9(0.7)       & 24.5(0.2)  & \tre{40.6(1.1)} & 39.5(1.7)       & \due{48.6(1.1)} \\
\midrule
\multirow{5}{*}{CPSC2018}
          & 12 leads   & 81.1     & \due{99.0}       & \tre{98.7}      & 95.5            & 97.6            & \uno{99.8} & 96.0            & \tre{98.7}      & 94.9            \\
          & Lead I     & 81.0     & 45.3             & \uno{81.8}      & 17.5            & \tre{45.9}      & 28.2       & 31.3            & 26.4            & \due{69.1}      \\
          & Lead II    & 81.0     & 49.7             & \uno{80.0}      & \due{73.7}      & 35.1            & 22.8       & 30.9            & 33.5            & \tre{72.7}      \\
          & One random & 81.8     & \due{61.0(1.0)}  & 47.8(2.4)       & \tre{57.8(2.8)} & 45.9(2.2)       & 21.7(1.4)  & 27.7(0.9)       & 43.1(2.4)       & \uno{72.3(1.4)} \\
          & Two random & 81.0     & \due{80.2(1.3)}* & \tre{78.7(2.0)} & 69.7(1.3)       & 64.6(1.3)       & 28.6(2.0)  & 43.0(0.8)       & 65.3(1.2)       & \uno{84.0(1.6)} \\
\bottomrule
\end{tabular}
}
\end{table*}

\begin{table*}[p]
\centering
{\small\textbf{Supplementary Table~\ref{tab:supp_spec80sens} (continued).}}\par\medskip
\setlength{\tabcolsep}{3pt}
\renewcommand{\arraystretch}{0.9}
\resizebox{\textwidth}{!}{
\begin{tabular}{llc cccccccc}
\toprule
Benchmark & Input      & Achieved & ECGFounder       & ECG-JEPA         & ST-MEM          & MERL            & ECGFM-KED   & HuBERT-ECG       & ECG-CPC          & \textbf{RELAY}   \\
\midrule\multirow{5}{*}{CPSC-Extra}
          & 12 leads   & 90.8     & \uno{92.6}       & 86.6             & 88.2            & 89.2            & \tre{89.7}* & 78.4             & \due{89.9}       & 76.6             \\
          & Lead I     & 90.8     & \tre{43.2}       & \uno{65.9}       & 25.9            & 35.3            & 28.1        & 32.1             & 41.7             & \due{56.5}       \\
          & Lead II    & 90.8     & 40.6             & \due{63.0}*      & \uno{68.3}      & 36.9            & 31.4        & 36.3             & 35.6             & \tre{60.0}       \\
          & One random & 91.3     & \uno{63.3(2.7)}  & 44.9(4.5)        & 46.2(2.5)       & 41.0(2.5)       & 29.5(4.2)   & 35.5(2.5)        & \tre{47.6(2.9)}  & \due{54.4(1.1)}  \\
          & Two random & 90.9     & \uno{78.3(1.3)}  & \tre{63.5(3.2)}  & 52.4(2.1)       & 54.1(4.2)       & 35.8(1.3)   & 43.6(3.2)        & 61.9(3.2)        & \due{64.1(2.8)}  \\
\midrule
\multirow{5}{*}{Chapman}
          & 12 leads   & 87.2     & \uno{95.3}       & \due{94.8}*      & 92.0            & 92.0            & 87.8        & 89.6             & \tre{94.3}*      & 91.5             \\
          & Lead I     & 87.2     & \tre{44.3}       & \due{60.8}       & 28.9            & 35.1            & 28.1        & 41.3             & 32.2             & \uno{66.2}       \\
          & Lead II    & 87.2     & 58.9             & \due{70.5}*      & \tre{61.8}      & 46.5            & 43.1        & 37.4             & 36.5             & \uno{75.6}       \\
          & One random & 87.7     & \due{59.5(1.7)}  & 40.5(2.0)        & \tre{45.0(1.6)} & 44.0(2.0)       & 30.7(2.2)   & 35.5(2.3)        & 39.6(1.9)        & \uno{68.0(2.8)}  \\
          & Two random & 87.3     & \due{76.5(1.0)}* & \tre{63.9(2.7)}  & 48.9(3.8)       & 55.4(1.7)       & 39.1(2.9)   & 45.0(2.6)        & 54.4(1.2)        & \uno{77.1(2.1)}  \\
\midrule
\multirow{5}{*}{Chapman Rhythm$^\dagger$}
          & 12 leads   & 83.4     & \due{98.9}*      & \uno{99.3}       & \tre{97.9}*     & 95.7            & 94.4        & 96.3             & 97.5*            & \tre{97.9}*      \\
          & Lead I     & 83.4     & \tre{65.7}       & \uno{88.2}       & 22.5            & 45.6            & 35.2        & 37.0             & 29.6             & \due{88.1}*      \\
          & Lead II    & 83.4     & 73.2             & \uno{94.8}       & \tre{81.4}      & 51.2            & 53.0        & 37.7             & 24.9             & \due{90.1}*      \\
          & One random & 84.5     & \due{80.6(3.1)}  & 51.5(2.9)        & \tre{61.7(2.3)} & 52.0(1.9)       & 34.3(4.4)   & 33.5(3.7)        & 42.1(3.6)        & \uno{86.4(2.5)}  \\
          & Two random & 83.5     & \due{92.2(0.7)}* & \tre{83.3(1.2)}  & 67.1(3.3)       & 61.6(2.0)       & 44.8(6.8)   & 44.6(3.4)        & 62.4(2.3)        & \uno{92.5(1.8)}  \\
\midrule
\multirow{5}{*}{Georgia}
          & 12 leads   & 86.9     & \tre{93.5}       & \uno{96.3}       & 86.5            & 91.0            & 84.9        & 75.8             & \due{93.9}       & 78.0             \\
          & Lead I     & 86.9     & \tre{48.8}       & \due{54.5}       & 21.9            & 34.5            & 32.6        & 24.2             & 36.9             & \uno{59.8}       \\
          & Lead II    & 86.9     & 44.9             & \tre{56.3}*      & \uno{60.7}      & 28.9            & 31.6        & 27.7             & 28.9             & \due{58.9}*      \\
          & One random & 87.6     & \due{55.8(1.8)}* & 31.3(2.0)        & \tre{42.0(3.1)} & 39.6(2.4)       & 30.4(1.5)   & 25.4(2.3)        & 38.5(2.0)        & \uno{56.2(2.4)}  \\
          & Two random & 87.0     & \uno{71.5(1.3)}  & 48.5(3.5)        & 45.0(1.8)       & 51.4(1.1)       & 41.1(1.2)   & 30.3(1.5)        & \tre{53.2(2.8)}  & \due{64.9(1.4)}  \\
\midrule
\multirow{5}{*}{Hefei}
          & 12 leads   & 87.7     & \tre{92.7}       & 89.7             & 91.7            & \due{94.3}*     & 91.3        & 89.9             & \uno{95.9}       & 92.5             \\
          & Lead I     & 87.8     & \tre{56.3}*      & \uno{60.9}       & 30.3            & 38.2            & 32.3        & 36.1             & 38.6             & \due{60.7}*      \\
          & Lead II    & 87.8     & 64.5             & \tre{66.4}       & \due{71.8}      & 43.9            & 36.9        & 44.4             & 41.2             & \uno{75.9}       \\
          & One random & 88.3     & \due{52.6(1.8)}  & 38.3(1.8)        & \tre{51.0(1.7)} & 40.2(3.5)       & 27.9(3.8)   & 36.6(2.7)        & 40.1(2.3)        & \uno{68.2(1.1)}  \\
          & Two random & 87.8     & 53.5(3.9)        & \due{55.3(2.1)}  & 53.3(2.6)       & 51.7(2.0)       & 36.6(2.5)   & 47.1(2.9)        & \tre{53.8(1.9)}  & \uno{77.4(1.1)}  \\
\midrule
\multirow{5}{*}{Ningbo}
          & 12 leads   & 86.5     & \due{98.0}       & 97.2             & 94.8            & \tre{97.6}      & 95.2        & 92.2             & \uno{98.5}       & 92.4             \\
          & Lead I     & 86.5     & 39.5             & \uno{73.4}       & 27.3            & 39.7            & 37.8        & \tre{42.7}       & 31.4             & \due{67.9}       \\
          & Lead II    & 86.5     & 55.5             & \uno{80.1}       & \tre{63.6}      & 48.0            & 33.3        & 44.2             & 40.9             & \due{74.4}       \\
          & One random & 87.8     & \due{62.2(1.6)}  & 25.9(3.8)        & 45.7(0.9)       & \tre{46.7(1.3)} & 34.7(1.2)   & 43.5(1.1)        & 37.3(0.7)        & \uno{68.3(1.5)}  \\
          & Two random & 86.7     & \uno{80.8(1.5)}  & \tre{63.0(1.7)}  & 53.9(1.8)       & 61.6(1.4)       & 49.0(1.5)   & 56.6(1.0)        & 57.8(1.7)        & \due{80.3(1.9)}* \\
\midrule
\multirow{5}{*}{PTB}
          & 12 leads   & 90.0     & \tre{54.3}*      & 53.9*            & 53.2*           & 54.2*           & 40.7        & \due{54.9}*      & \uno{58.1}       & 49.9*            \\
          & Lead I     & 90.0     & \due{38.7}*      & \tre{37.0}*      & 25.6            & 33.9*           & 28.4*       & \uno{41.8}       & 33.6*            & 34.3*            \\
          & Lead II    & 90.0     & 33.4             & \tre{43.6}*      & \uno{45.7}      & 31.3            & 23.7        & 40.6*            & 23.8             & \due{45.2}*      \\
          & One random & 90.0     & \uno{44.9(2.8)}  & \tre{39.1(4.8)}* & 37.4(2.0)       & 34.2(1.2)       & 29.4(3.3)   & 35.3(4.1)*       & \due{39.2(1.9)}* & 37.5(1.6)*       \\
          & Two random & 90.0     & \uno{52.6(1.2)}  & \tre{44.1(0.7)}* & 41.6(5.3)       & 36.9(1.3)       & 28.7(3.5)   & \due{46.3(3.1)}* & 42.6(1.6)        & 42.1(2.5)        \\
\midrule
\multirow{5}{*}{SPH}
          & 12 leads   & 88.2     & \due{97.6}*      & \uno{97.7}       & 95.0            & 92.9            & 90.5        & 93.8             & \tre{97.0}*      & 94.4             \\
          & Lead I     & 88.3     & \tre{49.2}       & \due{68.3}*      & 24.0            & 34.1            & 30.5        & 39.4             & 36.0             & \uno{72.2}       \\
          & Lead II    & 88.2     & 45.1             & \uno{82.3}       & \tre{71.8}      & 45.6            & 28.0        & 42.0             & 42.2             & \due{77.1}       \\
          & One random & 89.2     & \due{58.1(1.7)}  & 36.7(4.3)        & \tre{49.1(2.1)} & 44.4(3.0)       & 30.6(4.2)   & 33.7(1.9)        & 47.0(1.2)        & \uno{69.3(2.1)}  \\
          & Two random & 88.3     & \due{73.3(2.0)}  & \tre{68.1(3.9)}  & 54.0(1.4)       & 54.1(2.4)       & 35.5(3.4)   & 45.4(2.3)        & 57.3(2.8)        & \uno{79.7(1.8)}  \\
\midrule
\multirow{5}{*}{ZZU pECG}
          & 12 leads   & 89.8     & \due{87.9}*      & \uno{89.0}       & \tre{86.7}      & 86.0            & 80.6        & 83.0             & 86.6*            & 83.3             \\
          & Lead I     & 89.8     & \tre{60.1}       & \uno{70.6}       & 31.4            & 43.1            & 29.2        & 36.2             & 38.2             & \due{63.5}       \\
          & Lead II    & 89.8     & \tre{68.6}       & \uno{72.6}       & 68.2            & 52.9            & 60.8        & 35.0             & 46.2             & \due{72.0}*      \\
          & One random & 89.8     & \uno{71.8(1.1)}  & \tre{64.3(2.1)}  & 51.0(1.6)       & 50.2(2.1)       & 40.2(2.6)   & 37.9(1.4)        & 42.9(2.4)        & \due{67.4(1.9)}  \\
          & Two random & 89.8     & \uno{79.4(0.4)}  & \due{75.3(1.5)}  & 54.9(1.1)       & 65.1(1.3)       & 54.3(1.4)   & 46.9(0.9)        & 57.1(2.3)        & \tre{74.8(2.5)}  \\

\bottomrule
\end{tabular}
}
\end{table*}

\begin{table*}[p]
\centering
\caption{\textbf{Sensitivity at 95\% specificity.} For every condition, sensitivity is deliberately read from the ROC curve of the test set at the threshold giving a specificity above 95\%, then the macro average over conditions $\times 100$ is reported (conditions with no positive test sample are left out). Like AUROC, these values summarise the ROC curve: they are the highest reachable by any threshold at this operating point, not the result of a deployable threshold (Supplementary Table~\ref{tab:supp_threshold_transfer}). Achieved: specificity actually reached by the thresholds, averaged over conditions and models. When leads are sampled randomly, mean(std) over 5 seeds is reported. ST-MEM runs on the available leads only, with their true lead embeddings. *: not significantly worse than the best model of the row (paired bootstrap of the difference, 1000 resamples with thresholds re-derived, 95\% CI includes 0). Bold, underlined and italic values are the best, second-best and third-best of each row. The MIMIC-IV-ECG benchmarks are not included. $^\dagger$Evaluation-only benchmark.}
\label{tab:supp_sens95spec}
\setlength{\tabcolsep}{3pt}
\renewcommand{\arraystretch}{0.9}
\resizebox{\textwidth}{!}{
\begin{tabular}{llc cccccccc}
\toprule
Benchmark & Input      & Achieved & ECGFounder       & ECG-JEPA        & ST-MEM      & MERL        & ECGFM-KED  & HuBERT-ECG & ECG-CPC     & \textbf{RELAY}   \\
\midrule
\multirow{5}{*}{CODE 15\%}
          & 12 leads   & 95.0     & 95.5*            & 97.1*           & \uno{98.6}  & 93.9        & 92.0*      & 97.1*      & \tre{98.0}* & \due{98.5}*      \\
          & Lead I     & 95.0     & \due{79.4}*      & \tre{71.6}*     & 8.7         & 28.8        & 13.7       & 15.2       & 8.6         & \uno{80.3}       \\
          & Lead II    & 95.0     & \tre{73.3}       & \due{73.9}*     & 71.6        & 22.5        & 14.1       & 11.7       & 7.3         & \uno{84.5}       \\
          & One random & 95.0     & \due{78.6(2.2)}* & \tre{46.3(6.8)} & 32.0(4.0)   & 39.7(1.9)   & 16.1(1.8)  & 8.4(1.8)   & 9.5(1.6)    & \uno{79.0(0.6)}  \\
          & Two random & 95.0     & \due{90.3(1.4)}* & \tre{63.3(4.3)} & 43.9(7.4)   & 58.4(2.4)   & 30.0(3.6)  & 18.4(3.2)  & 19.9(2.5)   & \uno{93.1(0.8)}  \\
\midrule
\multirow{5}{*}{PTB-XL All}
          & 12 leads   & 95.0     & \due{72.6}*      & \tre{71.9}*     & 64.4        & 68.9        & 55.2       & 63.5       & \uno{73.9}  & 62.3             \\
          & Lead I     & 95.0     & \tre{19.7}       & \uno{31.9}      & 5.2         & 9.5         & 5.9        & 9.0        & 7.1         & \due{29.3}*      \\
          & Lead II    & 95.0     & 18.2             & \due{30.8}*     & \tre{30.5}* & 6.5         & 9.7        & 8.9        & 12.3        & \uno{32.4}       \\
          & One random & 95.1     & \due{21.7(1.0)}  & \tre{17.9(1.7)} & 13.7(1.9)   & 12.0(1.5)   & 7.8(0.4)   & 7.3(0.5)   & 13.2(1.2)   & \uno{26.2(1.4)}  \\
          & Two random & 95.0     & \uno{38.6(1.4)}  & \tre{26.9(2.4)} & 17.7(1.3)   & 20.1(0.9)   & 13.3(1.6)  & 12.5(1.3)  & 21.3(1.5)   & \due{36.3(0.7)}* \\
\midrule
\multirow{5}{*}{PTB-XL Diag$^\dagger$}
          & 12 leads   & 95.0     & \uno{76.2}       & 71.3            & 64.2        & \tre{73.3}* & 58.3       & 65.4       & \due{74.6}* & 68.1             \\
          & Lead I     & 95.0     & \tre{16.3}       & \uno{31.8}      & 5.2         & 11.4        & 6.7        & 7.6        & 6.5         & \due{28.9}*      \\
          & Lead II    & 95.0     & 15.3             & \tre{25.8}*     & \due{28.6}* & 6.9         & 12.1       & 8.9        & 11.9        & \uno{30.2}       \\
          & One random & 95.0     & \due{18.3(1.6)}  & \tre{17.7(1.3)} & 12.6(2.3)   & 13.1(2.5)   & 8.2(1.1)   & 6.6(0.8)   & 12.4(0.8)   & \uno{24.1(1.6)}  \\
          & Two random & 95.0     & \uno{37.0(1.2)}  & \tre{26.5(3.5)} & 15.7(2.2)   & 22.0(1.2)   & 12.9(2.5)  & 10.9(0.3)  & 20.6(1.7)   & \due{36.5(0.6)}* \\
\midrule
\multirow{5}{*}{PTB-XL Form$^\dagger$}
          & 12 leads   & 95.0     & \tre{55.5}       & \due{61.9}*     & 53.6        & 55.2        & 42.2       & 44.4       & \uno{64.7}  & 42.2             \\
          & Lead I     & 95.0     & \tre{11.0}       & \due{23.0}*     & 7.4         & 5.9         & 4.4        & 10.0       & 5.2         & \uno{24.6}       \\
          & Lead II    & 95.0     & 8.4              & \due{22.0}*     & \tre{17.9}* & 5.8         & 6.1        & 10.5       & 6.6         & \uno{22.5}       \\
          & One random & 95.0     & \tre{13.1(1.3)}  & \due{13.6(2.0)} & 10.5(1.9)   & 7.8(2.0)    & 6.3(0.8)   & 6.2(1.5)   & 10.2(2.2)   & \uno{21.5(2.3)}  \\
          & Two random & 95.0     & \due{23.3(1.2)}* & \tre{19.0(2.8)} & 14.2(0.8)   & 10.5(1.9)   & 7.9(0.9)   & 8.3(3.0)   & 12.7(1.7)   & \uno{25.6(0.8)}  \\
\midrule
\multirow{5}{*}{PTB-XL Rhythm$^\dagger$}
          & 12 leads   & 95.0     & \due{80.8}       & \uno{86.1}      & 76.8        & 71.8        & 63.0       & \tre{80.0} & 79.9        & 68.6             \\
          & Lead I     & 95.0     & \due{42.9}*      & \uno{46.8}      & 4.4         & 5.6         & 7.0        & 11.6       & 12.8        & \tre{38.5}*      \\
          & Lead II    & 95.1     & 41.8             & \uno{62.1}      & \due{56.5}* & 7.1         & 6.6        & 8.5        & 20.9        & \tre{51.6}*      \\
          & One random & 95.1     & \uno{46.4(2.7)}  & \tre{26.0(4.9)} & 22.5(4.6)   & 13.9(2.6)   & 8.4(4.9)   & 12.1(2.8)  & 19.4(3.2)   & \due{41.2(2.8)}* \\
          & Two random & 95.1     & \uno{64.0(3.0)}  & \tre{40.4(3.4)} & 28.8(4.8)   & 25.5(4.1)   & 22.7(7.0)  & 23.8(5.8)  & 34.2(2.6)   & \due{51.0(3.6)}  \\
\midrule
\multirow{5}{*}{PTB-XL Sub}
          & 12 leads   & 95.0     & \tre{73.1}*      & \due{73.3}*     & 65.8        & 72.1*       & 59.1       & 64.6       & \uno{74.7}  & 68.5             \\
          & Lead I     & 95.0     & \tre{16.1}       & \due{30.0}*     & 4.8         & 8.6         & 4.5        & 9.2        & 3.5         & \uno{31.5}       \\
          & Lead II    & 95.0     & 18.7             & \uno{32.7}      & \tre{27.0}* & 11.0        & 6.1        & 8.0        & 7.0         & \due{31.0}*      \\
          & One random & 95.0     & \due{24.9(2.5)}  & \tre{19.9(4.4)} & 13.8(2.6)   & 11.8(1.9)   & 7.3(2.2)   & 7.3(1.2)   & 10.2(1.4)   & \uno{27.9(2.2)}  \\
          & Two random & 95.0     & \uno{43.6(2.9)}  & \tre{27.3(3.4)} & 17.1(2.2)   & 23.0(3.7)   & 9.0(1.8)   & 15.1(0.6)  & 18.2(3.1)   & \due{40.0(3.6)}  \\
\midrule
\multirow{5}{*}{PTB-XL Super}
          & 12 leads   & 95.0     & \uno{69.3}       & 65.6            & 54.5        & \tre{67.8}* & 58.1       & 60.8       & \due{68.9}* & 61.9             \\
          & Lead I     & 95.0     & \tre{9.0}        & \due{23.2}      & 4.5         & 4.9         & 5.8        & 6.5        & 4.3         & \uno{31.0}       \\
          & Lead II    & 95.0     & 12.0             & \uno{31.4}      & \tre{18.1}  & 7.7         & 6.4        & 8.7        & 6.8         & \due{30.6}*      \\
          & One random & 95.0     & \due{15.1(0.6)}  & \tre{14.9(1.0)} & 8.9(0.5)    & 9.9(0.6)    & 6.7(0.7)   & 5.3(1.0)   & 6.2(0.8)    & \uno{22.7(1.2)}  \\
          & Two random & 95.0     & \due{25.0(0.9)}  & \tre{23.3(1.0)} & 12.9(0.7)   & 17.4(1.1)   & 9.0(1.0)   & 7.8(0.7)   & 10.1(0.8)   & \uno{32.1(1.8)}  \\
\midrule
\multirow{5}{*}{EchoNext}
          & 12 leads   & 95.0     & 33.1             & 35.3            & \tre{35.6}  & \due{36.6}  & 34.0       & 31.3       & \uno{39.5}  & 29.5             \\
          & Lead I     & 95.0     & \due{16.6}       & \uno{22.2}      & 5.7         & 10.7        & 10.0       & 11.9       & 11.7        & \tre{14.2}       \\
          & Lead II    & 95.0     & \uno{18.1}       & \due{16.2}*     & 11.4        & 10.6        & 7.4        & 11.4       & 8.0         & \tre{16.1}*      \\
          & One random & 95.0     & \uno{16.9(0.8)}  & \tre{13.4(1.5)} & 8.0(0.6)    & 9.6(0.5)    & 7.5(0.2)   & 8.9(0.5)   & 8.8(0.5)    & \due{14.3(0.8)}  \\
          & Two random & 95.0     & \due{21.4(1.3)}* & \tre{17.7(1.0)} & 10.3(0.5)   & 13.9(0.8)   & 8.1(0.9)   & 12.3(0.8)  & 15.4(1.1)   & \uno{21.6(1.0)}  \\
\midrule
\multirow{5}{*}{CPSC2018}
          & 12 leads   & 95.1     & \due{95.4}*      & \tre{94.5}      & 86.2        & 91.3        & \uno{97.0} & 86.9       & \tre{94.5}  & 82.6             \\
          & Lead I     & 95.1     & 17.6             & \uno{57.8}      & 3.8         & \tre{21.8}  & 9.7        & 8.0        & 7.4         & \due{38.7}       \\
          & Lead II    & 95.1     & 23.7             & \uno{66.2}      & \due{48.2}  & 12.2        & 6.2        & 7.0        & 13.3        & \tre{47.1}       \\
          & One random & 95.1     & \tre{36.0(1.0)}  & \due{42.1(1.2)} & 30.8(2.7)   & 21.6(2.3)   & 8.6(1.5)   & 7.1(1.0)   & 21.0(1.7)   & \uno{49.4(1.8)}  \\
          & Two random & 95.1     & \tre{56.6(1.8)}  & \due{57.3(2.0)} & 43.4(2.3)   & 38.9(2.0)   & 11.9(0.9)  & 18.8(1.1)  & 38.4(1.9)   & \uno{63.4(0.8)}  \\
\bottomrule
\end{tabular}
}
\end{table*}

\begin{table*}[p]
\centering
{\small\textbf{Supplementary Table~\ref{tab:supp_sens95spec} (continued).}}\par\medskip
\setlength{\tabcolsep}{3pt}
\renewcommand{\arraystretch}{0.9}
\resizebox{\textwidth}{!}{
\begin{tabular}{llc cccccccc}
\toprule
Benchmark & Input      & Achieved & ECGFounder       & ECG-JEPA         & ST-MEM           & MERL        & ECGFM-KED   & HuBERT-ECG & ECG-CPC     & \textbf{RELAY}   \\
\midrule\multirow{5}{*}{CPSC-Extra}
          & 12 leads   & 95.2     & \due{79.4}*      & 71.6             & 71.4             & 72.3        & \tre{74.8}* & 60.8       & \uno{80.3}  & 53.1             \\
          & Lead I     & 95.2     & \due{26.9}       & \uno{46.0}       & 2.4              & 17.1        & 7.3         & 10.0       & 15.0        & \tre{26.0}       \\
          & Lead II    & 95.2     & 30.5             & \uno{43.3}       & \due{37.9}*      & 9.1         & 8.3         & 13.1       & 15.1        & \tre{32.6}*      \\
          & One random & 95.2     & \uno{41.0(2.8)}  & \due{32.4(4.2)}  & 17.5(2.3)        & 18.0(2.0)   & 7.4(2.4)    & 9.1(0.7)   & 23.2(3.4)   & \tre{31.2(1.3)}  \\
          & Two random & 95.2     & \uno{57.4(3.5)}  & \tre{38.0(3.3)}  & 22.4(2.4)        & 29.4(2.9)   & 12.9(2.9)   & 19.7(2.3)  & 37.6(2.4)   & \due{39.6(3.0)}  \\
\midrule
\multirow{5}{*}{Chapman}
          & 12 leads   & 95.0     & \due{82.2}*      & \uno{84.0}       & 75.1             & 73.3        & 61.3        & 70.7       & \tre{81.8}* & 76.3             \\
          & Lead I     & 95.1     & \tre{17.7}       & \due{39.2}*      & 7.0              & 15.1        & 5.7         & 11.3       & 11.5        & \uno{39.4}       \\
          & Lead II    & 95.1     & 26.1             & \uno{50.4}       & \tre{33.0}       & 21.2        & 11.4        & 11.8       & 12.3        & \due{45.5}*      \\
          & One random & 95.1     & \due{31.8(1.3)}  & \tre{31.4(3.2)}  & 17.2(1.8)        & 18.4(0.7)   & 8.7(1.5)    & 9.9(1.5)   & 14.3(1.3)   & \uno{39.4(2.2)}  \\
          & Two random & 95.1     & \uno{50.6(2.1)}  & \tre{47.0(3.0)}  & 21.4(3.2)        & 27.1(1.5)   & 14.2(2.8)   & 18.2(1.4)  & 25.5(2.1)   & \due{49.2(2.1)}* \\
\midrule
\multirow{5}{*}{Chapman Rhythm$^\dagger$}
          & 12 leads   & 95.0     & \due{95.6}       & \uno{97.6}       & 93.0             & 88.2        & 81.7        & 91.1       & \tre{94.8}* & 93.2             \\
          & Lead I     & 95.1     & \tre{38.2}       & \due{55.1}*      & 3.0              & 17.2        & 7.9         & 9.8        & 11.5        & \uno{65.5}       \\
          & Lead II    & 95.1     & 46.8             & \uno{85.4}       & \tre{58.5}       & 23.3        & 13.0        & 10.7       & 10.3        & \due{69.4}       \\
          & One random & 95.1     & \due{54.4(1.2)}  & \tre{45.2(5.2)}  & 32.4(4.0)        & 24.0(0.8)   & 11.6(3.3)   & 11.3(3.1)  & 19.0(6.7)   & \uno{65.0(3.0)}  \\
          & Two random & 95.1     & \uno{79.5(0.6)}  & \tre{63.4(5.1)}  & 40.3(2.9)        & 39.0(2.9)   & 23.6(6.9)   & 22.5(2.1)  & 36.9(4.6)   & \due{77.9(4.3)}* \\
\midrule
\multirow{5}{*}{Georgia}
          & 12 leads   & 95.1     & 79.8             & \uno{88.5}       & 67.5             & \tre{81.9}  & 65.1        & 52.3       & \due{82.4}  & 53.8             \\
          & Lead I     & 95.1     & \tre{25.5}       & \uno{33.5}       & 5.4              & 15.9        & 11.1        & 4.2        & 11.1        & \due{27.3}*      \\
          & Lead II    & 95.1     & 18.0             & \uno{34.0}       & \due{33.4}*      & 10.2        & 12.4        & 7.2        & 8.9         & \tre{32.9}*      \\
          & One random & 95.1     & \uno{33.0(3.5)}  & \tre{23.4(2.7)}  & 16.6(1.5)        & 17.9(1.5)   & 10.0(0.9)   & 4.9(0.9)   & 16.1(2.4)   & \due{29.3(1.0)}  \\
          & Two random & 95.1     & \uno{44.4(1.1)}  & \tre{30.2(5.0)}  & 20.0(1.9)        & 27.4(2.1)   & 15.9(2.0)   & 8.7(0.7)   & 29.1(2.4)   & \due{35.2(0.7)}  \\
\midrule
\multirow{5}{*}{Hefei}
          & 12 leads   & 95.0     & 79.5             & 80.6             & 79.8             & \due{88.0}* & \tre{82.5}  & 80.1       & \uno{88.7}  & 79.7             \\
          & Lead I     & 95.1     & \tre{35.7}       & \due{39.6}       & 7.3              & 13.0        & 6.9         & 14.8       & 8.6         & \uno{45.9}       \\
          & Lead II    & 95.1     & \due{48.7}       & 41.2             & \tre{47.8}       & 17.7        & 11.0        & 18.1       & 17.2        & \uno{54.5}       \\
          & One random & 95.1     & \due{27.7(5.1)}  & \due{27.7(3.5)}  & \tre{21.7(3.0)}  & 16.5(3.4)   & 8.2(3.0)    & 11.1(3.0)  & 14.5(3.6)   & \uno{43.7(3.7)}  \\
          & Two random & 95.1     & \tre{33.0(4.1)}  & \due{37.3(3.1)}  & 25.0(2.3)        & 26.5(1.8)   & 13.3(2.4)   & 17.4(1.7)  & 24.8(3.2)   & \uno{56.6(2.7)}  \\
\midrule
\multirow{5}{*}{Ningbo}
          & 12 leads   & 95.0     & 92.6*            & \tre{93.4}*      & 85.5             & \due{93.9}* & 85.9        & 77.5       & \uno{94.3}  & 81.0             \\
          & Lead I     & 95.0     & 18.8             & \uno{54.5}       & 6.0              & 13.9        & 16.5        & \tre{19.5} & 11.6        & \due{45.1}       \\
          & Lead II    & 95.0     & 29.0             & \uno{61.9}       & \tre{34.3}       & 20.5        & 14.9        & 18.6       & 17.5        & \due{54.1}       \\
          & One random & 95.0     & \due{37.9(2.3)}* & \tre{34.1(2.7)}  & 18.0(1.2)        & 20.6(1.0)   & 13.2(1.6)   & 19.2(0.1)  & 17.2(2.2)   & \uno{40.8(1.6)}  \\
          & Two random & 95.0     & \uno{60.6(2.7)}  & \tre{49.3(1.8)}  & 27.9(2.2)        & 35.7(1.3)   & 24.5(1.0)   & 32.2(1.1)  & 31.9(0.6)   & \due{56.8(1.0)}  \\
\midrule
\multirow{5}{*}{PTB}
          & 12 leads   & 95.6     & \uno{27.1}       & \tre{25.4}*      & 18.1*            & \due{27.0}* & 8.6         & 13.5*      & 19.4*       & 19.5*            \\
          & Lead I     & 95.6     & \uno{13.3}       & \due{10.5}*      & 3.3*             & 8.7*        & \tre{10.4}* & 9.2*       & 6.6*        & 7.7*             \\
          & Lead II    & 95.6     & 8.5*             & \due{12.9}*      & \uno{14.1}       & \tre{9.5}*  & 8.4*        & 6.1*       & 4.6         & 4.7*             \\
          & One random & 95.6     & \uno{16.6(1.5)}  & \due{14.3(1.8)}* & 11.5(4.2)*       & 9.3(1.8)*   & 4.9(1.9)    & 7.9(3.4)   & 6.9(2.7)    & \tre{13.8(3.0)}* \\
          & Two random & 95.6     & \uno{21.8(1.7)}  & \tre{17.4(1.7)}* & \due{21.0(6.3)}* & 10.4(2.6)   & 4.6(1.7)    & 13.9(3.2)* & 5.5(1.2)    & 15.2(1.2)        \\
\midrule
\multirow{5}{*}{SPH}
          & 12 leads   & 95.0     & \due{90.7}*      & \uno{93.8}       & 80.2             & 82.7        & 75.7        & 80.0       & \tre{90.2}* & 83.4             \\
          & Lead I     & 95.0     & \tre{25.1}       & \due{50.8}*      & 7.6              & 13.2        & 14.0        & 16.3       & 20.1        & \uno{52.1}       \\
          & Lead II    & 95.0     & 26.9             & \uno{63.3}       & \tre{47.0}       & 23.1        & 13.2        & 15.6       & 23.9        & \due{61.7}*      \\
          & One random & 95.1     & \tre{37.2(1.5)}  & \due{41.7(4.3)}* & 20.0(1.2)        & 20.1(2.7)   & 10.1(1.1)   & 10.1(0.9)  & 19.9(2.3)   & \uno{43.4(2.6)}  \\
          & Two random & 95.0     & \tre{52.7(2.1)}* & \uno{57.0(1.8)}  & 25.9(1.9)        & 28.8(2.1)   & 13.5(3.7)   & 22.4(2.8)  & 30.2(2.0)   & \due{56.2(3.4)}* \\
\midrule
\multirow{5}{*}{ZZU pECG}
          & 12 leads   & 95.0     & \tre{65.7}*      & \uno{70.8}       & 65.0             & 61.3        & 51.8        & 57.4       & \due{69.3}* & 59.2             \\
          & Lead I     & 95.0     & \tre{30.3}       & \uno{44.6}       & 4.8              & 15.1        & 10.0        & 10.6       & 9.4         & \due{33.5}       \\
          & Lead II    & 95.0     & 34.0             & \uno{50.8}       & \tre{40.2}       & 18.0        & 23.8        & 8.6        & 18.5        & \due{46.0}*      \\
          & One random & 95.0     & \due{40.6(1.9)}* & \uno{40.8(3.7)}  & 19.8(1.7)        & 22.8(0.5)   & 14.0(2.3)   & 12.8(1.4)  & 16.2(1.5)   & \tre{40.2(1.8)}* \\
          & Two random & 95.0     & \uno{53.2(0.8)}  & \due{50.6(1.2)}* & 24.4(1.7)        & 33.1(1.5)   & 22.0(1.7)   & 16.7(0.7)  & 27.2(1.6)   & \tre{48.7(1.6)}* \\

\bottomrule
\end{tabular}
}
\end{table*}

\begin{table*}[p]
\centering
\caption{\textbf{Specificity at 90\% sensitivity.} For every condition, specificity is deliberately read from the ROC curve of the test set at the threshold giving a sensitivity above 90\%, then the macro average over conditions $\times 100$ is reported (conditions with no positive test sample are left out). Like AUROC, these values summarise the ROC curve: they are the highest reachable by any threshold at this operating point, not the result of a deployable threshold (Supplementary Table~\ref{tab:supp_threshold_transfer}). Achieved: sensitivity actually reached by the thresholds, averaged over conditions and models. It exceeds the target because sensitivity moves in large steps when a condition has few positives, which makes the reported specificities conservative. When leads are sampled randomly, mean(std) over 5 seeds is reported. ST-MEM runs on the available leads only, with their true lead embeddings. *: not significantly worse than the best model of the row (paired bootstrap of the difference, 1000 resamples with thresholds re-derived, 95\% CI includes 0). Bold, underlined and italic values are the best, second-best and third-best of each row. The MIMIC-IV-ECG benchmarks are not included. $^\dagger$Evaluation-only benchmark.}
\label{tab:supp_spec90sens}
\setlength{\tabcolsep}{3pt}
\renewcommand{\arraystretch}{0.9}
\resizebox{\textwidth}{!}{
\begin{tabular}{llc cccccccc}
\toprule
Benchmark & Input      & Achieved & ECGFounder       & ECG-JEPA        & ST-MEM          & MERL            & ECGFM-KED  & HuBERT-ECG      & ECG-CPC         & \textbf{RELAY}  \\
\midrule
\multirow{5}{*}{CODE 15\%}
          & 12 leads   & 92.9     & 97.7*            & \tre{98.1}*     & \tre{98.1}*     & 96.7*           & 96.3*      & \uno{98.5}      & \due{98.4}*     & 97.8*           \\
          & Lead I     & 92.9     & \uno{88.0}       & \tre{86.5}*     & 15.1            & 32.9            & 41.0       & 14.5            & 25.0            & \due{87.0}*     \\
          & Lead II    & 93.0     & \due{82.8}*      & \tre{82.2}*     & 75.8            & 28.0            & 39.9       & 9.8             & 20.9            & \uno{90.2}      \\
          & One random & 93.7     & \due{84.0(2.9)}* & 26.5(4.5)       & \tre{47.2(6.7)} & 45.7(3.9)       & 32.2(2.2)  & 11.4(1.0)       & 31.0(5.6)       & \uno{87.8(2.2)} \\
          & Two random & 93.3     & \due{93.8(1.6)}* & 53.6(4.1)       & 59.8(4.3)       & \tre{66.5(1.5)} & 46.1(6.6)  & 17.4(1.7)       & 35.1(5.6)       & \uno{95.0(1.0)} \\
\midrule
\multirow{5}{*}{PTB-XL All}
          & 12 leads   & 94.3     & \due{84.6}*      & \tre{83.6}*     & 77.0            & 79.8            & 74.9       & 77.4            & \uno{84.9}      & 77.3            \\
          & Lead I     & 94.3     & \tre{35.0}       & \due{38.4}      & 16.8            & 17.9            & 18.2       & 18.9            & 17.9            & \uno{48.9}      \\
          & Lead II    & 94.3     & 30.9             & \tre{41.3}      & \due{43.0}      & 17.9            & 17.0       & 20.9            & 22.4            & \uno{51.8}      \\
          & One random & 94.5     & \due{35.3(0.9)}  & 25.2(1.1)       & \tre{27.4(1.0)} & 21.7(1.3)       & 19.9(0.9)  & 17.9(1.0)       & 23.2(1.1)       & \uno{44.6(2.8)} \\
          & Two random & 94.4     & \due{54.4(1.4)}* & \tre{36.7(2.3)} & 31.9(0.9)       & 30.5(1.5)       & 25.1(2.3)  & 22.4(0.9)       & 34.7(1.2)       & \uno{55.8(1.5)} \\
\midrule
\multirow{5}{*}{PTB-XL Diag$^\dagger$}
          & 12 leads   & 94.4     & \uno{86.7}       & \tre{83.9}      & 76.4            & 82.0            & 76.6       & 78.5            & \due{85.1}*     & 82.0            \\
          & Lead I     & 94.4     & \tre{35.7}       & \due{36.3}      & 17.5            & 18.9            & 19.3       & 15.8            & 15.6            & \uno{48.3}      \\
          & Lead II    & 94.5     & 27.7             & \tre{38.6}      & \due{40.3}      & 20.3            & 19.9       & 19.1            & 23.7            & \uno{53.1}      \\
          & One random & 94.5     & \due{33.2(1.3)}  & 24.8(1.3)       & \tre{26.3(2.2)} & 23.2(1.2)       & 20.8(1.2)  & 16.8(2.1)       & 22.4(0.8)       & \uno{41.5(4.2)} \\
          & Two random & 94.5     & \due{53.8(1.1)}* & \tre{35.3(3.2)} & 30.7(0.9)       & 31.4(1.4)       & 25.3(3.3)  & 20.6(1.7)       & 33.8(1.7)       & \uno{56.3(1.9)} \\
\midrule
\multirow{5}{*}{PTB-XL Form$^\dagger$}
          & 12 leads   & 93.2     & \tre{76.1}       & \due{77.8}*     & 73.7            & 75.9            & 71.9       & 65.7            & \uno{81.2}      & 71.8            \\
          & Lead I     & 93.2     & \tre{24.8}       & \due{34.9}      & 15.5            & 13.5            & 13.4       & 18.8            & 20.4            & \uno{44.2}      \\
          & Lead II    & 93.2     & 22.9             & \tre{30.4}      & \due{37.1}*     & 14.9            & 12.1       & 15.2            & 18.3            & \uno{41.1}      \\
          & One random & 93.4     & \due{25.5(1.4)}  & \tre{22.8(2.7)} & 21.3(1.9)       & 17.2(2.0)       & 13.9(3.2)  & 13.9(1.4)       & 19.6(2.6)       & \uno{42.1(1.9)} \\
          & Two random & 93.3     & \due{42.9(4.3)}  & \tre{34.2(2.3)} & 25.1(1.7)       & 24.3(1.7)       & 18.1(2.9)  & 17.8(1.4)       & 26.5(3.0)       & \uno{49.3(3.0)} \\
\midrule
\multirow{5}{*}{PTB-XL Rhythm$^\dagger$}
          & 12 leads   & 94.8     & \due{88.6}*      & \uno{91.1}      & 84.0            & 76.8            & 70.9       & 86.7*           & \tre{87.4}      & 68.4            \\
          & Lead I     & 94.8     & \tre{50.1}*      & \due{52.7}*     & 17.1            & 18.3            & 19.2       & 28.9            & 27.2            & \uno{57.8}      \\
          & Lead II    & 94.8     & 53.7             & \uno{69.6}      & \tre{62.4}*     & 14.0            & 15.3       & 34.9            & 23.3            & \due{62.8}*     \\
          & One random & 95.0     & \due{57.0(3.1)}* & 31.0(2.5)       & \tre{40.2(2.9)} & 20.5(3.9)       & 24.3(6.6)  & 26.8(1.3)       & 30.9(4.4)       & \uno{59.2(3.5)} \\
          & Two random & 94.9     & \uno{74.3(3.9)}  & 47.3(1.7)       & 45.5(2.5)       & 35.1(6.4)       & 32.7(5.2)  & 34.9(0.9)       & \tre{49.9(4.4)} & \due{63.0(3.2)} \\
\midrule
\multirow{5}{*}{PTB-XL Sub}
          & 12 leads   & 92.9     & \due{81.1}*      & \tre{80.8}*     & 77.0            & 80.0*           & 70.8       & 76.5            & 80.4*           & \uno{81.9}      \\
          & Lead I     & 92.9     & \due{38.8}       & \tre{34.6}      & 11.4            & 18.7            & 10.5       & 17.0            & 13.9            & \uno{51.4}      \\
          & Lead II    & 92.9     & 30.8             & \due{40.6}      & \tre{37.3}      & 15.8            & 16.0       & 19.3            & 15.8            & \uno{50.4}      \\
          & One random & 93.2     & \due{37.1(1.9)}  & 18.2(2.7)       & \tre{23.9(2.4)} & 16.7(0.8)       & 12.7(1.5)  & 14.0(1.2)       & 16.4(1.5)       & \uno{42.4(1.6)} \\
          & Two random & 93.1     & \uno{57.7(0.4)}  & 26.7(3.0)       & 28.1(3.3)       & \tre{28.2(1.1)} & 15.0(1.6)  & 20.9(3.0)       & 23.9(1.6)       & \due{53.1(0.6)} \\
\midrule
\multirow{5}{*}{PTB-XL Super}
          & 12 leads   & 90.1     & \uno{81.5}       & 77.8            & 68.0            & \tre{79.5}*     & 72.0       & 73.1            & \due{80.3}*     & 74.8            \\
          & Lead I     & 90.1     & \tre{15.6}       & \due{27.3}      & 11.0            & 7.5             & 8.9        & 14.6            & 5.6             & \uno{42.2}      \\
          & Lead II    & 90.1     & 20.0             & \tre{27.0}      & \due{31.2}      & 11.3            & 10.8       & 14.1            & 11.6            & \uno{43.3}      \\
          & One random & 90.5     & \due{19.6(0.7)}  & 14.6(0.5)       & \tre{17.0(1.2)} & 14.5(0.5)       & 12.0(0.4)  & 11.6(0.5)       & 9.6(0.5)        & \uno{30.7(0.7)} \\
          & Two random & 90.5     & \due{37.1(0.9)}  & 17.5(1.0)       & \tre{22.5(1.0)} & 21.4(0.4)       & 13.8(0.8)  & 15.4(0.7)       & 12.8(1.0)       & \uno{43.0(0.9)} \\
\midrule
\multirow{5}{*}{EchoNext}
          & 12 leads   & 90.7     & \tre{52.3}*      & \due{54.9}*     & 51.1*           & 51.0            & 50.8       & 44.3            & \uno{55.3}      & 48.4            \\
          & Lead I     & 90.8     & \uno{37.9}       & \due{34.0}*     & 11.0            & 11.7            & 14.1       & 22.2            & 16.4            & \tre{27.5}      \\
          & Lead II    & 90.8     & \uno{37.2}       & \due{30.7}      & 20.9            & 16.6            & 11.9       & 20.8            & 9.1             & \tre{26.3}      \\
          & One random & 91.0     & \uno{30.9(0.8)}  & \tre{17.1(0.9)} & 15.8(0.9)       & 13.7(1.2)       & 10.7(1.1)  & \tre{17.1(0.4)} & 14.3(1.4)       & \due{24.7(1.2)} \\
          & Two random & 90.9     & \uno{36.2(1.8)}  & 21.7(2.1)       & 20.5(1.7)       & 20.5(0.5)       & 11.9(0.6)  & \tre{24.7(1.4)} & 23.6(1.0)       & \due{32.2(1.7)} \\
\midrule
\multirow{5}{*}{CPSC2018}
          & 12 leads   & 90.7     & \due{97.5}*      & \due{97.5}*     & 88.6            & 94.2            & \uno{99.3} & 91.6            & \tre{96.8}*     & 90.3            \\
          & Lead I     & 90.7     & \tre{33.0}       & \uno{70.2}      & 10.6            & 30.3            & 20.7       & 22.5            & 16.4            & \due{57.4}      \\
          & Lead II    & 90.7     & 34.4             & \uno{65.0}      & \due{59.0}*     & 21.9            & 14.8       & 21.0            & 20.7            & \tre{56.4}      \\
          & One random & 90.9     & \due{43.0(1.5)}  & \tre{41.2(1.4)} & 40.7(3.1)       & 29.7(1.2)       & 12.3(1.3)  & 16.9(0.8)       & 28.1(1.3)       & \uno{58.5(1.9)} \\
          & Two random & 90.8     & \due{68.2(2.4)}* & \tre{60.5(3.1)} & 51.9(2.1)       & 48.7(2.8)       & 17.1(1.5)  & 30.8(1.1)       & 50.2(0.6)       & \uno{69.7(0.9)} \\
\bottomrule
\end{tabular}
}
\end{table*}

\begin{table*}[p]
\centering
{\small\textbf{Supplementary Table~\ref{tab:supp_spec90sens} (continued).}}\par\medskip
\setlength{\tabcolsep}{3pt}
\renewcommand{\arraystretch}{0.9}
\resizebox{\textwidth}{!}{
\begin{tabular}{llc cccccccc}
\toprule
Benchmark & Input      & Achieved & ECGFounder       & ECG-JEPA         & ST-MEM          & MERL            & ECGFM-KED  & HuBERT-ECG  & ECG-CPC          & \textbf{RELAY}   \\
\midrule\multirow{5}{*}{CPSC-Extra}
          & 12 leads   & 97.2     & \uno{89.4}       & 80.7             & 79.2            & \tre{83.0}      & 82.5       & 70.4        & \due{85.5}       & 64.3             \\
          & Lead I     & 97.2     & \tre{36.2}       & \uno{55.1}       & 19.1            & 30.8            & 22.0       & 26.1        & 34.6             & \due{45.7}       \\
          & Lead II    & 97.2     & 33.9             & \due{51.0}       & \uno{59.4}      & 30.7            & 24.8       & 28.6        & 29.3             & \tre{48.0}       \\
          & One random & 97.3     & \uno{54.7(3.5)}  & \tre{39.3(3.7)}  & 37.2(2.3)       & 32.9(2.7)       & 23.3(4.2)  & 28.3(1.9)   & 38.8(4.7)        & \due{44.9(1.2)}  \\
          & Two random & 97.2     & \uno{70.7(1.6)}  & 52.3(3.2)        & 43.6(2.6)       & 46.0(3.6)       & 29.3(1.7)  & 35.2(2.1)   & \due{54.5(2.8)}  & \tre{53.6(1.7)}  \\
\midrule
\multirow{5}{*}{Chapman}
          & 12 leads   & 94.6     & \uno{91.6}       & \due{90.7}*      & 86.5            & 86.0            & 79.5       & 83.3        & \tre{89.2}*      & 85.2             \\
          & Lead I     & 94.6     & \tre{36.2}       & \due{52.8}       & 22.3            & 27.9            & 20.5       & 33.5        & 25.7             & \uno{56.8}       \\
          & Lead II    & 94.6     & 48.7             & \due{62.3}*      & \tre{52.6}      & 34.0            & 33.3       & 26.0        & 26.6             & \uno{65.3}       \\
          & One random & 94.8     & \due{48.4(2.6)}  & \tre{36.1(2.0)}  & 35.5(2.2)       & 34.0(2.2)       & 21.9(1.0)  & 26.6(1.8)   & 29.9(1.1)        & \uno{58.1(2.5)}  \\
          & Two random & 94.7     & \due{66.6(1.3)}* & \tre{53.1(3.1)}  & 38.9(3.9)       & 43.6(2.2)       & 28.8(2.7)  & 34.3(2.5)   & 42.5(1.0)        & \uno{67.2(0.9)}  \\
\midrule
\multirow{5}{*}{Chapman Rhythm$^\dagger$}
          & 12 leads   & 92.0     & \due{97.5}*      & \uno{98.6}       & 96.2*           & 92.2            & 90.1       & 93.6        & \tre{96.5}*      & 96.0*            \\
          & Lead I     & 92.0     & \tre{54.5}       & \due{79.8}*      & 15.9            & 31.2            & 24.8       & 28.6        & 20.7             & \uno{82.7}       \\
          & Lead II    & 92.0     & 63.1             & \uno{90.7}       & \tre{72.7}      & 33.2            & 44.2       & 26.2        & 14.9             & \due{84.2}*      \\
          & One random & 92.4     & \due{73.8(2.5)}  & 48.7(3.5)        & \tre{50.6(2.9)} & 40.6(1.4)       & 21.8(3.4)  & 23.4(3.2)   & 30.4(3.6)        & \uno{79.5(2.5)}  \\
          & Two random & 92.2     & \uno{89.3(0.9)}  & \tre{69.8(2.4)}  & 55.7(4.3)       & 51.0(2.2)       & 31.4(7.9)  & 34.6(2.8)   & 48.9(2.4)        & \due{88.4(2.0)}* \\
\midrule
\multirow{5}{*}{Georgia}
          & 12 leads   & 94.9     & \tre{86.9}       & \uno{91.7}       & 78.8            & 85.6            & 77.4       & 66.1        & \due{88.4}       & 65.9             \\
          & Lead I     & 94.9     & \tre{38.9}       & \due{40.2}       & 14.7            & 26.1            & 25.2       & 16.7        & 29.1             & \uno{48.3}       \\
          & Lead II    & 94.9     & 36.4             & \tre{47.7}*      & \uno{48.5}      & 19.6            & 23.2       & 19.8        & 20.4             & \due{48.1}*      \\
          & One random & 95.0     & \due{44.8(1.1)}* & 26.6(1.7)        & \tre{31.8(3.3)} & 29.5(2.4)       & 21.8(1.1)  & 17.1(2.2)   & 28.0(2.8)        & \uno{44.9(1.7)}  \\
          & Two random & 95.0     & \uno{60.5(1.2)}  & 37.5(2.5)        & 33.5(1.2)       & \tre{40.4(1.7)} & 30.0(1.0)  & 21.6(1.7)   & 40.3(2.3)        & \due{53.1(1.7)}  \\
\midrule
\multirow{5}{*}{Hefei}
          & 12 leads   & 94.5     & 84.3             & 85.8             & 88.6            & \due{91.7}*     & \tre{89.3} & 86.0        & \uno{91.9}       & 88.7             \\
          & Lead I     & 94.6     & \tre{48.4}       & \due{51.2}*      & 22.4            & 31.7            & 26.9       & 28.4        & 31.4             & \uno{55.0}       \\
          & Lead II    & 94.5     & 55.8             & \tre{56.5}       & \due{63.3}      & 36.6            & 31.6       & 36.3        & 35.0             & \uno{70.5}       \\
          & One random & 94.7     & \due{44.0(2.2)}  & 33.3(1.7)        & \tre{41.6(2.6)} & 32.8(2.8)       & 20.5(4.1)  & 28.9(2.6)   & 31.1(2.1)        & \uno{61.2(1.8)}  \\
          & Two random & 94.6     & \tre{44.9(4.4)}  & 43.1(2.4)        & 42.9(1.9)       & 42.9(0.8)       & 28.7(2.7)  & 37.5(2.1)   & \due{45.3(2.7)}  & \uno{69.9(1.7)}  \\
\midrule
\multirow{5}{*}{Ningbo}
          & 12 leads   & 94.6     & \due{96.3}       & \tre{95.7}       & 91.2            & 95.6            & 92.3       & 89.0        & \uno{97.3}       & 89.2             \\
          & Lead I     & 94.6     & 31.8             & \uno{64.2}       & 18.6            & 30.6            & 28.9       & \tre{33.3}  & 23.2             & \due{58.5}       \\
          & Lead II    & 94.6     & 46.1             & \uno{72.9}       & \tre{51.8}      & 38.1            & 24.5       & 35.5        & 31.0             & \due{64.3}       \\
          & One random & 95.2     & \due{50.2(1.6)}  & 20.6(2.1)        & 34.3(1.5)       & \tre{36.0(1.2)} & 25.4(1.5)  & 33.7(1.4)   & 26.9(1.3)        & \uno{57.5(1.2)}  \\
          & Two random & 95.0     & \due{71.9(1.0)}* & 42.0(2.4)        & 41.0(1.3)       & \tre{50.5(1.3)} & 37.8(2.1)  & 46.2(1.1)   & 45.0(1.9)        & \uno{72.2(2.3)}  \\
\midrule
\multirow{5}{*}{PTB}
          & 12 leads   & 97.6     & \tre{47.4}*      & 44.7*            & 44.1*           & 46.9*           & 33.2       & \due{49.0}* & \uno{49.8}       & 38.5*            \\
          & Lead I     & 97.6     & \tre{33.4}*      & \due{33.6}*      & 17.1            & 25.4*           & 21.5       & \uno{34.8}  & 28.3*            & 25.6*            \\
          & Lead II    & 97.6     & 26.6*            & \due{35.7}*      & \tre{35.3}*     & 25.5            & 17.4       & 28.3*       & 18.6             & \uno{37.2}       \\
          & One random & 97.6     & \uno{36.4(2.0)}  & \tre{30.5(4.1)}* & 28.5(2.2)       & 25.7(1.8)       & 22.7(4.2)  & 26.3(4.5)*  & \due{32.7(2.4)}* & 28.4(2.2)*       \\
          & Two random & 97.6     & \uno{42.6(1.7)}  & \due{36.1(1.5)}* & 32.1(4.9)       & 27.0(1.5)       & 22.7(3.6)  & 35.2(2.6)*  & \tre{35.3(1.7)}* & 33.6(3.4)        \\
\midrule
\multirow{5}{*}{SPH}
          & 12 leads   & 95.1     & \uno{95.5}       & \tre{95.0}*      & 91.8            & 89.6            & 86.0       & 91.2        & \due{95.2}*      & 90.4             \\
          & Lead I     & 95.1     & \tre{40.7}       & \due{60.5}*      & 17.7            & 29.0            & 23.5       & 29.9        & 28.5             & \uno{65.1}       \\
          & Lead II    & 95.1     & 37.0             & \uno{74.8}       & \tre{62.6}      & 36.4            & 22.0       & 30.7        & 33.1             & \due{67.6}       \\
          & One random & 95.4     & \due{49.3(1.9)}  & 33.1(3.2)        & \tre{39.6(2.5)} & 34.8(2.4)       & 23.5(4.4)  & 25.2(1.7)   & 37.2(0.4)        & \uno{59.7(2.3)}  \\
          & Two random & 95.3     & \due{64.6(2.6)}  & \tre{54.0(3.1)}  & 43.9(1.5)       & 44.6(2.5)       & 27.7(4.0)  & 37.0(2.8)   & 46.6(3.7)        & \uno{70.8(2.1)}  \\
\midrule
\multirow{5}{*}{ZZU pECG}
          & 12 leads   & 95.7     & \due{84.2}*      & \uno{84.5}       & 81.5            & 81.5            & 75.6       & 77.0        & \tre{82.4}*      & 76.9             \\
          & Lead I     & 95.7     & \tre{52.4}       & \uno{62.5}       & 25.7            & 35.7            & 22.6       & 29.0        & 31.7             & \due{55.0}       \\
          & Lead II    & 95.7     & 60.2             & \due{64.8}*      & \tre{62.8}      & 46.3            & 51.4       & 27.5        & 38.3             & \uno{65.4}       \\
          & One random & 95.8     & \uno{65.3(2.1)}  & \tre{54.2(2.8)}  & 42.9(2.1)       & 41.9(1.3)       & 32.3(2.8)  & 31.2(1.0)   & 34.8(2.4)        & \due{59.4(2.2)}  \\
          & Two random & 95.7     & \uno{74.4(0.5)}  & \tre{67.4(1.3)}  & 46.5(0.9)       & 57.0(0.8)       & 45.0(1.5)  & 39.4(0.8)   & 48.7(2.1)        & \due{67.8(2.0)}  \\

\bottomrule
\end{tabular}
}
\end{table*}

\end{appendices}

\clearpage%
{\renewcommand\refname{Supplementary References}%
}

\end{document}